\documentclass[10pt,twocolumn,letterpaper]{article}

\usepackage[pagenumbers]{cvpr} %

\newcommand{\hyt}[1]{#1}
\definecolor{teal}{HTML}{008080}
\definecolor{maroon}{HTML}{800000}
\newcommand{\ymyi}[1]{#1}
\usepackage{booktabs}
\usepackage{graphicx}
\usepackage{colortbl}  %
\definecolor{rankfirst}{RGB}{244,128,128}
\definecolor{ranksecond}{RGB}{248,180,130}
\definecolor{rankthird}{RGB}{255,230,153}
\newcommand{\best}[1]{\cellcolor{rankfirst}#1}
\newcommand{\second}[1]{\cellcolor{ranksecond}#1}
\newcommand{\third}[1]{\cellcolor{rankthird}#1}
\newcommand{\rankbox}[2]{{\setlength{\fboxsep}{0.8pt}\colorbox{#1}{\strut #2}}}
\usepackage{multirow}
\usepackage{makecell}  %
\usepackage{adjustbox} %
\usepackage{algorithm}
\usepackage{algpseudocode}
\definecolor{cvprblue}{rgb}{0.21,0.49,0.74}
\usepackage[pagebackref,breaklinks,colorlinks,allcolors=cvprblue]{hyperref}

\def\confName{CVPR}
\def\confYear{2027}

\title{DSD-GS: Snapshot Dynamic--Static Decomposition of Gaussian Splatting\\for Efficient Dynamic Scene Reconstruction and Fast Rendering}

\author{Youngtae Han \qquad Youngmin Yi\\
Department of Artificial Intelligence Engineering, Sogang University\\
{\small\url{https://young0tete.github.io/dsd-gs/}}
}
\makeatletter
\def\@maketitle{%
   \newpage\null
   \vskip -0.1in
   \begin{center}%
      {\Large \bf \@title \par}%
      \vspace*{12pt}%
      {\large\lineskip .3em\begin{tabular}[t]{c}\@author\end{tabular}\par}%
      \vskip .3em
   \end{center}}
\makeatother
\newcommand{\suppcite}[1]{~\cite{#1}}
\begin{document}

\twocolumn[{%
\renewcommand\twocolumn[1][]{#1}%
\maketitle
\begin{center}
\centering
\captionsetup{type=figure}
\includegraphics[width=\textwidth]{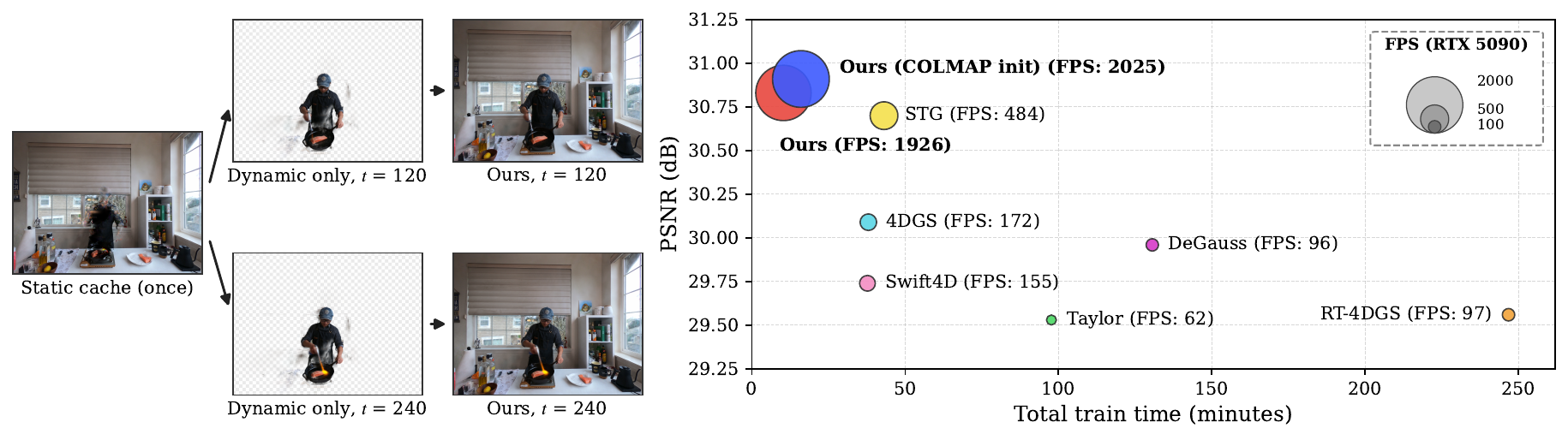}
\captionof{figure}{Left: DSD-GS caches the rendering of the frozen static Gaussians once per camera and, at each time $t$, rasterizes only the dynamic Gaussians over the cache. Right: PSNR versus total train time, including initialization, on Neural 3D~\cite{dynerf}.}
\label{fig:teaser}
\end{center}%
\vspace{0.5em}
}]

\begin{abstract}
Dynamic Gaussian splatting reconstructs multi-view dynamic scenes with high fidelity, but \hyt{its \ymyi{cost} limits} AR/VR and edge applications. 
\hyt{Existing dynamic--static decomposition methods jointly \ymyi{optimize geometry and motion, repeatedly rendering and updating} static Gaussians. We \ymyi{propose} DSD-GS, a snapshot dynamic--static decomposition framework that learns} \ymyi{scene geometry before motion, freezes static Gaussians before dynamic optimization, and caches their rendering. Subsequent training and rendering thus process only dynamic Gaussians. Bidirectional Label Correction, Keyframe Propagation, and Emergent-Content Seeding address labeling errors and content emerging after the first frame.}
\ymyi{%
To further reduce dynamic rendering cost, we propose Render-Cost Regularization, as we observe that it is governed by the tile coverage and visible duration of Gaussians rather than their number.} %
\ymyi{ %
Finally, feed-forward Gaussian splatting can replace COLMAP but predicts inaccurate depth. Prior work removes such Gaussians, whereas our Multi-View Ray Relocation corrects their depth.} %
\ymyi{ %
On Neural 3D and Technicolor, DSD-GS accelerates reconstruction by 3–4$\times$ and rendering by 4–5$\times$ on an RTX 5090 and about 8$\times$ on a Jetson AGX Orin compared with STG, while achieving comparable quality.
}
\end{abstract}

\section{Introduction}
\label{sec:intro}

\hyt{Dynamic Gaussian splatting reconstructs multi-view dynamic scenes with high fidelity, but its cost limits applications such as AR/VR content creation and volumetric video, which need results right after capture.}
\hyt{Training takes tens of minutes to hours per scene after several minutes of COLMAP~\cite{colmap} initialization, and rendering is not real-time on resource-constrained devices. For example, STG~\cite{stg} renders Technicolor~\cite{technicolor} at only 12.9 FPS on a Jetson AGX Orin.}
\hyt{A key source of this cost is that existing methods~\cite{stg,taylor,4dgs,rt-4dgs} treat every primitive alike, learning motion, computing gradients, and rendering at every frame even for static primitives.}

Existing dynamic--static decomposition methods~\cite{swift4d,mixvoxels,degauss} \hyt{exploit the fact that static regions such as the background need no temporal model.}
\hyt{However, they learn the geometry and motion of each primitive jointly and label it along the way, so they still optimize and render the static components at every iteration.}
\hyt{To address this limitation, we propose DSD-GS, a snapshot-based dynamic--static decomposition framework that learns the geometry of the scene first and its motion afterward.}
\hyt{We freeze the static Gaussians trained on the first frame before dynamic optimization and cache their rendering, so that each dynamic-training iteration} renders and optimizes only the dynamic Gaussians, which shortens the training on Technicolor by 1.7$\times$ compared with rendering the frozen static Gaussians at every iteration.
\hyt{Compositing the dynamic Gaussians over this cache instead of sorting them jointly with the static Gaussians (Static-Caching Rasterization) also prevents} the static Gaussians from occluding the dynamic ones and weakening their gradients, \hyt{which improves the PSNR by 0.78\,dB on Technicolor.}

However, the snapshot \ymyi{approach} raises two challenges.
\hyt{First, unlike existing methods~\cite{ex4dgs,degauss} that \ymyi{progressively} refine \ymyi{dynamic–static} labels while optimizing over all frames and views, the snapshot \ymyi{determines the initial labels using optical flow before} dynamic optimization.}
\ymyi{Consequently, misclassified Gaussians may remain throughout training. To address this, we propose Bidirectional Label Correction, which revisits the initial optical-flow-based labels before full-sequence dynamic training. Specifically, we relabel dynamic Gaussians that exhibit negligible motion along their flow trajectories as static (D$\rightarrow$S), and static Gaussians whose position gradients are inconsistent across keyframes as dynamic (S$\rightarrow$D).
}
\hyt{Second, since the dynamic Gaussians are initialized only from the first frame, they \ymyi{struggle to capture} motion far from it and cannot represent content that emerges later, such as newly appearing objects or disoccluded regions.
}
\ymyi{To address this, we propose a dynamic expansion strategy that propagates dynamic Gaussians along optical-flow trajectories to later keyframes}
\hyt{%
and seeds new Gaussians with feed-forward Gaussian splatting (FFGS) in the regions that no first-frame pixel reaches (Emergent-Content Seeding).}

\hyt{With the static part cached, the remaining rendering cost lies in the dynamic Gaussians.}
Existing methods~\cite{mini-splatting,gaussianspa,4dgs1k} accelerate rendering by reducing the number of Gaussians, \hyt{but we observe that the per-frame rendering time is explained by the number of screen tiles covered by the Gaussians ($R^2=0.996$) rather than their number ($R^2=0.39$).}
\hyt{We therefore \ymyi{propose to regularize the cost of each Gaussian over a sequence (Render-Cost Regularization), which penalizes} each \ymyi{Gaussian's tile coverage multiplied by its visible duration} during training, accelerating rendering from 425 to 1033 FPS on Technicolor without additional pruning.} %
Since Keyframe Propagation \ymyi{increases} the number of dynamic Gaussians, we also skip Gaussians whose opacity at the current time is \hyt{too low to change any pixel} (Temporal Culling), without which the rendering speed falls to 401 FPS.

\hyt{Once training and rendering are accelerated, initialization takes a larger share of the total time, as dense COLMAP on the first frame~\cite{4dgs,degauss,swift4d} takes about 6 minutes on Neural 3D~\cite{dynerf}, more than half of our training time, and sparse SfM on all frames~\cite{stg,taylor} takes about 16 minutes.}
\hyt{Pixel-aligned FFGS can replace COLMAP in seconds, but its predicted depth is inaccurate on camera configurations unlike its training data, lowering the PSNR of existing methods by 1.2--8.8\,dB on Technicolor.}
\hyt{MVSGaussian~\cite{mvsgaussian} removes the Gaussians whose depths are inconsistent across views, but on a camera rig, dense cameras produce 25--51$\times$ as many Gaussians as a sparse set, whereas a sparse set loses 64--96\% of them.}
\hyt{We instead keep all Gaussians predicted from a sparse set of cameras and correct their depths along their pixel rays using the other training cameras (Multi-View Ray Relocation), which also reduces the drop of existing methods to 0.3--2.9\,dB.}

On Neural 3D and Technicolor, DSD-GS accelerates total reconstruction, including initialization and training, by 3--4$\times$ and rendering by 4--5$\times$ on an RTX 5090 and about 8$\times$ on a Jetson AGX Orin compared with STG, while maintaining comparable or better quality. In temporally varying regions, it achieves the best PSNR, SSIM, and LPIPS, outperforming the strongest baseline by 1.38\,dB on Neural 3D and 0.50\,dB on Technicolor.

In summary, our contributions are as follows:
\begin{itemize}
  \item \textbf{Snapshot dynamic--static decomposition.} \ymyi{We propose a novel snapshot-based dynamic–static decomposition that identifies and refines the decomposition before dynamic optimization, so that the static Gaussians are frozen and rendered once per camera, with Keyframe Propagation and Bidirectional Label Correction that make this early decision practical.}
  \item \textbf{Render-Cost Regularization.} %
  \ymyi{We propose a render-cost regularizer based on the screen tiles covered by each dynamic Gaussian and its visible duration, which targets the actual rasterization workload rather than the number of Gaussians and, together with Temporal Culling, accelerates the remaining dynamic rendering without relying on aggressive pruning.
  }
\item \textbf{\ymyi{Fast initialization with depth correction.}} %
\ymyi{We propose Multi-View Ray Relocation, which corrects the inaccurate depths of pixel-aligned FFGS initialization by searching for cross-view-consistent locations along the camera rays instead of discarding inconsistent Gaussians, reducing the initialization overhead of dense COLMAP while preserving coverage and also improving existing reconstruction methods.}
\end{itemize}

\section{Related Work}
\label{sec:related}

\paragraph{Dynamic Gaussian Splatting.}
NeRF~\cite{nerf} and its variants~\cite{mip-nerf,zip-nerf,nsvf,point-nerf} represent static scenes with radiance fields, and NeRF-based methods extend them to dynamic scenes with time-conditioned fields~\cite{dynerf}, deformation fields~\cite{d-nerf,nerfies,hypernerf}, or space-time plane factorizations~\cite{hexplane,k-planes}, but volume rendering requires dense sampling along each ray, which limits their rendering speed.
3DGS~\cite{3dgs} and its extensions~\cite{sugar,2dgs,scaffold-gs,mip-splatting,fregs} represent static scenes with explicit Gaussian primitives for real-time rendering, and dynamic Gaussian splatting extends this representation to dynamic scenes.
One line of work keeps Gaussians in a canonical space and deforms them with neural fields~\cite{4dgs,d3dgs,sc-gs,ed3dgs}.
Another line models the attributes of each Gaussian as explicit functions of time~\cite{stg,gaussian-flow,taylor,ex4dgs} or directly optimizes 4D Gaussians in space and time~\cite{rt-4dgs}.
Dynamic Gaussian splatting has also been studied in streaming settings~\cite{3dgstream,instant-gs,4dgcpro}.

\paragraph{Dynamic--Static Decomposition.}
Several methods for dynamic scene representation separate static and dynamic content according to their goals~\cite{nerfplayer,mixvoxels,swift4d,ex4dgs,degauss,desire-gs}.
For efficiency, MixVoxels~\cite{mixvoxels} represents static voxels with a lightweight model, Swift4D~\cite{swift4d} applies a 4D hash deformation only to the Gaussians labeled dynamic from the temporal variance of pixels, and Ex4DGS~\cite{ex4dgs} separates the Gaussians with the largest motion as dynamic but still assigns linear motion to static ones.
For quality, DeGauss~\cite{degauss} decomposes scenes into foreground and background Gaussians with a learned probabilistic mask for distractor-free reconstruction.
These methods decide the labels while optimizing over the whole sequence, so their static part is rendered and optimized at every iteration and its rendering cannot be cached during training.
Moreover, caching the static part only at rendering time changes the order in which static and dynamic content are composited, so the rendered images differ from those they are trained on (\cref{sec:supp_scr_renderonly}).
In contrast, DSD-GS freezes the static Gaussians before dynamic optimization and replaces them with a cache used in both training and rendering, so the static part is neither rendered nor optimized during dynamic training and caching does not change the rendered images.

\begin{figure*}[t]
\centering
\includegraphics[width=\linewidth]{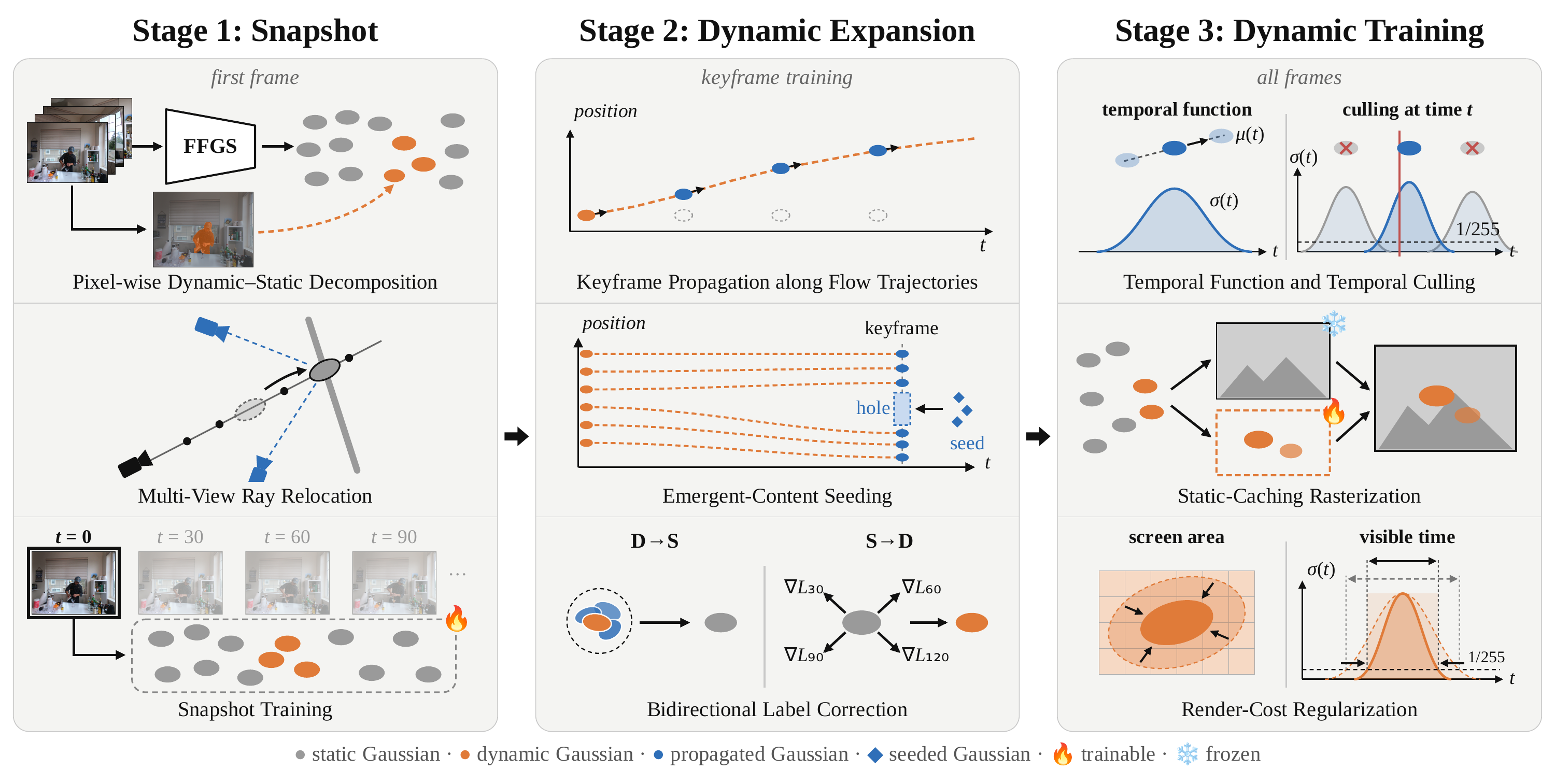}
\caption{Overview of DSD-GS. \textbf{Stage 1.} Initialized Gaussians from the first frame are labeled by optical flow, relocated along their rays, and trained, after which the static Gaussians are frozen. \textbf{Stage 2.} Dynamic Gaussians are propagated to keyframes, emergent content is seeded, and labels are corrected. \textbf{Stage 3.} Only the dynamic Gaussians are trained over all frames under Render-Cost Regularization and composited over the cached static rendering.
}
\label{fig:overview}
\end{figure*}

\paragraph{Efficient Gaussian Splatting.}
Most Gaussian splatting methods for dynamic scenes initialize from COLMAP~\cite{colmap} point clouds, which takes several minutes per scene, whereas feed-forward Gaussian splatting (FFGS) predicts Gaussians from posed images in a single forward pass.
Among them, pixel-aligned methods~\cite{pixelsplat,mvsplat,depthsplat,gps-gaussian,lgm,gs-lrm} predict one Gaussian per pixel, so the number of Gaussians depends on the number and resolution of input views; recent methods decouple the Gaussians from pixels~\cite{tokengs,tokensplat,offthegrid}.
MVSGaussian~\cite{mvsgaussian} initializes Gaussians with a pixel-aligned FFGS and removes those whose depths are inconsistent across views.
With dense cameras, it produces too many Gaussians for efficient training and rendering, whereas with sparse cameras, it removes too many of them (\cref{sec:supp_ffinit}).
DSD-GS predicts Gaussians from a few views with a pixel-aligned FFGS, whose alignment lets each pixel's motion label its Gaussian, and corrects their depths along their pixel rays with the other training views instead of removing them.

To accelerate rendering, existing methods reduce the number of Gaussians.
For static scenes, Mini-Splatting~\cite{mini-splatting} and GaussianSpa~\cite{gaussianspa} prune or simplify Gaussians, and for dynamic scenes, 4DGS-1K~\cite{4dgs1k} removes Gaussians with short lifespans and skips inactive Gaussians at each frame.
However, the rendering time depends more on the number of screen tiles each Gaussian covers and its visible duration than on the number of Gaussians (\cref{sec:supp_cost}).
DSD-GS therefore regularizes this rendering cost during training, without additional pruning.

\section{Method}
\label{sec:method}

\subsection{Overview}
\label{sec:overview}
Given synchronized videos from a fixed multi-view camera rig, DSD-GS represents the scene with a static Gaussian set $\mathcal{G}_s$ and a dynamic Gaussian set $\mathcal{G}_d$ whose attributes are functions of time.
Our key idea is to \ymyi{separate the two sets early}, freeze the static Gaussians, and cache their rendering for each camera, so that \ymyi{repeated} training and rendering computation \ymyi{involves} only the dynamic set.
As shown in \cref{fig:overview}, the pipeline \ymyi{consists of} three stages.

\ymyi{
In the snapshot stage (\cref{sec:snapshot}), we initialize Gaussians from the first frame and assign initial static/dynamic labels using the optical flow at each Gaussian's corresponding pixel, given by projection for COLMAP initialization or by the pixel alignment of feed-forward Gaussian splatting (FFGS).
For FFGS initialization, Multi-View Ray Relocation corrects inaccurate depth estimates. We then train all Gaussians on the first frame and freeze the static set.
}

Since the first frame alone neither covers later content nor labels every Gaussian correctly, the dynamic expansion stage (\cref{sec:expansion}) %
\ymyi{addresses both limitations before full-sequence training.} 
It propagates the dynamic Gaussians along flow trajectories \ymyi{to keyframes}, seeds \ymyi{new} Gaussians where content emerges, and corrects the labels in both directions.

\ymyi{Finally}, the dynamic training stage (\cref{sec:training}) learns the temporal functions of the dynamic Gaussians over all frames. 
\ymyi{
Static-Caching Rasterization composites them over the cached static rendering, avoiding repeated static rasterization and preventing frozen static Gaussians from attenuating their gradients.
}
\ymyi{As} the dynamic Gaussians then dominate the remaining cost, Temporal Culling and \ymyi{Render-Cost Regularization further} reduce the rendering cost. %

\subsection{Snapshot}
\label{sec:snapshot}
\paragraph{Pixel-wise Dynamic--Static Decomposition.}
Given the first-frame images $\mathcal{I}_0=\{I_0^v\}_{v\in\mathcal{V}_{\text{in}}}$ of a sparse set of input views, a pixel-aligned feed-forward Gaussian splatting encoder $\mathcal{E}$~\cite{depthsplat} predicts one Gaussian per pixel in a single forward pass:
\begin{equation}
\mathcal{G}_{\text{init}} = \mathcal{E}(\mathcal{I}_0) = \{ g_{u,v} \mid (u,v)\in\Omega \},
\label{eq:ffgs}
\end{equation}
where $\Omega$ is the pixel domain and $g_{u,v}=\{\boldsymbol{\mu}_{u,v}, \mathbf{s}_{u,v}, \mathbf{r}_{u,v}, \mathbf{c}_{u,v}, \sigma_{u,v}\}$ holds the position, scale, rotation, color and opacity of the Gaussian at pixel $(u,v)$.
Because each Gaussian corresponds to a pixel, a per-pixel motion label directly becomes a per-Gaussian label.
We compute optical flow $F_{0\rightarrow \ell\Delta t}$ from the first frame to frames sampled at a fixed interval $\Delta t$, $\ell\in\{1,\dots,L\}$, and measure the motion of each pixel $(u,v)$ by the flow magnitude $M^\ell_{u,v}=\lVert F_{0\rightarrow \ell\Delta t}(u,v)\rVert_2$.
A Gaussian is labeled dynamic if its pixel moves in at least one sampled frame:
\begin{equation}
g_{u,v} \in
\begin{cases}
\mathcal{G}_s, & \text{if } M^\ell_{u,v} < \tau_M \;\; \forall \ell,\\
\mathcal{G}_d, & \text{otherwise},
\end{cases}
\label{eq:flowlabel}
\end{equation}
where $\tau_M$ is a threshold defined relative to the image width.

\paragraph{Multi-View Ray Relocation.}
The depth predicted by the feed-forward encoder $\mathcal{E}$ is inaccurate on camera rigs that differ from its training data, and the resulting Gaussians float in front of or behind the true surface.
Inspired by plane-sweep stereo~\cite{collins1996,gallup2007}, we verify the depth of each Gaussian along the pixel ray of its source camera by photometric consistency across the training views.
Unlike multi-view stereo, which estimates dense depth from scratch and leaves holes where matching fails, we search only around the depth predicted by $\mathcal{E}$ and keep the Gaussian there when matching fails.
For each candidate depth $d_m=\beta_m\hat{d}_{u,v}$ along the ray, we project the $5\times5$ source patch around $(u,v)$ placed at $d_m$ into the other training views and, following~\cite{kang2001}, average the matching cost over the better half $\mathcal{V}_m$ of the visible views to be robust to occlusion,
\begin{equation}
C(d_m) = \frac{1}{|\mathcal{V}_{m}|}\sum_{j\in\mathcal{V}_m} \frac{1-\rho_j(d_m)}{2},
\label{eq:reloc}
\end{equation}
where $\rho_j(d_m)$ is the zero-mean normalized cross-correlation (ZNCC) between the source patch and the patch sampled in view $j$ at the projection of $\mathbf{o}+d_m\mathbf{v}_{u,v}$, $\mathbf{o}$ is the source camera center, and $\mathbf{v}_{u,v}$ is the ray direction scaled to unit depth.
When the minimum $d^\ast=\beta^\ast\hat{d}_{u,v}$ of $C$ is distinct and changes the depth by more than a margin, the Gaussian is moved to $\boldsymbol{\mu}_{u,v}=\mathbf{o}+d^\ast\mathbf{v}_{u,v}$ and its scale is multiplied by $\beta^\ast$.
The correction takes a few seconds per scene (\cref{sec:supp_reloc}).

\paragraph{Snapshot Training.}
We optimize all Gaussians as a static 3DGS model on the first-frame images with
\begin{equation}
\mathcal{L}_{\text{photo}} = (1-\lambda)\,\mathcal{L}_1 + \lambda\,\mathcal{L}_{\text{D-SSIM}}.
\label{eq:photo}
\end{equation}
After this stage $\mathcal{G}_s$ is frozen and no longer optimized.

\subsection{Dynamic Expansion}
\label{sec:expansion}
\paragraph{Flow Trajectories.}
Dynamic expansion relies on the trajectories of the first-frame pixels.
For each input view $v\in\mathcal{V}_{\text{in}}$, we track every pixel $\mathbf{p}$ of the first frame by chaining optical flow at an interval of $\Delta\tau$ frames,
\begin{equation}
\mathbf{p}_v(t+\Delta\tau) = \mathbf{p}_v(t) + F^v_{t\rightarrow t+\Delta\tau}\big(\mathbf{p}_v(t)\big), \quad \mathbf{p}_v(0)=\mathbf{p},
\label{eq:track}
\end{equation}
and interpolate the positions between the tracked frames.

\paragraph{Keyframe Propagation along Flow Trajectories.}
Since the dynamic Gaussians from the snapshot alone hardly learn the motion far from the first frame, Keyframe Propagation duplicates the dynamic set at keyframes $t_k$ so that each copy learns the motion around its keyframe.
Rather than simply duplicating the Gaussians, we initialize the position and velocity of each copy from the flow trajectories and its temporal center as $\mu^\tau_i=t_k$.
Since the trajectories give only 2D positions, we unproject the trajectory position of each Gaussian at $t_k$ with its first-frame depth in each input view and average the resulting points over the views (\cref{sec:supp_traj}).

\paragraph{Emergent-Content Seeding.}
Content that appears after the first frame, such as background revealed behind a moving person or an object entering the scene, has no Gaussian responsible for it.
We detect it at each keyframe as holes, the pixels that no flow trajectory of the first-frame pixels reaches.
Only for keyframes whose holes exceed a small fraction of the image, we run $\mathcal{E}$ on the keyframe images and seed Gaussians at the hole pixels, which are corrected by Multi-View Ray Relocation and added to the copy of that keyframe as dynamic Gaussians with zero velocity.

\paragraph{Bidirectional Label Correction.}
Since \cref{eq:flowlabel} labels each Gaussian by the single-view flow magnitude $M^\ell_{u,v}$, the labels may contain errors in both directions.
Dynamic-to-static (D$\rightarrow$S) correction projects each dynamic Gaussian into all input views where it is visible and checks the flow trajectories of the projected pixels.
If it barely moves in all of them, we regard its motion as unnecessary to learn and relabel it static before propagation.
Static-to-dynamic (S$\rightarrow$D) correction instead uses conflicts of position gradients.
After propagation and seeding, a short keyframe training fine-tunes the newly added dynamic Gaussians.
After keyframe training, we compute the position gradient $\mathbf{g}_{i,k,c}=\partial\mathcal{L}_{k,c}/\partial\boldsymbol{\mu}_i$ of each static Gaussian $i$ under the loss of camera $c$ at keyframe $k$, average it over the cameras $\mathcal{C}_k$ of the keyframe into $\mathbf{g}_{i,k}$, and measure how inconsistent these gradients are across keyframes, since a static Gaussian that actually moves is pulled in different directions by different keyframes,
\begin{equation}
\eta_i = \frac{\sqrt{\sum_k \lVert \mathbf{g}_{i,k}-\bar{\mathbf{g}}_i \rVert^2}}{\sqrt{\sum_k \frac{1}{|\mathcal{C}_k|}\sum_{c\in\mathcal{C}_k} \lVert \mathbf{g}_{i,k,c} \rVert^2}},
\label{eq:gradswitch}
\end{equation}
where $\bar{\mathbf{g}}_i$ is the mean over keyframes.
The denominator normalizes by the magnitude of the per-camera gradients, which is large at edges where the cameras disagree regardless of motion.
Static Gaussians with $\eta_i>\lambda_\eta$ are relabeled dynamic and added to the copies of all keyframes.

\subsection{Dynamic Training}
\label{sec:training}
\paragraph{Temporal Function and Temporal Culling.} 
Following STG~\cite{stg}, each dynamic Gaussian $i$ has a temporal opacity $\sigma_i(t)$, a polynomial position $\boldsymbol{\mu}_i(t)$ and a polynomial rotation $\mathbf{r}_i(t)$:
\begin{align}
\sigma_i(t) &= \sigma_i \exp\!\big(-s^\tau_i \lvert t-\mu^\tau_i \rvert^2\big), \label{eq:topacity}\\
\boldsymbol{\mu}_i(t) &= \textstyle\sum_{n=0}^{n_\mu} \mathbf{m}_{i,n}(t-\mu^\tau_i)^n, \label{eq:tmotion}\\
\mathbf{r}_i(t) &= \textstyle\sum_{n=0}^{n_r} \mathbf{q}_{i,n}(t-\mu^\tau_i)^n, \label{eq:trot}
\end{align}
where $\sigma_i$ is the peak opacity and $\mu^\tau_i$ and $s^\tau_i$ are the temporal center and scale.
Keyframe Propagation multiplies the number of dynamic Gaussians, but each propagated Gaussian contributes only near its keyframe.
At time $t$ we therefore skip every Gaussian with $\sigma_i(t)<1/255$, which cannot change any pixel, in both training and rendering, so that the per-frame cost does not grow with the sequence length.

\paragraph{Static-Caching Rasterization.}
Our snapshot dynamic--static decomposition enables efficient Static-Caching Rasterization.
Because $\mathcal{G}_s$ is frozen and the cameras are fixed, its rendering from each camera never changes.
We render it once per camera and cache the image $C_{\text{static}}$.
At each iteration we rasterize only the dynamic Gaussians that pass Temporal Culling, obtaining their color and remaining transmittance
\begin{align}
C_{\text{dyn}} &= \sum_{i\in\mathcal{G}_d} \mathbf{c}_i\,\alpha_i(t) \prod_{j<i}\big(1-\alpha_j(t)\big), \label{eq:dyn}\\
T_{\text{dyn}} &= \prod_{i\in\mathcal{G}_d}\big(1-\alpha_i(t)\big), \label{eq:tdyn}
\end{align}
where $\alpha_i(t)$ is the opacity of Gaussian $i$ at the pixel, given by $\sigma_i(t)$ and its projected 2D Gaussian, and composite them over the cache as
\begin{equation}
C = C_{\text{dyn}} + T_{\text{dyn}}\, C_{\text{static}}.
\label{eq:scr}
\end{equation}
The per-iteration and per-frame cost thus depends only on the dynamic Gaussians.
The same compositing procedure is used during training and rendering, ensuring that the dynamic Gaussians are optimized consistently with the final rendering process.
Applying the cache only at rendering time would composite the Gaussians in a depth order different from that in training, which degrades quality (\cref{sec:supp_scr_renderonly}).
Because the dynamic Gaussians are always composited in front of the cache, frozen static Gaussians cannot occlude them during training, which also improves the reconstruction of moving content (\cref{sec:supp_scr_optim}).
This composition still correctly renders most occlusions caused by static foreground objects, and an optional depth test handles view-dependent ones, as needed in 360$^\circ$ capture (\cref{sec:supp_static_occluder}).

\paragraph{Render-Cost Regularization.} 
The cost of rendering a dynamic Gaussian depends on the number of screen tiles it covers and the number of frames in which it is visible.
Tile-based rasterization duplicates, sorts, and blends one entry per Gaussian--tile pair, so the frame time is dominated by the number of pairs ($R^2=0.996$, \cref{sec:supp_cost}).
With Temporal Culling, a Gaussian enters this process only in the frames in which it is visible.
For a Gaussian with maximum scale $\bar{s}_i=\max(\mathbf{s}_i)$ at depth $d_i$ under focal length $f$, we approximate the number of $16\times16$ tiles it covers by
\begin{equation}
A_i = \frac{\pi\,(3 f \bar{s}_i / d_i)^2}{256},
\label{eq:area}
\end{equation}
and, from \cref{eq:topacity}, the duration over which its opacity exceeds the culling threshold $1/255$ by
\begin{equation}
T_i = 2\sqrt{\max\!\big(0,\,\ln(255\,\mathrm{sg}(\sigma_i))\big)/s^\tau_i},
\label{eq:visibletime}
\end{equation}
where $\mathrm{sg}(\cdot)$ stops the gradient, so the regularizer shortens the visible duration through $s^\tau_i$ without lowering the opacity.
We add the total cost to the training loss as
\begin{equation}
\mathcal{L} = \mathcal{L}_{\text{photo}} + \frac{\lambda_c}{C_0}\sum_{i\in\mathcal{G}_d} A_i T_i,
\label{eq:costloss}
\end{equation}
where $C_0$ is the value of $\sum_i A_i T_i$ at the start of regularization, which normalizes the cost across scenes so that a single $\lambda_c$ applies to all scenes.
This makes the dynamic Gaussians smaller and shorter-lived without additional pruning, which speeds up rendering with little loss in quality.

\section{Experiments}
\label{sec:experiments}

\subsection{Setup}
\label{sec:setup}
\paragraph{Benchmarks.}
We evaluate our method on two multi-view dynamic scene benchmarks captured with fixed camera rigs.
Neural 3D~\cite{dynerf} contains six scenes of 300 frames, and following prior work, we hold out camera 00 for testing.
Technicolor~\cite{technicolor} contains five scenes with objects that enter after the first frame and severe disocclusion, which we use with the full sequences, holding out camera 10 for testing.
More details are given in \cref{sec:supp_bench}.

\paragraph{Baselines.}
We compare with STG~\cite{stg}, Taylor~\cite{taylor}, 4DGS~\cite{4dgs}, Swift4D~\cite{swift4d}, DeGauss~\cite{degauss}, and RT-4DGS~\cite{rt-4dgs}, among which Swift4D and DeGauss perform dynamic--static decomposition.
Each baseline follows its original COLMAP initialization, denoted D0 for dense COLMAP on the first frame and SA for sparse COLMAP on all frames in \cref{tab:main_dynerf,tab:main_technicolor}.
FF denotes our feed-forward initialization on the first frame without Multi-View Ray Relocation (MVR), whose results with relocation are given in \cref{sec:supp_ffinit}.
We also report a variant of our method initialized with D0, in which Emergent-Content Seeding uses sparse COLMAP instead of feed-forward Gaussian splatting.

\paragraph{Metrics.}
We measure reconstruction quality with PSNR, SSIM~\cite{ssim}, and LPIPS~\cite{lpips} with AlexNet~\cite{alexnet}, averaged over all frames of the test camera.
For efficiency, we report the initialization time, the training time, and their sum in minutes, and the rendering FPS.
All methods are trained and evaluated on an RTX 5090 GPU, and the FPS is additionally measured on a Jetson AGX Orin.
Hyperparameters, baseline settings, and the timing protocol are in \cref{sec:supp_impl}.

\subsection{Comparison}
\begin{table}[t]
\centering
\caption{Quantitative comparison on Neural 3D~\cite{dynerf}. Init.: initialization (\cref{sec:setup}). Times are in minutes. The \rankbox{rankfirst}{best}, \rankbox{ranksecond}{second-best}, and \rankbox{rankthird}{third-best} results are highlighted.}
\label{tab:main_dynerf}
\scriptsize
\setlength{\tabcolsep}{1.6pt}
\begin{adjustbox}{max width=\linewidth}
\begin{tabular}{lc ccc ccc cc}
\toprule
Method & Init. & PSNR$\uparrow$ & SSIM$\uparrow$ & LPIPS$\downarrow$ & \makecell{Init.\\time$\downarrow$} & \makecell{Train\\time$\downarrow$} & \makecell{Total\\time$\downarrow$} & \makecell{FPS$\uparrow$\\(5090)} & \makecell{FPS$\uparrow$\\(Orin)} \\
\midrule
STG~\cite{stg} & SA & \third{30.70} & \best{0.940} & 0.057 & 15.9 & 27.2 & 43.1 & \third{483.7} & \third{35.6} \\
Taylor~\cite{taylor} & SA & 29.53 & 0.927 & 0.078 & 15.9 & 81.8 & 97.7 & 62.5 & 4.9 \\
4DGS~\cite{4dgs} & D0 & 30.09 & 0.929 & 0.065 & 5.9 & 32.1 & 38.0 & 172.1 & 18.0 \\
Swift4D~\cite{swift4d} & D0 & 29.74 & 0.932 & 0.055 & 11.1 & 26.6 & 37.7 & 155.1 & 16.1 \\
DeGauss~\cite{degauss} & D0 & 29.96 & 0.935 & 0.060 & 5.9 & 124.7 & 130.6 & 96.3 & 14.9 \\
RT-4DGS~\cite{rt-4dgs} & D0 & 29.56 & 0.935 & 0.059 & 5.9 & 240.9 & 246.8 & 97.4 & 11.3 \\
\midrule
STG~\cite{stg} & FF & 29.90 & 0.933 & 0.069 & 0.1 & \third{24.4} & \third{24.5} & 483.2 & 32.2 \\
Taylor~\cite{taylor} & FF & 29.85 & 0.926 & 0.076 & 0.1 & 71.3 & 71.4 & 66.7 & 5.8 \\
4DGS~\cite{4dgs} & FF & 30.28 & 0.926 & 0.072 & 0.1 & 33.1 & 33.2 & 153.0 & 15.8 \\
Swift4D~\cite{swift4d} & FF & 30.30 & \third{0.937} & \third{0.051} & 5.2 & 26.5 & 31.7 & 147.7 & 14.4 \\
DeGauss~\cite{degauss} & FF & 30.48 & \third{0.937} & 0.056 & 0.1 & 121.4 & 121.5 & 95.4 & 14.8 \\
\midrule
Ours & D0 & \best{30.91} & \second{0.939} & \second{0.050} & 6.7 & \best{9.3} & \second{16.0} & \best{2025.0} & \best{307.4} \\
Ours & FF+MVR & \second{30.83} & 0.932 & \best{0.045} & 0.7 & \second{9.6} & \best{10.3} & \second{1926.2} & \second{280.0} \\
\bottomrule
\end{tabular}
\end{adjustbox}
\end{table}

\begin{table}[t]
\centering
\caption{Quantitative comparison on Technicolor~\cite{technicolor}. Init.: initialization (\cref{sec:setup}). Times are in minutes. The \rankbox{rankfirst}{best}, \rankbox{ranksecond}{second-best}, and \rankbox{rankthird}{third-best} results are highlighted.}
\label{tab:main_technicolor}
\scriptsize
\setlength{\tabcolsep}{1.6pt}
\begin{adjustbox}{max width=\linewidth}
\begin{tabular}{lc ccc ccc cc}
\toprule
Method & Init. & PSNR$\uparrow$ & SSIM$\uparrow$ & LPIPS$\downarrow$ & \makecell{Init.\\time$\downarrow$} & \makecell{Train\\time$\downarrow$} & \makecell{Total\\time$\downarrow$} & \makecell{FPS$\uparrow$\\(5090)} & \makecell{FPS$\uparrow$\\(Orin)} \\
\midrule
STG~\cite{stg} & SA & \second{31.83} & \second{0.906} & \second{0.104} & 20.0 & 60.4 & 80.4 & 205.8 & 12.9 \\
Taylor~\cite{taylor} & SA & 31.33 & 0.893 & \third{0.128} & 20.0 & 218.5 & 238.5 & 50.0 & 3.5 \\
4DGS~\cite{4dgs} & D0 & 27.96 & 0.822 & 0.208 & 5.3 & 65.1 & 70.4 & 107.4 & 10.5 \\
Swift4D~\cite{swift4d} & D0 & 29.82 & 0.889 & 0.132 & 10.6 & \third{39.4} & 50.0 & 114.7 & 11.3 \\
DeGauss~\cite{degauss} & D0 & 23.14 & 0.837 & 0.180 & 5.3 & 178.3 & 183.6 & 92.7 & \third{15.7} \\
RT-4DGS~\cite{rt-4dgs} & D0 & 30.15 & 0.859 & 0.171 & 5.3 & 272.9 & 278.2 & 102.5 & 9.4 \\
\midrule
STG~\cite{stg} & FF & 23.01 & 0.702 & 0.297 & 0.1 & 49.8 & 49.9 & \third{237.7} & 15.5 \\
Taylor~\cite{taylor} & FF & 27.13 & 0.823 & 0.201 & 0.1 & 127.5 & 127.6 & 69.5 & 5.2 \\
4DGS~\cite{4dgs} & FF & 26.73 & 0.796 & 0.233 & 0.1 & 68.8 & 68.9 & 91.1 & 8.4 \\
Swift4D~\cite{swift4d} & FF & 27.72 & 0.851 & 0.154 & 5.4 & 42.1 & \third{47.5} & 94.6 & 8.4 \\
DeGauss~\cite{degauss} & FF & 20.21 & 0.737 & 0.271 & 0.1 & 176.2 & 176.3 & 95.9 & 15.2 \\
\midrule
Ours & D0 & \best{32.14} & \best{0.909} & \best{0.093} & 6.2 & \best{24.4} & \second{30.6} & \second{992.2} & \second{91.3} \\
Ours & FF+MVR & \third{31.68} & \third{0.900} & \second{0.104} & 0.6 & \second{26.0} & \best{26.6} & \best{1032.8} & \best{105.6} \\
\bottomrule
\end{tabular}
\end{adjustbox}
\end{table}

\begin{figure*}[t]
\centering
\includegraphics[width=\linewidth]{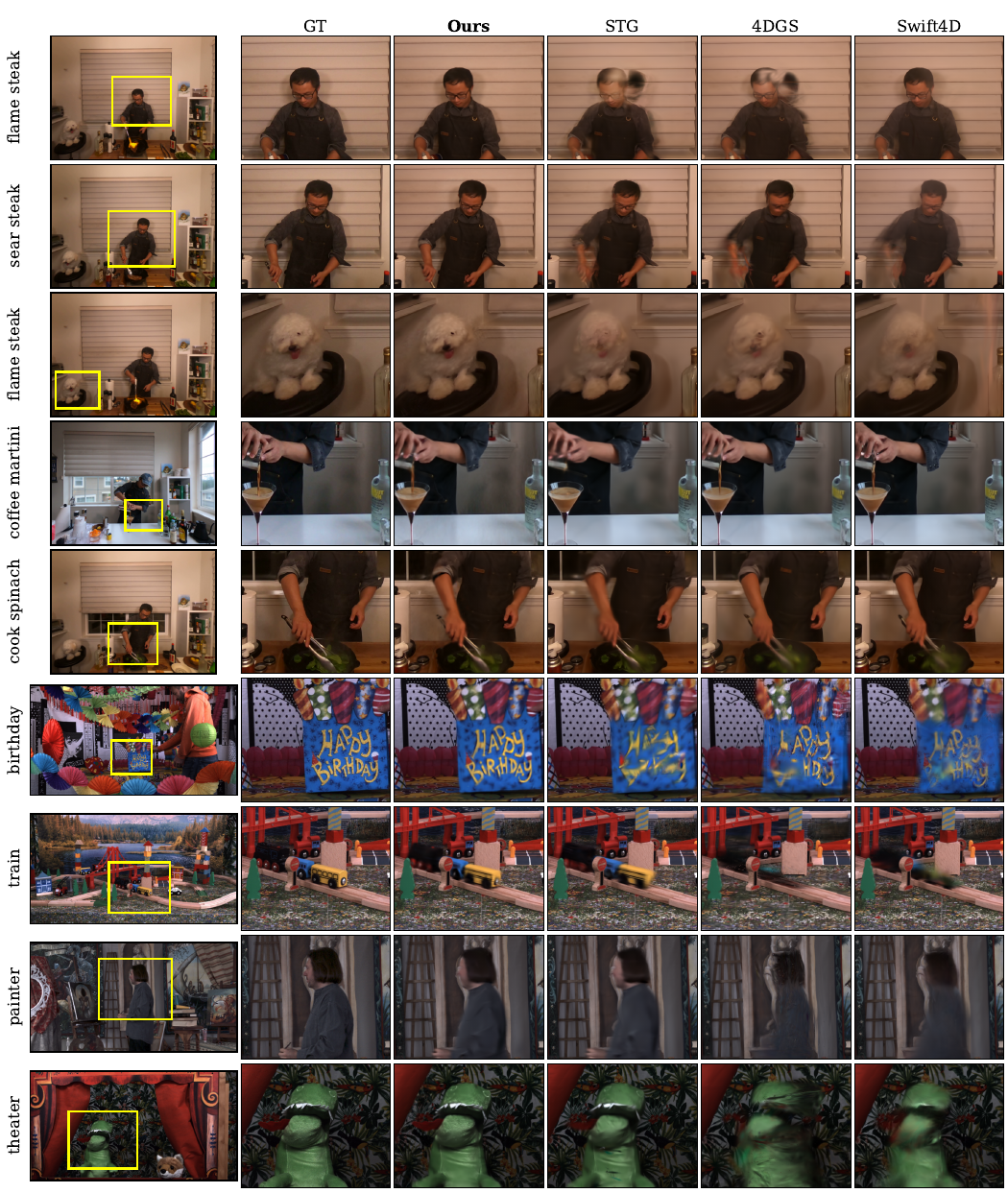}
\caption{Qualitative comparison with STG~\cite{stg}, 4DGS~\cite{4dgs}, and Swift4D~\cite{swift4d} on Neural 3D~\cite{dynerf} (top five rows) and Technicolor~\cite{technicolor} (bottom four rows).}
\label{fig:qual}
\end{figure*}
\paragraph{Quantitative comparison.}
\Cref{tab:main_dynerf,tab:main_technicolor} compare DSD-GS with the baselines.
On Neural 3D, DSD-GS initialized with FF+MVR achieves higher PSNR and lower LPIPS than all baselines, while reducing the total time from 43.1 to 10.3 minutes compared with STG and rendering at 1926 FPS on an RTX 5090 GPU and 280 FPS on a Jetson AGX Orin, about 4$\times$ and 8$\times$ faster than STG.
On Technicolor, it achieves PSNR and LPIPS comparable to those of STG, while its total time is 3.0$\times$ shorter and its rendering is 5.0$\times$ and 8.2$\times$ faster on the two devices.
With D0 initialization, DSD-GS achieves the best PSNR on both benchmarks.
The feed-forward model predicts accurate geometry on Neural 3D, whose camera configuration resembles its training data, but inaccurate geometry on Technicolor, whose camera configuration lies outside it.
On Technicolor, the PSNR of the baselines with FF thus drops by 1.2--8.8\,dB, and relocation reduces this drop to 0.3--2.9\,dB (\cref{sec:supp_ffinit}).

\paragraph{Qualitative comparison.}
As shown in \cref{fig:qual}, the baselines exhibit ghosting artifacts and distorted moving objects.
In \emph{flame steak}, STG and 4DGS show ghosting of the head.
In \emph{coffee martini}, the baselines render duplicated or smeared fingers.
In the other scenes, the moving objects of the baselines are blurred or lose their shape.
DSD-GS shows fewer of these artifacts.

\paragraph{Temporally varying regions.}
Since image-level metrics are dominated by static regions, we also evaluate the temporally varying regions, in which the ground-truth color deviates from its temporal median.
In these regions, DSD-GS ranks first in all three metrics on both benchmarks and, with FF+MVR initialization, improves the PSNR over STG by 2.35\,dB on Neural 3D and 0.50\,dB on Technicolor (\cref{sec:supp_region}).

\subsection{Ablation}
\begin{table}[t]
\centering
\caption{Ablation study on Technicolor~\cite{technicolor}. Each row removes one component from the full model. $^\dagger$ and $^\ddagger$ mark components for quality and efficiency, respectively. Train time is in minutes.}
\label{tab:ablation}
\small
\setlength{\tabcolsep}{4pt}
\begin{adjustbox}{max width=\linewidth}
\begin{tabular}{l cc cc}
\toprule
Variant & PSNR$\uparrow$ & LPIPS$\downarrow$ & \makecell{Train\\time$\downarrow$} & FPS$\uparrow$ \\
\midrule
Full model & 31.68 & 0.104 & 26.0 & 1033 \\
\midrule
\multicolumn{5}{l}{\textit{Snapshot}} \\
\quad w/o Dynamic--Static Decomposition$^\ddagger$ & 31.79 & 0.097 & \textbf{61.6} & \textbf{316} \\
\quad w/o Multi-View Ray Relocation$^\dagger$ & \textbf{26.85} & \textbf{0.163} & 25.2 & 1126 \\
\midrule
\multicolumn{5}{l}{\textit{Dynamic Expansion}} \\
\quad w/o Keyframe Propagation$^\dagger$ & \textbf{29.52} & \textbf{0.128} & 17.5 & 1603 \\
\quad w/o Emergent-Content Seeding$^\dagger$ & \textbf{31.08} & \textbf{0.111} & 25.6 & 1149 \\
\multicolumn{5}{l}{Bidirectional Label Correction} \\
\quad w/o D$\rightarrow$S Correction$^\ddagger$ & 31.67 & 0.102 & \textbf{29.2} & \textbf{1018} \\
\quad w/o S$\rightarrow$D Correction$^\dagger$ & \textbf{31.24} & \textbf{0.107} & 21.7 & 1503 \\
\midrule
\multicolumn{5}{l}{\textit{Dynamic Training}} \\
\quad w/o Temporal Culling$^\ddagger$ & 31.49 & 0.106 & \textbf{33.4} & \textbf{401} \\
\quad w/o Static-Caching Rasterization$^{\dagger\ddagger}$ & \textbf{30.90} & \textbf{0.105} & \textbf{44.4} & \textbf{193} \\
\quad w/o Render-Cost Regularization$^\ddagger$ & 31.62 & 0.103 & \textbf{31.8} & \textbf{425} \\
\bottomrule
\end{tabular}
\end{adjustbox}
\end{table}

\Cref{tab:ablation} removes each component from the full model on Technicolor.
Without the dynamic--static decomposition, where all Gaussians are dynamic\hyt{, which also removes Static-Caching Rasterization and label correction},
the PSNR is 0.11\,dB higher, but the training takes 2.4$\times$ longer, and the rendering is 3.3$\times$ slower.
Among the components for quality, removing Multi-View Ray Relocation causes the largest drop of 4.83\,dB, which shows that the feed-forward depth cannot be used as is.
Without Keyframe Propagation, the PSNR drops by 2.16\,dB, as the Gaussians from the first frame cannot follow the motion far from the snapshot (\cref{sec:supp_perframe}).
Without Emergent-Content Seeding, the PSNR drops by 0.60\,dB, mainly in the frames where a new person is in view (\cref{sec:supp_perframe}). %
\ymyi{Without} S$\rightarrow$D correction, \ymyi{it drops} by 0.44\,dB.
The components for efficiency barely change the quality but reduce the cost.
Without Temporal Culling or Render-Cost Regularization, the rendering speed falls from 1033 to 401 and 425 FPS, and without D$\rightarrow$S correction the training takes 3.2 minutes longer.
Static-Caching Rasterization serves both goals, as removing it lowers the rendering speed to 193 FPS and the PSNR by 0.78\,dB (\cref{sec:supp_scr_optim}).

\section{Limitations}
\label{sec:limitations}
DSD-GS has two limitations.
\ymyi{First, the rendering speedup from Static-Caching Rasterization relies on fixed rendering viewpoints, as commonly used in multi-view dynamic scene benchmarks~\cite{dynerf,technicolor}. In free-viewpoint playback, static Gaussians must be rasterized for each new viewpoint. However, training remains accelerated because the capture cameras are fixed, as is common in multi-view dynamic capture~\cite{dynerf,technicolor,broxton2020immersive,joo2015panoptic,isik2023humanrf,cheng2023dnarendering}.}
With Render-Cost Regularization also applied to the static Gaussians, DSD-GS \ymyi{maintains} the quality close to that of the full model and still renders faster than all baselines on both benchmarks on an RTX 5090 and an Orin, reaching over 30\,FPS on average on the Orin with a larger weight $\lambda_c$ (\cref{sec:supp_freecam}).
Second, the feed-forward initialization is less accurate \ymyi{for camera configurations outside its training distribution.}
Multi-View Ray Relocation corrects \ymyi{most errors}, but on Technicolor the FF variant remains 0.46\,dB below the D0 variant.

\section{Conclusion}
\label{sec:conclusion}
We present DSD-GS, \hyt{an efficient dynamic scene reconstruction method built on snapshot dynamic--static decomposition}.
\hyt{
\ymyi{It learns geometry before} motion, freezing} the static scene before dynamic optimization and \hyt{caching} its rendering, 
so that training \ymyi{involves} only the dynamic Gaussians. 
Dynamic expansion recovers emergent content and corrects labeling errors.
Render-Cost Regularization accelerates rendering by penalizing the screen area and visible duration of each dynamic Gaussian during training.
Multi-View Ray Relocation corrects \ymyi{inaccurate} depths of feed-forward initialization, \ymyi{enabling fast initialization with quality close to COLMAP.}
On Neural 3D and Technicolor, 
\ymyi{DSD-GS reduces total reconstruction time by 3--4$\times$ and accelerates rendering by 4--5$\times$ over STG} on a desktop GPU and \ymyi{renders} at over 100\,FPS \ymyi{even} on an embedded device, \ymyi{while} achieving \ymyi{overall} quality comparable to existing methods and the best quality in temporally varying regions.

\clearpage

{
    \small
    \bibliographystyle{ieeenat_fullname}
    \bibliography{main,references,supp}
}

\clearpage
\setcounter{page}{1}
\maketitlesupplementary
\appendix

This appendix summarizes the notation (\cref{sec:supp_notation}) and gives the details of the method (\cref{sec:supp_method}) and of the implementation (\cref{sec:supp_impl}).
\Cref{sec:supp_analysis} analyzes the components, the sensitivity to the hyperparameters, the render-cost trade-off, the feed-forward initialization, and the temporally varying regions.
\Cref{sec:supp_scr_discussion} discusses Static-Caching Rasterization, including its effect on optimization, free-viewpoint rendering, and static occluders.
\Cref{sec:supp_qual} shows additional qualitative comparisons, and \cref{sec:supp_perscene} gives the per-scene results.

\section{Notation}
\label{sec:supp_notation}
\Cref{tab:supp_notation} summarizes the notation used in \cref{sec:method} and in this appendix.

\begin{table}[!t]
\centering
\caption{Notation.}
\label{tab:supp_notation}
\scriptsize
\setlength{\tabcolsep}{2pt}
\renewcommand{\arraystretch}{1.0}
\begin{tabular}{@{}l p{0.66\linewidth}@{}}
\toprule
Symbol & Meaning \\
\midrule
\multicolumn{2}{@{}l}{\textit{Scene and Gaussians}} \\
$\mathcal{G}_s$, $\mathcal{G}_d$ & static and dynamic Gaussian sets \\
$\boldsymbol{\mu}$, $\mathbf{s}$, $\mathbf{r}$, $\mathbf{c}$, $\sigma$ & position, scale, rotation, color, and opacity of a Gaussian \\
\midrule
\multicolumn{2}{@{}l}{\textit{Snapshot (\cref{sec:snapshot})}} \\
$\mathcal{V}_{\text{in}}$, $I_0^v$, $\mathcal{I}_0$ & feed-forward input views, first-frame image of view $v$, and the set of these images \\
$\mathcal{E}$, $\mathcal{G}_{\text{init}}$ & pixel-aligned feed-forward encoder and its predicted Gaussians \\
$(u,v)$, $\Omega$, $g_{u,v}$ & pixel, pixel domain, and the Gaussian predicted at the pixel \\
$F_{0\rightarrow \ell\Delta t}$, $\Delta t$, $L$ & optical flow from the first frame to sampled frame $\ell$, sampling interval, and number of sampled frames \\
$M^\ell_{u,v}$, $\tau_M$ & flow magnitude at pixel $(u,v)$ and labeling threshold \\
$\hat{d}_{u,v}$, $d_m$, $\beta_m$ & predicted depth, candidate depth, and depth factor \\
$P_{\text{src}}$ & source patch around pixel $(u,v)$ \\
$\mathbf{o}$, $\mathbf{v}_{u,v}$ & source camera center and ray direction scaled to unit depth \\
$\rho_j$, $C(d_m)$, $\mathcal{V}_m$ & ZNCC in view $j$, matching cost of a candidate, and the better half of the visible views \\
$d^\ast$, $\beta^\ast$ & selected depth and its factor \\
$\mathcal{L}_{\text{photo}}$, $\lambda$ & photometric loss and its D-SSIM weight \\
\midrule
\multicolumn{2}{@{}l}{\textit{Dynamic expansion (\cref{sec:expansion})}} \\
$\mathbf{p}_v(t)$, $F^v_{t\rightarrow t+\Delta\tau}$, $\Delta\tau$ & tracked position of a first-frame pixel in view $v$, optical flow, and tracking interval \\
$t_k$, $K$, $\mathcal{G}_d^k$ & time of keyframe $k$, number of keyframes after the first frame, and dynamic copy at keyframe $k$ \\
$\mathcal{L}_{k,c}$, $\mathcal{C}_k$ & loss of camera $c$ at keyframe $k$ and the cameras of keyframe $k$ \\
$\mathbf{g}_{i,k,c}$, $\mathbf{g}_{i,k}$, $\bar{\mathbf{g}}_i$ & position gradient of static Gaussian $i$ for camera $c$ at keyframe $k$, its mean over $\mathcal{C}_k$, and its mean over keyframes \\
$\eta_i$, $\lambda_\eta$ & gradient inconsistency and S$\rightarrow$D threshold \\
\midrule
\multicolumn{2}{@{}l}{\textit{Dynamic training (\cref{sec:training})}} \\
$\sigma_i(t)$, $\boldsymbol{\mu}_i(t)$, $\mathbf{r}_i(t)$ & temporal opacity, position, and rotation of dynamic Gaussian $i$ \\
$\mu^\tau_i$, $s^\tau_i$ & temporal center and temporal scale \\
$\mathbf{m}_{i,n}$, $\mathbf{q}_{i,n}$, $n_\mu$, $n_r$ & polynomial coefficients and degrees of position and rotation \\
$\alpha_i(t)$ & opacity of Gaussian $i$ at a pixel at time $t$ \\
$C_{\text{static}}$, $C_{\text{dyn}}$, $T_{\text{dyn}}$, $C$ & cached static image, dynamic color, dynamic transmittance, and composited color \\
$\bar{s}_i$, $d_i$, $f$ & maximum scale, depth, and focal length \\
$A_i$, $T_i$ & number of covered tiles and number of visible frames \\
$\mathrm{sg}(\cdot)$ & stop-gradient \\
$\lambda_c$, $C_0$, $n_c$ & render-cost weight, total cost at the start of regularization, and its start iteration \\
\midrule
\multicolumn{2}{@{}l}{\textit{Appendix}} \\
$\mathcal{V}_{\text{tr}}$ & training cameras \\
$N_{\text{snap}}$, $N_{\text{kf}}$, $N_{\text{dyn}}$ & iterations of Snapshot Training, keyframe training, and dynamic training \\
$\Gamma_v(\mathbf{p})$, $\mathcal{V}_i(t_k)$ & trajectory of first-frame pixel $\mathbf{p}$ in view $v$, and input views used to place Gaussian $i$ at $t_k$ \\
$\mathcal{P}_v(t_k)$, $H_v(t_k)$ & valid trajectory positions and hole mask of view $v$ at $t_k$ \\
$\tau_D$ & D$\rightarrow$S motion threshold \\
$\mathcal{G}_d(t)$, $\mathcal{T}_i$, $r_i$ & dynamic Gaussians rasterized at time $t$, tiles overlapped by Gaussian $i$, and its screen-space radius \\
$P$, $N$ & number of Gaussian--tile pairs and number of visible dynamic Gaussians in a frame \\
$a$, $b$, $c$, $\kappa$ & coefficients of the frame-time model and per-element costs of the rasterizer stages \\
\bottomrule
\end{tabular}
\end{table}

\section{Method Details}
\label{sec:supp_method}

\subsection{Full Pipeline}
\label{sec:supp_algo}
\Cref{alg:pipeline} summarizes the full pipeline of DSD-GS, and \cref{alg:scr} details Static-Caching Rasterization.
Let $\mathcal{V}_{\text{tr}}$ be the training cameras, which exclude the test camera, and $\mathcal{V}_{\text{in}}\subset\mathcal{V}_{\text{tr}}$ the four input views of the feed-forward encoder.
The keyframes are $t_0=0<t_1<\dots<t_K$, and $\mathcal{G}_d^k$ denotes the copy of the dynamic Gaussians at keyframe $t_k$, where $\mathcal{G}_d^0$ is the dynamic part of the snapshot.
Snapshot Training updates $\mathcal{G}_s$ and $\mathcal{G}_d^0$ on the first frame, keyframe training updates only the copies $\mathcal{G}_d^1,\dots,\mathcal{G}_d^K$, and dynamic training updates only $\mathcal{G}_d$ while $\mathcal{G}_s$ stays frozen and cached.

\subsection{Multi-View Ray Relocation}
\label{sec:supp_reloc}
\begin{figure*}[h]
\centering
\includegraphics[width=\linewidth]{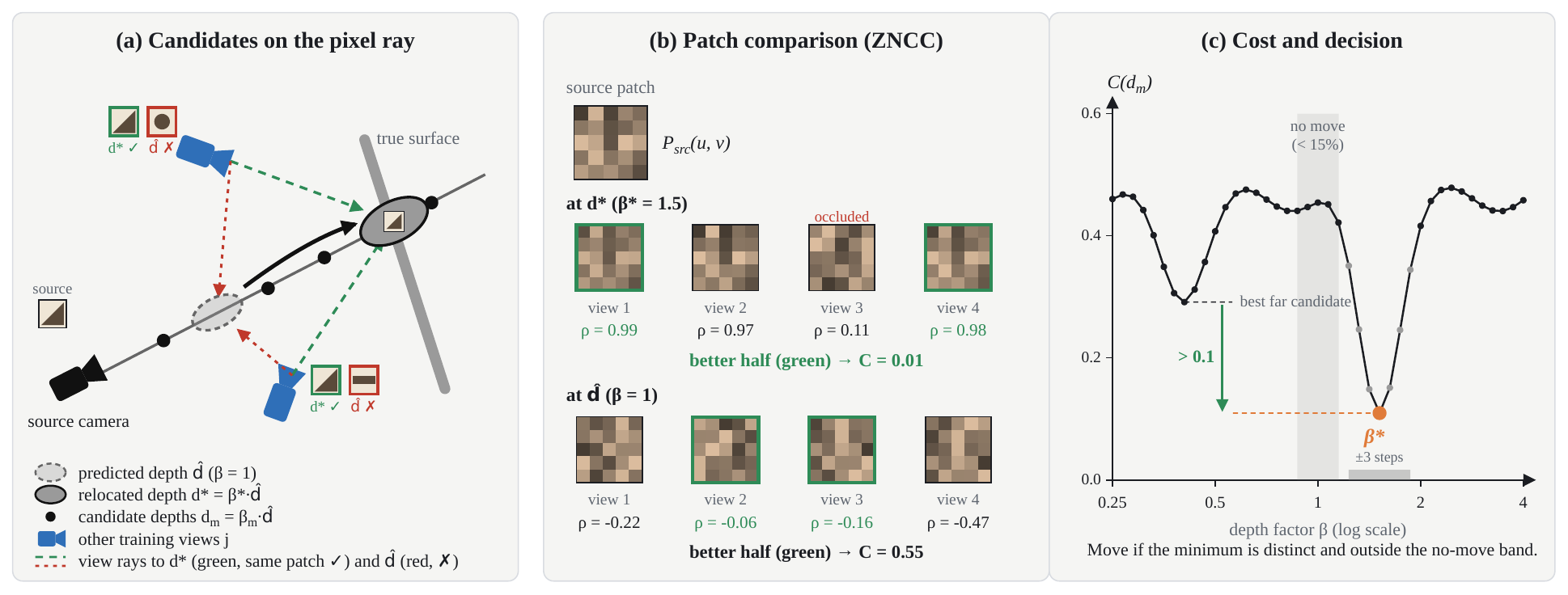}
\caption{Multi-View Ray Relocation. (a) Candidate depths on the pixel ray of the source camera. At the relocated depth $d^\ast$, the other views see the same patch as the source view, whereas at the predicted depth $\hat{d}$ they see different patches. (b) Patch comparison by ZNCC, averaging the better half of the visible views. (c) Cost over the depth factor and the move decision. Panels (b) and (c) are illustrative.}
\label{fig:supp_reloc}
\end{figure*}

As shown in \cref{fig:supp_reloc}(a), Multi-View Ray Relocation places each Gaussian at candidate depths on the ray from its source camera through its pixel.
It checks whether the other training views see the same image patch at each candidate and moves the Gaussian only when one candidate is clearly better than the predicted depth.

The candidate depths are $d_m=\beta_m\hat{d}_{u,v}$ with 41 factors spaced logarithmically in $[0.25, 4]$,
\begin{equation}
\beta_m = 0.25\cdot 16^{\,m/40}, \quad m=0,\dots,40.
\label{eq:supp_beta}
\end{equation}
Matching is performed on images resized to a width of 512 pixels.
At each candidate, the $5\times5$ source patch around $(u,v)$ is placed at $d_m$ and projected into the other training views.
As shown in \cref{fig:supp_reloc}(b), the projected patch is compared with the source patch by ZNCC.
For the source patch $P$ and the patch $Q$ sampled in view $j$, both in RGB, $\rho_j$ in \cref{eq:reloc} is
\begin{equation}
\rho_j = \frac{\sum_{\mathbf{q},k}\tilde{P}_k(\mathbf{q})\,\tilde{Q}_k(\mathbf{q})}{\sqrt{\sum_{\mathbf{q},k}\tilde{P}_k(\mathbf{q})^2\,\sum_{\mathbf{q},k}\tilde{Q}_k(\mathbf{q})^2}},
\label{eq:supp_zncc}
\end{equation}
where $\mathbf{q}$ runs over the 25 patch pixels, $k$ over the color channels, and $\tilde{P}_k=P_k-\bar{P}_k$ and $\tilde{Q}_k=Q_k-\bar{Q}_k$ subtract the patch means $\bar{P}_k$ and $\bar{Q}_k$ of channel $k$.
Following~\cite{kang2001} for robustness to occlusion, the cost averages $(1-\rho_j)/2$ over the $|\mathcal{V}_m|=\max(2,\mathrm{round}(n/2))$ of the $n$ visible views with the highest $\rho_j$, and is set to 1 when fewer than two views are visible.
Let $m^\ast=\arg\min_m C(d_m)$.
As shown in \cref{fig:supp_reloc}(c), the minimum is distinct when it is lower by more than 0.1 than the cost of every candidate at least four steps away,
\begin{equation}
\min_{|m-m^\ast|\ge 4} C(d_m) - C(d_{m^\ast}) > 0.1,
\label{eq:supp_uniq}
\end{equation}
so that a flat or multi-modal cost leaves the Gaussian at its predicted depth.
If the distinct minimum lies within a margin of 15\% of the predicted depth, the prediction is regarded as correct and kept, so a Gaussian is moved only when $\lvert\ln\beta^\ast\rvert>\ln 1.15$.
A moved Gaussian is placed at $d_{m^\ast}$ along its pixel ray, and its scale is multiplied by $\beta^\ast$, which keeps its projected size in the source view.

On an RTX 5090 GPU, relocation takes about 0.8\,s per scene on Technicolor and about 5\,s on Neural 3D.

\paragraph{Relation to multi-view stereo.}
The matching follows plane-sweep stereo~\cite{collins1996,gallup2007} with the view selection of~\cite{kang2001}, but its role differs from multi-view stereo.
Multi-view stereo estimates a new depth map for every view over the full depth range and fuses the maps into a new point cloud, so it leaves holes where matching fails, such as in textureless or occluded regions.
Relocation instead only verifies the depth predicted for each feed-forward Gaussian.
Where matching fails and the minimum is not distinct, it keeps the Gaussian at its predicted depth, so every Gaussian is retained and no holes appear (\cref{sec:supp_ffinit}).
A moved Gaussian keeps its other attributes and moves only along its own pixel ray.

\begin{algorithm}[t]
\caption{Full pipeline of DSD-GS.}
\label{alg:pipeline}
\begin{algorithmic}[1]
\Require Images $I_t^v$ ($v\in\mathcal{V}_{\text{tr}}$) with camera parameters, input views $\mathcal{V}_{\text{in}}$, keyframes $t_0<\dots<t_K$
\Ensure Static Gaussians $\mathcal{G}_s$, dynamic Gaussians $\mathcal{G}_d$, static caches $C_{\text{static}}^v$
\State \textbf{Stage 1: Snapshot} (\cref{sec:snapshot})
\State $\mathcal{G}_{\text{init}} \gets \mathcal{E}(\{I_0^v\}_{v\in\mathcal{V}_{\text{in}}})$ \Comment{\cref{eq:ffgs}}
\State Split $\mathcal{G}_{\text{init}}$ into $\mathcal{G}_s$ and $\mathcal{G}_d^0$ by flow labels \Comment{\cref{eq:flowlabel}}
\State Relocate $\mathcal{G}_s\cup\mathcal{G}_d^0$ \Comment{MVR, \cref{sec:supp_reloc}}
\For{$n=1,\dots,N_{\text{snap}}$} \Comment{Snapshot Training}
  \State Render $\mathcal{G}_s\cup\mathcal{G}_d^0$ at a camera of $t_0$
  \State Update $\boldsymbol{\mathcal{G}_s}$, $\boldsymbol{\mathcal{G}_d^0}$ by $\mathcal{L}_{\text{photo}}$
\EndFor
\Statex \vspace{-4pt}
\State \textbf{Stage 2: Dynamic expansion} (\cref{sec:expansion})
\State Track flow trajectories over all frames \Comment{\cref{sec:supp_traj}}
\State Move barely moving Gaussians of $\mathcal{G}_d^0$ to $\mathcal{G}_s$ \Comment{D$\rightarrow$S}
\For{$k=1,\dots,K$}
  \State $\mathcal{G}_d^k \gets \mathcal{G}_d^0$ moved to $t_k$ \Comment{Keyframe Propagation}
  \State Add seeds in emergent regions of $t_k$ to $\mathcal{G}_d^k$ \Comment{ECS}
\EndFor
\For{$n=1,\dots,N_{\text{kf}}$} \Comment{Keyframe training}
  \State Render $\mathcal{G}_s\cup\mathcal{G}_d^k$ at a camera of $t_k$, $k\geq1$
  \State Update $\boldsymbol{\mathcal{G}_d^k}$ by $\mathcal{L}_{\text{photo}}$
\EndFor
\State Move high-gradient Gaussians of $\mathcal{G}_s$ to all $\mathcal{G}_d^k$ \Comment{S$\rightarrow$D}
\Statex \vspace{-4pt}
\State \textbf{Stage 3: Dynamic training} (\cref{sec:training})
\State $\mathcal{G}_d \gets \bigcup_{k=0}^{K}\mathcal{G}_d^k$, \ $C_{\text{static}}^v \gets \textsc{Rasterize}(\mathcal{G}_s, v)$ for all $v$
\For{$n=1,\dots,N_{\text{dyn}}$} \Comment{Dynamic training}
  \State Cull $\mathcal{G}_d$ at $t$ of $I_t^v$ \Comment{Temporal Culling}
  \State $\hat{I}_t^v \gets \textsc{SCR}(\mathcal{G}_d, C_{\text{static}}^v, v, t)$ \Comment{\cref{alg:scr}}
  \State Update $\boldsymbol{\mathcal{G}_d}$ by $\mathcal{L}_{\text{photo}}$ and the render cost \Comment{RCR}
\EndFor
\State \Return $\mathcal{G}_s$, $\mathcal{G}_d$, $\{C_{\text{static}}^v\}$
\end{algorithmic}
\end{algorithm}

\subsection{Flow Trajectories and Keyframe Propagation}
\label{sec:supp_traj}
\paragraph{Flow trajectories.}
Trajectories are computed only in the four feed-forward input views, with optical flow at an interval of $\Delta\tau=15$ frames.
A trajectory stays valid at a tracked frame only if it was valid at the previous tracked frame, remains inside the image, and passes the forward--backward check
\begin{equation}
\big\lVert \mathbf{f}+F^v_{t+\Delta\tau\rightarrow t}\big(\mathbf{p}_v(t)+\mathbf{f}\big)\big\rVert_2 < 1 + 0.05\,\lVert\mathbf{f}\rVert_2,
\label{eq:supp_fb}
\end{equation}
where $\mathbf{f}=F^v_{t\rightarrow t+\Delta\tau}(\mathbf{p}_v(t))$ is the forward flow and the error is in pixels.
Once a trajectory becomes invalid, it is not used afterward.

\paragraph{Keyframe Propagation.}
The keyframes are placed every 30 frames, with the first one at frame 15.
Let $\Gamma_v(\mathbf{p})=\{\mathbf{p}_v(t)\}_t$ be the trajectory of a first-frame pixel $\mathbf{p}$ in input view $v$ from \cref{eq:track}.

Each dynamic Gaussian is copied to every keyframe and placed by following the trajectories of its first-frame pixels in the input views.
The 3D position of the copy of Gaussian $i$ at keyframe $t_k$ is
\begin{equation}
\boldsymbol{\mu}_i(t_k) = \frac{1}{|\mathcal{V}_i(t_k)|}\sum_{v\in\mathcal{V}_i(t_k)} \pi_v^{-1}\big(\mathbf{p}_v(t_k),\, d_{i,v}\big),
\label{eq:propagate}
\end{equation}
that is, for each input view $v\in\mathcal{V}_i(t_k)$ in which Gaussian $i$ is not occluded in the first frame and its trajectory has not been discarded by $t_k$, we take the position $\mathbf{p}_v(t_k)$ of the trajectory $\Gamma_v(\mathbf{p}_{i,v})$ of its first-frame pixel $\mathbf{p}_{i,v}$, unproject it with its first-frame depth $d_{i,v}$ by $\pi_v^{-1}$, and average the resulting points over the views.

The set $\mathcal{V}_i(t_k)$ excludes the views in which the trajectory does not follow the Gaussian, and Gaussians without any remaining view are placed from their neighbors.
Views in which the Gaussian is hidden behind other content in the first frame are excluded, since the trajectory of its pixel there follows the occluding content rather than the Gaussian.
To detect them, the centers of all Gaussians with opacity above 0.1 are projected into each input view, and the nearest center depth is kept at each pixel.
A Gaussian is regarded as occluded in a view when its depth exceeds this nearest depth at its pixel by more than 3\%.
A Gaussian with an empty $\mathcal{V}_i(t_k)$, which is occluded in all input views or whose trajectories have all been discarded, cannot be placed by \cref{eq:propagate}.
Since nearby dynamic Gaussians usually belong to the same moving object, it is moved by the median displacement of its 16 nearest neighbors that have a non-empty $\mathcal{V}_i(t_k)$, within a radius of 3\% of the diagonal of the bounding box of the dynamic Gaussians.

The velocity of a copy is initialized from the positions given by \cref{eq:propagate} five frames before and after $t_k$,
\begin{equation}
\dot{\boldsymbol{\mu}}_i(t_k) = \frac{\boldsymbol{\mu}_i(t_k+5)-\boldsymbol{\mu}_i(t_k-5)}{10},
\label{eq:supp_velocity}
\end{equation}
per frame, where the positions at $t_k\pm5$, which lie between tracked frames, are linearly interpolated, and the window is clipped at the ends of the sequence.

\paragraph{Keyframe training.}
During keyframe training, the static Gaussians stay frozen, each copy is trained only with the images of its keyframe, and the dynamic Gaussians of the snapshot are not updated.

\subsection{Seeding and Label-Correction Criteria}
\label{sec:supp_criteria}
\paragraph{Emergent-Content Seeding.}
Emergent-Content Seeding adds dynamic Gaussians for content that is absent from the first frame, such as a person entering the scene, since no propagated Gaussian can represent it.
Such content appears at holes, the pixels of a keyframe that no flow trajectory of the first-frame pixels reaches, so seeding first finds the holes.

For each feed-forward input view and keyframe, every valid trajectory position marks its pixel and the pixels within 2 pixels as covered, where pixels are counted at a width of 512.
With $\mathcal{P}_v(t_k)$ the set of valid trajectory positions in view $v$ at $t_k$, the hole mask is the uncovered region after a morphological opening,
\begin{equation}
H_v(t_k) = \mathrm{open}\Big(\neg\,\mathrm{dilate}_2\big(\mathcal{P}_v(t_k)\big)\Big),
\label{eq:supp_hole}
\end{equation}
where $\mathrm{dilate}_2$ dilates by 2 pixels, and connected regions smaller than 50 pixels are then removed.

A keyframe is seeded only when the remaining holes cover at least 0.5\% of the image on average over the input views.
For a seeded keyframe, the feed-forward model is run on its input views, and its Gaussians at the hole pixels on a grid with a stride of 2 pixels are added to the copy of the keyframe as dynamic Gaussians with zero velocity.

\paragraph{D$\rightarrow$S correction.}
The flow label of a Gaussian in the snapshot stage uses the flow of a single input view, the view whose pixel predicts the Gaussian, so Gaussians that barely move can still be labeled dynamic, for instance when the flow of that view is noisy.
Such Gaussians do not need to learn motion.
The flow trajectories observe the same Gaussian in all input views in which it is visible, which gives a more reliable decision.
For each dynamic Gaussian, the trajectories of its pixels in the input views in which it is visible are read, as in \cref{sec:supp_traj}.
The Gaussian is relabeled static when it is visible in at least one input view and, in every such view $v$ with projected pixel $\mathbf{p}$,
\begin{equation}
\frac{|\{t: \mathbf{p}_v(t)\ \text{valid}\}|}{|\{t\}|} \ge 0.5
\quad\text{and}\quad
\max_{t:\,\text{valid}} \big\lVert \mathbf{p}_v(t)-\mathbf{p} \big\rVert_2 < \tau_D,
\label{eq:supp_d2s}
\end{equation}
where $t$ runs over the tracked frames and $\tau_D$ is the motion threshold (\cref{tab:supp_hparams}).
The Gaussian is relabeled before propagation, so it is not copied to the keyframes.

\paragraph{S$\rightarrow$D correction.}
The flow labels can also miss moving Gaussians, for instance when their motion lies along the viewing direction of the labeling view, is too small, or occurs in textureless regions where the flow fails, and these Gaussians stay frozen as static.
Such a static Gaussian is pulled in different directions by the losses of different keyframes, which the gradient inconsistency $\eta_i$ detects.
After keyframe training, the position gradients $\mathbf{g}_{i,k,c}$ in \cref{eq:gradswitch} are computed once for all static Gaussians, rendering each training camera $c$ of keyframe $t_k$ with $\mathcal{G}_s\cup\mathcal{G}_d^k$.
Static Gaussians with gradient inconsistency $\eta_i>\lambda_\eta=0.4$ are removed from $\mathcal{G}_s$, and a copy of each is added to every $\mathcal{G}_d^k$ with temporal center $t_k$ and zero velocity.
Since this happens before the static cache is computed, the cache contains only the remaining static Gaussians.

\subsection{Static-Caching Rasterization}
\label{sec:supp_scr}
\Cref{alg:scr} gives Static-Caching Rasterization for one camera and time step.
Static-Caching Rasterization is used both in dynamic training and in rendering, and in dynamic training the gradients reach only the dynamic Gaussians.
The cache of each training camera is computed once after dynamic expansion (\cref{alg:pipeline}).
At test time, the cache of the test camera is computed by lines 1 to 3 at the first frame and reused for all other frames.
The cache is used from dynamic training onward (\cref{sec:supp_scr_renderonly}).

\begin{algorithm}[t]
\caption{Static-Caching Rasterization ($\textsc{SCR}$) of camera $v$ at time $t$.}
\label{alg:scr}
\begin{algorithmic}[1]
\Require Static Gaussians $\mathcal{G}_s$, dynamic Gaussians $\mathcal{G}_d$, camera $v$, time $t$, cache $C_{\text{static}}^v$ (empty at first)
\Ensure Rendered image $\hat{I}$
\If{$C_{\text{static}}^v$ is empty}
  \State $C_{\text{static}}^v \gets \textsc{Rasterize}(\mathcal{G}_s, v)$ \Comment{once per camera}
\EndIf
\State $\mathcal{G}_d(t) \gets \{\, i\in\mathcal{G}_d \mid \sigma_i(t)\ge 1/255 \,\}$ \Comment{Temporal Culling, \cref{eq:topacity}}
\For{each pixel $\mathbf{p}$} \Comment{one rasterization pass}
  \State $C \gets 0$, $T \gets 1$
  \For{$g_i\in\mathcal{G}_d(t)$ overlapping the tile of $\mathbf{p}$, in front-to-back order}
    \State $C \gets C + T\,\alpha_i\,\mathbf{c}_i$, \quad $T \gets T\,(1-\alpha_i)$
    \If{$T < \epsilon$} \textbf{break} \EndIf
  \EndFor
  \State $\hat{I}(\mathbf{p}) \gets C + T\, C_{\text{static}}^v(\mathbf{p})$ \Comment{\cref{eq:scr}}
\EndFor
\State \Return $\hat{I}$
\end{algorithmic}
\end{algorithm}

\subsection{Derivation of the Render-Cost Terms}
\label{sec:supp_cost}
This section derives the render-cost term of \cref{eq:costloss} and validates it by measurement.
The frame-time model expresses the time of tile-based rasterization through the number of Gaussian--tile pairs under assumptions (A1) and (A2).
The total cost extends it to a sequence under assumption (A3), which gives $\sum_i A_iT_i$.
The validation tests the model and the assumptions by regressions on 45 models, and the loss term describes how the cost is added to training.
With the static part cached, only the dynamic Gaussians that pass Temporal Culling are rasterized, and $\mathcal{G}_d(t)$ below denotes these Gaussians at time $t$.

\paragraph{Frame-time model.}
Let $\mathcal{T}_i$ be the set of $16\times16$ tiles that Gaussian $i$ overlaps in a frame.
The rasterizer of 3DGS~\cite{3dgs} projects every Gaussian and creates, for each pair of a Gaussian and an overlapped tile, a key that concatenates the tile ID and the depth of the Gaussian, with the Gaussian ID as its value.
The number of keys is
\begin{equation}
P = \sum_i \lvert\mathcal{T}_i\rvert = \sum_{\tau} \big\lvert\{\, i : \tau\in\mathcal{T}_i \,\}\big\rvert,
\label{eq:supp_npairs}
\end{equation}
the number of Gaussian--tile pairs, which also equals the total number of Gaussians listed over all tiles $\tau$.
The keys are sorted with a radix sort, which groups them by tile in front-to-back order, and the Gaussians of each tile are blended in this order.
Let $N$ be the number of Gaussians with $\mathcal{T}_i\ne\emptyset$, so that $P\ge N$.
We assume that (A1) the keys have a fixed length, so the radix sort runs a fixed number of passes over the keys, and that (A2) the blending cost per pair, which depends on the number of pixels the pair actually blends and on early termination, stays approximately constant across frames.
With the per-unit costs $\kappa_{\text{proj}}$, $\kappa_{\text{key}}$, and $\kappa_{\text{sort}}$ of projecting one Gaussian, creating one key, and sorting one key,
\begin{align}
t_{\text{proj}} &= \kappa_{\text{proj}}N, \label{eq:supp_tproj}\\
t_{\text{key}}+t_{\text{sort}} &= (\kappa_{\text{key}}+\kappa_{\text{sort}})P \quad\text{by (A1)}, \label{eq:supp_tsort}\\
t_{\text{blend}} &\approx \kappa_{\text{blend}}P \quad\text{by (A2)}, \label{eq:supp_tblend}
\end{align}
where $0\le\kappa_{\text{blend}}\le256\,\kappa_{\text{px}}$ for the cost $\kappa_{\text{px}}$ of blending one pixel, since a pair blends at most the 256 pixels of its tile.
Adding the fixed cost $a$ of a frame, which covers kernel launches, the computation of tile ranges, buffer initialization, and the composition with the cache, gives
\begin{equation}
t_{\text{frame}} \approx a + b\,N + c\,P,
\label{eq:supp_tframe}
\end{equation}
with $b=\kappa_{\text{proj}}$ and $c=\kappa_{\text{key}}+\kappa_{\text{sort}}+\kappa_{\text{blend}}$.

\paragraph{Total cost.}
We further assume that (A3) a dynamic Gaussian overlaps the same number of tiles in every frame in which it is visible, which holds when its depth changes little during these frames.
A Gaussian with maximum scale $\bar{s}_i$ at depth $d_i$ projects to a screen-space Gaussian whose radius of three standard deviations under focal length $f$ is $r_i\approx 3f\bar{s}_i/d_i$ pixels.
Approximating its footprint by a disk of radius $r_i$ and neglecting partly covered tiles,
\begin{equation}
\lvert\mathcal{T}_i\rvert \approx \frac{\pi r_i^2}{16^2} = A_i,
\quad\text{so}\quad
P(t) \approx \sum_{i\in\mathcal{G}_d(t)} A_i .
\label{eq:supp_pairs}
\end{equation}
By \cref{eq:topacity}, $\sigma_i(t)\ge 1/255$ if and only if $\lvert t-\mu^\tau_i\rvert \le \sqrt{\ln(255\sigma_i)/s^\tau_i}$, so Gaussian $i$ belongs to $\mathcal{G}_d(t)$ for $T_i$ frames in \cref{eq:visibletime}.
Summing over the frames and exchanging the two sums,
\begin{equation}
\sum_t P(t) \approx \sum_t\sum_{i\in\mathcal{G}_d(t)} A_i
= \sum_{i\in\mathcal{G}_d} A_i T_i .
\label{eq:supp_totalcost}
\end{equation}

\paragraph{Validation.}
We verify three points by measurement, namely that the frame time follows \cref{eq:supp_tframe} with the term $cP$ dominant, that assumptions (A1)--(A3) hold, and that $\sum_i A_iT_i$ tracks the total number of pairs over a sequence as in \cref{eq:supp_totalcost}.
To this end, we measure 45 models on the five Technicolor scenes, namely the full model and eight variants, which are the seven ablation variants in \cref{tab:ablation} that keep Dynamic--Static Decomposition and Static-Caching Rasterization and a variant without both label corrections.
For every frame of the test camera, we record the rendering time on an RTX 5090 GPU, the number $P$ of Gaussian--tile pairs, and the number $N$ of visible dynamic Gaussians, 14{,}463 frames in total.
For the full model, we also record the time of each kernel of the rasterizer to test (A1) and (A2).
\Cref{tab:supp_costfit} lists the linear regressions (R1)--(R6), with times in ms and $P$ and $N$ in millions, so that a slope gives the increase in time in ms per million pairs or Gaussians.

\begin{table}[t]
\centering
\caption{Regressions that validate the render-cost model on Technicolor~\cite{technicolor}. Times are in ms, and $P$, $N$, and the pair totals of (R6) are in millions.}
\label{tab:supp_costfit}
\scriptsize
\setlength{\tabcolsep}{3pt}
\begin{adjustbox}{max width=\linewidth}
\begin{tabular}{@{}c l c c c@{}}
\toprule
 & Regression & Intercept & Slope & $R^2$ \\
\midrule
(R1) & $t_{\text{frame}}$ on $P$ & 0.30 & 0.34 & 0.996 \\
(R2) & $t_{\text{frame}}$ on $N$ & 0.53 & 5.73 & 0.39 \\
(R3) & $t_{\text{frame}}$ on $N$ and $P$ (\cref{eq:supp_tframe}) & $a=0.29$ & $b=0.11$, $c=0.34$ & 0.997 \\
\midrule
(R4) & $t_{\text{blend}}$ on $P$ (A2) & 0.05 & 0.20 (0.20--0.28) & 0.998 \\
(R5) & $t_{\text{sort}}$ on $P$ (A1), per scene & $-0.32$--0.04 & 0.03--0.13 & 0.85--0.99 \\
\midrule
(R6) & $\sum_t P(t)$ on $\sum_i A_iT_i$ (\cref{eq:supp_totalcost}), per scene & 91--996 & 0.08--0.43 & 0.74--0.99 \\
\bottomrule
\end{tabular}
\end{adjustbox}
\end{table}

\begin{figure}[h]
\centering
\includegraphics[width=\linewidth]{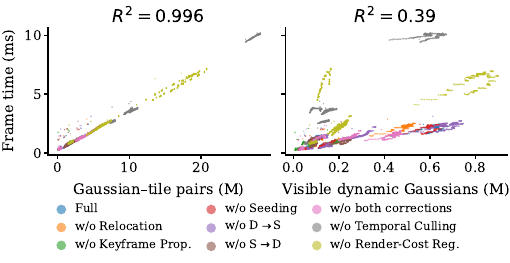}
\caption{Frame time against the number $P$ of Gaussian--tile pairs (left) and the number $N$ of visible dynamic Gaussians (right) for every test frame of the 45 models on Technicolor~\cite{technicolor}, corresponding to (R1) and (R2) in \cref{tab:supp_costfit}. The frame time follows $P$ but not $N$.}
\label{fig:supp_rendercost}
\end{figure}

The frame time follows $P$ rather than $N$.
With $P$ alone (R1), $R^2$ is 0.996 and one million more pairs add 0.34\,ms per frame (\cref{fig:supp_rendercost}), whereas $N$ alone (R2) gives $R^2=0.39$ and cannot model the frame time.
Fitting \cref{eq:supp_tframe} with both (R3) also holds with $R^2=0.997$ and gives $c=0.34$\,ms and $b=0.11$\,ms per million.
An average frame has $P=3.79$ million pairs and $N=0.19$ million visible dynamic Gaussians, since a dynamic Gaussian overlaps 20 tiles on average.
The term $bN$ is thus 0.02\,ms, or 1.3\% of the average frame time of 1.59\,ms, against 1.28\,ms, or 80\%, for $cP$, so the term $cP$ dominates.

The kernel times support (A1) and (A2), and the tile counts of each Gaussian support (A3).
The sorting kernel, which takes 15--22\% of the kernel time, grows linearly with $P$ within each scene (R5), which supports (A1) within a scene.
The blending kernel, which takes 39--63\% of the kernel time, grows linearly with $P$ with a similar slope in all scenes (R4), and a pair blends 216--241 of the 256 pixels of its tile with a variation below 2\% across frames, which supports (A2).
Over its visible frames, the number of tiles that a dynamic Gaussian overlaps varies by 7--12\% in the median, which supports (A3).
$A_i$ underestimates the measured tile count because of tile quantization but correlates with it at 0.85--0.94 in log scale, which suffices since the cost is normalized by $C_0$.

Finally, (R6) excludes the variant without Temporal Culling, since it rasterizes every dynamic Gaussian in every frame and its $T_i$ is therefore not the interval of \cref{eq:supp_totalcost}.
Within each scene, $\sum_i A_i T_i$ then tracks the number of pairs summed over the sequence with $R^2$ of 0.74--0.99 (R6).
The slope differs across scenes, which does not affect the loss since it is normalized by $C_0$ of each scene.

\paragraph{Loss term.}
By \cref{eq:supp_tframe,eq:supp_totalcost} and (R1)--(R6), the rendering time of a sequence increases with $\sum_i A_i T_i$, so we add this cost to the training loss in \cref{eq:costloss}.
Normalizing it by the cost $C_0$ at the start of regularization removes the scale difference across scenes, so a single weight $\lambda_c$ applies to all scenes.
The peak opacity in $T_i$ is detached by stop-gradient, so that the regularizer reduces the cost by shrinking the screen area and the visible duration instead of lowering opacities.
At each iteration, $A_i$ is computed with the depth $d_i$ and the focal length $f$ of a training camera of the batch, only the dynamic Gaussians in front of that camera and inside its view frustum are summed, and $T_i$ is clipped to the sequence length.
The regularization starts at iteration $n_c=3.5$K, when $C_0$ is recorded, and is skipped for 500 iterations after each opacity reset of STG~\cite{stg}, which occurs every 3K iterations until densification ends at 15K.

\section{Implementation Details}
\label{sec:supp_impl}

\subsection{Benchmarks}
\label{sec:supp_bench}
\paragraph{Neural 3D.}
Neural 3D is captured by 18--21 cameras at $2704\times2028$ and shows a person cooking in an indoor kitchen.
The person is present from the first frame, and new content appears only locally, such as the flame of the torch in \emph{flame\_salmon} and \emph{flame\_steak} and the poured liquid in \emph{coffee\_martini}.
Its scenes contain 300 frames each and use 17--20 training cameras, and following prior work~\cite{stg,4dgs}, their images are downsampled by a factor of two to $1352\times1014$.
Camera 00 is held out for testing.

\paragraph{Technicolor.}
Technicolor is captured by a $4\times4$ rig of 16 cameras, and following prior work~\cite{stg}, we use five of its scenes.
Objects absent from the first frame enter the scene later, namely people in \emph{Birthday} and \emph{Painter}, puppets in \emph{Theater}, and trains in \emph{Train}.
We use the full sequences, which contain 304, 200, 372, 391, and 340 frames (\emph{Birthday}, \emph{Fabien}, \emph{Painter}, \emph{Theater}, and \emph{Train}), with 15 training cameras at $2048\times1088$.
Camera 10 is held out for testing.

\paragraph{Panoptic Studio.}
To evaluate static occluders under 360$^\circ$ capture (\cref{sec:supp_static_occluder}), we use the \emph{office} sequence of the Panoptic Studio~\cite{joo2015panoptic}, in which a person walks behind a static monitor.
We use 150 frames captured by 28 HD cameras inside a dome at $640\times360$ and hold out HD camera 15 for testing.

For all three benchmarks, camera poses are provided by the datasets.

\subsection{Hyperparameters and Timing}
\label{sec:supp_hparams}
\paragraph{Pretrained models.}
The feed-forward encoder $\mathcal{E}$ is DepthSplat~\cite{depthsplat} with its released checkpoint trained on RealEstate10K~\cite{re10k} and DL3DV~\cite{dl3dv}.
All optical flow, for both the flow labeling and the flow trajectories, is computed by FlowFormer~\cite{huang2022flowformer} with its released Sintel checkpoint.

\paragraph{Hyperparameters.}
\Cref{tab:supp_hparams} lists the hyperparameters, which are shared by all scenes.
The motion thresholds $\tau_M$ and $\tau_D$ are given as a fraction of the image width, so that the same values apply to both benchmarks.
On Technicolor and Neural 3D, whose images are 2048 and 1352 pixels wide, the flow threshold $\tau_M$ of 0.117\% corresponds to about 2.4 and 1.6 pixels, and the D$\rightarrow$S motion threshold $\tau_D$ of 0.44\% to about 9 and 6 pixels, respectively.

\begin{table}[t]
\centering
\caption{Hyperparameters of DSD-GS.}
\label{tab:supp_hparams}
\scriptsize
\setlength{\tabcolsep}{3pt}
\begin{adjustbox}{max width=\linewidth}
\begin{tabular}{@{}l l@{}}
\toprule
Hyperparameter & Value \\
\midrule
\multicolumn{2}{@{}l}{\textit{Snapshot}} \\
Flow threshold $\tau_M$ & 0.117\% of the image width \\
Flow sampling interval $\Delta t$ & 10 frames \\
Snapshot Training $N_{\text{snap}}$ & 10K iterations \\
\midrule
\multicolumn{2}{@{}l}{\textit{Dynamic expansion}} \\
Tracking interval $\Delta\tau$ & 15 frames \\
Keyframe interval & 30 frames \\
Keyframe training $N_{\text{kf}}$ & 3K iterations \\
D$\rightarrow$S motion threshold $\tau_D$ & 0.44\% of the image width \\
S$\rightarrow$D threshold $\lambda_\eta$ & 0.4 \\
\midrule
\multicolumn{2}{@{}l}{\textit{Dynamic training}} \\
Dynamic training $N_{\text{dyn}}$ & 20K iterations \\
Render-cost weight $\lambda_c$ & 0.003 \\
\midrule
Optimizer & Adam \\
\bottomrule
\end{tabular}
\end{adjustbox}
\end{table}

\paragraph{Baselines.}
All baselines are trained with their official implementations and configurations on the same full-length sequences and test cameras as DSD-GS, and the test camera is excluded from their initialization.

\paragraph{Timing.}
All initialization and training times are measured on a single RTX 5090 GPU.
The initialization time of DSD-GS covers all steps other than the optimization, such as the feed-forward prediction, the flow labeling, Multi-View Ray Relocation, the flow trajectories, and Emergent-Content Seeding.
When DSD-GS is initialized with dense COLMAP on the first frame, this COLMAP replaces the feed-forward prediction and Multi-View Ray Relocation, Emergent-Content Seeding uses sparse COLMAP on the keyframes instead of the feed-forward model, and the other steps are unchanged.
The initialization time of the baselines covers COLMAP, either sparse on all frames or dense on the first frame.
For Swift4D~\cite{swift4d}, it also includes the computation of the per-pixel temporal standard deviation maps.
For both, data preparation such as image downsampling is excluded.
As in the baselines, the training time includes only the optimization, which for DSD-GS covers Snapshot Training, keyframe training, and dynamic training, and excludes data loading.
For the rendering speed, the first 10 frames of the baselines are discarded as warm-up, whereas for DSD-GS no frame is discarded, so its time includes building the static cache.

\section{Analysis}
\label{sec:supp_analysis}

\subsection{Qualitative Ablation}
\label{sec:supp_qual_ablation}
\begin{figure*}[t]
\centering
\includegraphics[width=\linewidth]{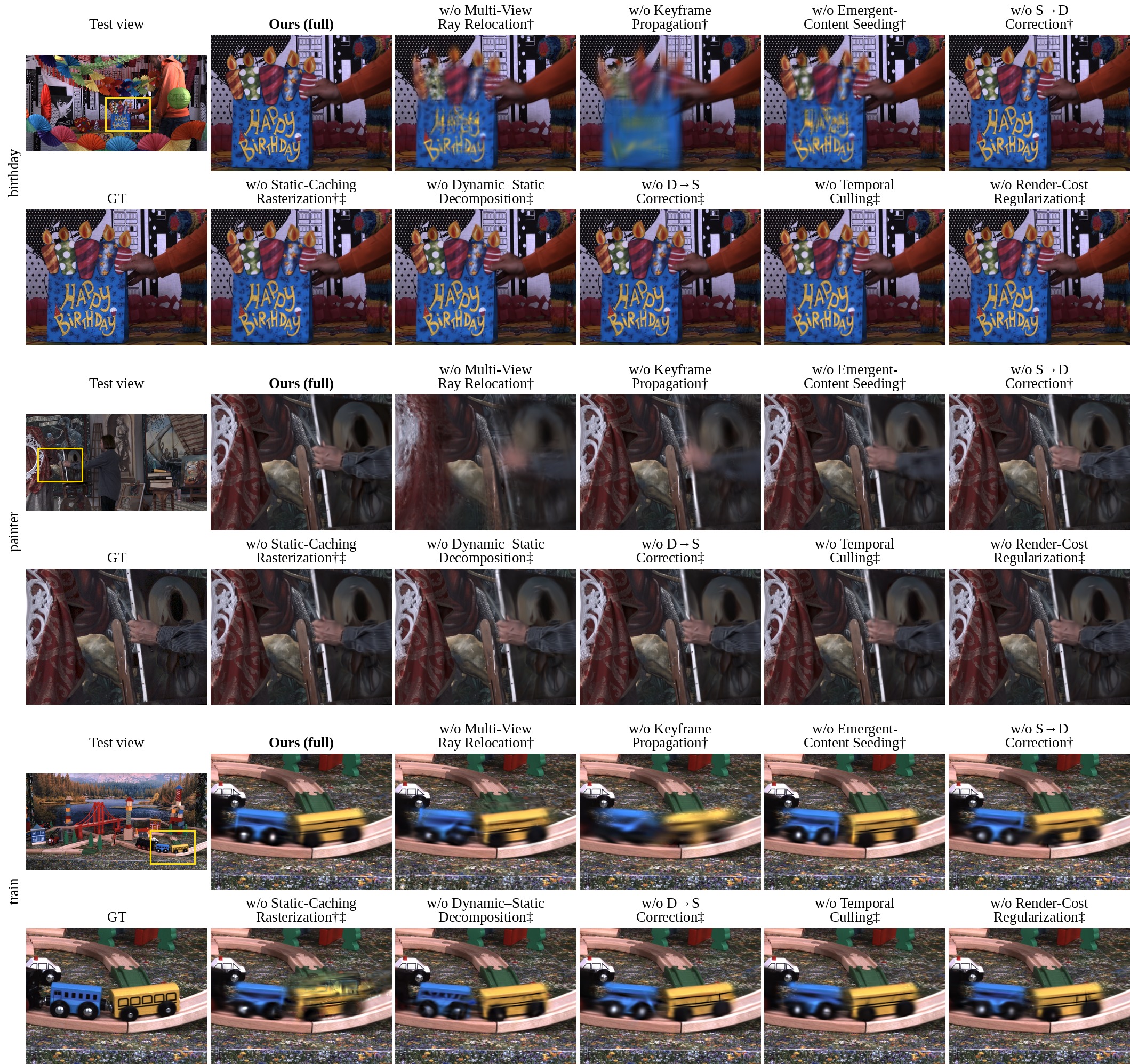}
\caption{Qualitative ablation on \emph{Birthday}, \emph{Painter}, and \emph{Train} of Technicolor~\cite{technicolor}. The yellow box in the test view marks the crop. $^\dagger$ and $^\ddagger$ mark the components for quality and efficiency, respectively, as in \cref{tab:ablation}.}
\label{fig:supp_qual_ablation}
\end{figure*}

\Cref{fig:supp_qual_ablation} shows the renderings of the full model and of the nine ablation variants in \cref{tab:ablation} on three Technicolor scenes, where $^\dagger$ and $^\ddagger$ mark the components for quality and efficiency, respectively.
Most variants without a quality component show visible errors.
Without Multi-View Ray Relocation, the static background is blurred, most clearly in \emph{Painter}.
Without Keyframe Propagation, content that has moved far from the first frame is blurred, such as the bag in \emph{Birthday}, the arm in \emph{Painter}, and the train in \emph{Train}.
Without Emergent-Content Seeding, the letters on the bag in \emph{Birthday} are blurred.
Without Static-Caching Rasterization, artifacts appear on the moving train in \emph{Train}.
The variants without an efficiency component render nearly the same as the full model, and they differ in training time and rendering speed (\cref{tab:ablation}).

\subsection{Per-Frame Ablation}
\label{sec:supp_perframe}
We analyze the per-frame PSNR to verify that Keyframe Propagation and Emergent-Content Seeding improve the frames that they target.

\begin{figure*}[t]
\centering
\includegraphics[width=\linewidth]{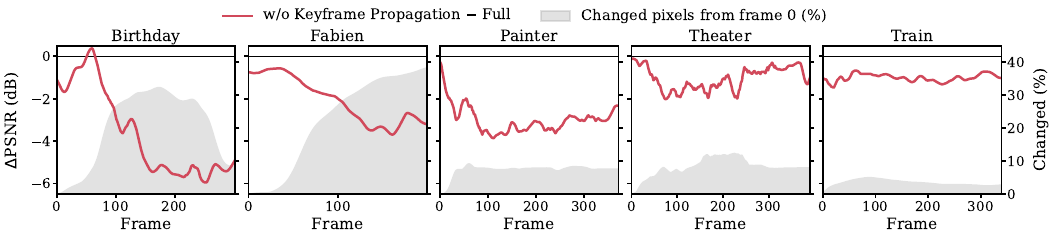}
\caption{Per-frame PSNR difference between the model without Keyframe Propagation and the full model on Technicolor~\cite{technicolor}, smoothed over 15 frames. The gray area shows the fraction of pixels whose ground-truth color in the test view differs from that of frame~0.}
\label{fig:supp_kf_perframe}
\end{figure*}

Keyframe Propagation places copies of the dynamic Gaussians at keyframes so that content that moves far from the first frame can be represented.
To see this effect, \cref{fig:supp_kf_perframe} plots the per-frame PSNR difference between the model without it and the full model, together with the fraction of pixels that have changed from the first frame.
On \emph{Birthday} and \emph{Fabien}, the drop grows as more pixels change from the first frame and reaches 6.0\,dB and 3.7\,dB, respectively.
On \emph{Painter}, the drop rises to about 3\,dB once a person enters at frame 16 and stays there while the person moves in the scene.
On \emph{Train}, where the fewest pixels change, the drop stays around 1\,dB.
On \emph{Theater}, the drop shrinks after frame 250, when the content returns closer to that of the first frame.

\begin{figure}[t]
\centering
\includegraphics[width=\linewidth]{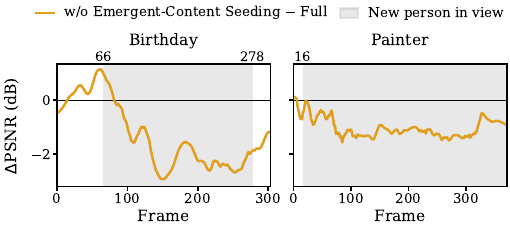}
\caption{Per-frame PSNR difference between the model without Emergent-Content Seeding and the full model on the two Technicolor~\cite{technicolor} scenes in which a person enters the view, smoothed over 9 frames. The gray area marks the frames in which the person is visible in the test view, with the first and last frames shown above.}
\label{fig:supp_ecs_perframe}
\end{figure}

Emergent-Content Seeding adds dynamic Gaussians for content that is absent from the first frame, such as a person who enters the scene.
To see this effect, \cref{fig:supp_ecs_perframe} plots the per-frame PSNR difference between the model without it and the full model on the two scenes in which a person enters the view, together with the frames in which the person is visible.
On \emph{Birthday}, the model without seeding is not worse before the person enters at frame 66, drops sharply once the person enters, by up to 2.9\,dB, and partly recovers after the person leaves at frame 278.
On \emph{Painter}, the person enters at frame 16 and stays until the end, and the drop of about 1\,dB lasts accordingly.

\subsection{Sensitivity Analysis}
\label{sec:supp_sens}
To examine how sensitive DSD-GS is to its hyperparameters, \cref{tab:supp_sens} varies each hyperparameter to a smaller and a larger value on three Technicolor scenes (\emph{Painter}, \emph{Theater}, and \emph{Train}), keeping the other hyperparameters at their defaults and reusing the same Snapshot Training except for the labeling hyperparameters.
On Technicolor, the flow is computed at a width of 256 pixels, where the default $\tau_M$ of 0.117\% of the image width (\cref{sec:supp_hparams}) corresponds to 0.3 pixels, so the table gives $\tau_M$ in pixels.
Except for the keyframe interval and the S$\rightarrow$D threshold, the PSNR changes by at most 0.16\,dB, which shows that DSD-GS is insensitive to these hyperparameters.
The keyframe interval and the S$\rightarrow$D threshold trade efficiency for quality, and we choose their defaults to balance the two.
Doubling the keyframe interval to 60 frames lowers the PSNR by 0.30\,dB, while halving it to 15 frames raises the PSNR by 0.08\,dB at 15\% longer training.
Halving the S$\rightarrow$D threshold $\lambda_\eta$ raises the PSNR by 0.12\,dB, but it relabels many more Gaussians as dynamic, which lengthens the training by 60\% and halves the rendering speed.

\begin{table}[t]
\centering
\caption{Sensitivity to hyperparameters on three Technicolor~\cite{technicolor} scenes. Each row reports a smaller, the default (bold), and a larger value.}
\label{tab:supp_sens}
\scriptsize
\setlength{\tabcolsep}{2.5pt}
\begin{adjustbox}{max width=\linewidth}
\begin{tabular}{@{}l c c c@{}}
\toprule
Hyperparameter & Values & PSNR$\uparrow$ & LPIPS$\downarrow$ \\
\midrule
\multicolumn{4}{@{}l}{\textit{Snapshot}} \\
Flow threshold $\tau_M$ (px) & 0.15 / \textbf{0.3} / 0.6 & 31.20 / 31.26 / 31.15 & 0.103 / 0.104 / 0.104 \\
Flow interval $\Delta t$ (frames) & 5 / \textbf{10} / 20 & 31.18 / 31.26 / 31.10 & 0.104 / 0.104 / 0.104 \\
\midrule
\multicolumn{4}{@{}l}{\textit{Dynamic expansion}} \\
Keyframe interval (frames) & 15 / \textbf{30} / 60 & 31.34 / 31.26 / 30.96 & 0.101 / 0.104 / 0.108 \\
Occlusion depth tolerance & 1.5 / \textbf{3} / 6\% & 31.13 / 31.26 / 31.16 & 0.104 / 0.104 / 0.105 \\
Occlusion opacity & 0.05 / \textbf{0.1} / 0.2 & 31.21 / 31.26 / 31.22 & 0.104 / 0.104 / 0.104 \\
Neighbor count & 8 / \textbf{16} / 32 & 31.24 / 31.26 / 31.21 & 0.104 / 0.104 / 0.104 \\
Neighbor radius & 1.5 / \textbf{3} / 6\% & 31.18 / 31.26 / 31.16 & 0.104 / 0.104 / 0.104 \\
Seeding hole fraction & 0.25 / \textbf{0.5} / 1\% & 31.17 / 31.26 / 31.19 & 0.104 / 0.104 / 0.104 \\
Seeding stride (px) & 1 / \textbf{2} / 4 & 31.17 / 31.26 / 31.19 & 0.102 / 0.104 / 0.105 \\
Hole dilation (px) & 1 / \textbf{2} / 4 & 31.11 / 31.26 / 31.17 & 0.105 / 0.104 / 0.105 \\
Hole minimum area (px) & 25 / \textbf{50} / 100 & 31.25 / 31.26 / 31.19 & 0.104 / 0.104 / 0.105 \\
D$\rightarrow$S motion threshold $\tau_D$ (\% of width) & 0.22 / \textbf{0.44} / 0.89 & 31.22 / 31.26 / 31.18 & 0.104 / 0.104 / 0.105 \\
D$\rightarrow$S validity ratio & 0.25 / \textbf{0.5} / 0.75 & 31.18 / 31.26 / 31.27 & 0.105 / 0.104 / 0.104 \\
S$\rightarrow$D threshold $\lambda_\eta$ & 0.2 / \textbf{0.4} / 0.8 & 31.38 / 31.26 / 31.13 & 0.100 / 0.104 / 0.106 \\
\bottomrule
\end{tabular}
\end{adjustbox}
\end{table}

\subsection{Render-Cost Trade-Off}
\label{sec:supp_costtradeoff}
To examine how Render-Cost Regularization trades quality for rendering speed, \cref{tab:supp_costtradeoff} varies the render-cost weight $\lambda_c$ and the form of the cost on the five Technicolor scenes.
From $\lambda_c=0.001$ to the default $\lambda_c=0.003$, the rendering speed rises from 875 to 1033\,FPS with the same PSNR.
Larger weights trade quality for speed gradually, and $\lambda_c=0.1$ renders 1.75$\times$ faster than the default at a PSNR loss of 0.28\,dB.
With the same weight, penalizing only the covered tiles $A_i$ gives the same quality and penalizing the L1 norm of the scales lowers the PSNR by 0.10\,dB, while both render 12\% and 20\% slower than the product $A_iT_i$, which also penalizes the visible duration.

\begin{table}[t]
\centering
\caption{Render-cost weight and cost form on five Technicolor~\cite{technicolor} scenes. The default is $\lambda_c=0.003$ with the cost $A_iT_i$.}
\label{tab:supp_costtradeoff}
\scriptsize
\setlength{\tabcolsep}{3pt}
\begin{tabular}{@{}l ccccc@{}}
\toprule
Setting & PSNR$\uparrow$ & SSIM$\uparrow$ & LPIPS$\downarrow$ & \makecell{Train\\time$\downarrow$} & FPS$\uparrow$ \\
\midrule
$\lambda_c=0.001$ & 31.68 & 0.900 & 0.103 & 27.2 & 875 \\
$\lambda_c=0.003$ (default) & 31.68 & 0.900 & 0.104 & 26.0 & 1033 \\
$\lambda_c=0.01$ & 31.59 & 0.900 & 0.104 & 24.5 & 1325 \\
$\lambda_c=0.03$ & 31.48 & 0.898 & 0.106 & 23.3 & 1591 \\
$\lambda_c=0.1$ & 31.40 & 0.897 & 0.108 & 22.7 & 1808 \\
\midrule
Cost $A_i$ only & 31.65 & 0.900 & 0.102 & 27.1 & 914 \\
Cost $\lVert\mathbf{s}_i\rVert_1$ & 31.58 & 0.900 & 0.102 & 27.2 & 824 \\
\bottomrule
\end{tabular}
\end{table}

\subsection{Feed-Forward Initialization}
\label{sec:supp_ffinit}

\paragraph{Sparse COLMAP initialization.}
Since dense COLMAP is slow, sparse COLMAP on the first frame may seem an alternative.
The number of sparse points, however, depends on the number of images.
Static scenes are captured with hundreds of images, and sparse SfM yields 32K--247K points per scene on Mip-NeRF 360\suppcite{mipnerf360}.
With the 16--21 cameras of our benchmarks, sparse SfM on the first frame yields only 3.6K--4.5K points on Neural 3D and 3.1K--16K on Technicolor.
Methods that initialize from a single frame therefore use dense COLMAP~\cite{4dgs,degauss,swift4d}.
Initializing DSD-GS from first-frame sparse COLMAP lowers the PSNR from 30.83 to 29.82\,dB and raises the LPIPS from 0.045 to 0.058 on Neural 3D.
Feed-forward initialization instead predicts a Gaussian for every pixel of the four input views, which gives more than 100K Gaussians on both benchmarks.

\paragraph{Unknown camera poses.}
When the camera poses are unknown, they can be estimated by COLMAP before the feed-forward initialization.
On the first frames of Neural 3D, the pose estimation with feature extraction, matching, and the mapper takes 14.4\,s on average.
Adding it to the 41\,s initialization of DSD-GS still keeps the initialization far below the 341\,s of dense COLMAP reconstruction.

\paragraph{Camera configuration and feed-forward depth.}
The feed-forward depth, however, is inaccurate on Technicolor.
Relocation moves 40\% of the feed-forward Gaussians by more than 15\% in depth on Technicolor, whereas it moves only 1.5\% on Neural 3D.
To find the cause, we compare the camera configurations.
The feed-forward model that we use, DepthSplat~\cite{depthsplat}, is trained on RealEstate10K~\cite{re10k} and DL3DV~\cite{dl3dv}, whose input views are frames sampled from handheld videos. %
\Cref{tab:supp_camconfig} compares the input views of the training data with those used for initialization on the two benchmarks.
The input views of Neural 3D lie within this distribution, with a wide field of view and optical axes that differ by more than 20$^\circ$.
The input views of Technicolor come from a planar camera array with nearly parallel optical axes and a narrow field of view, which occur in only 1.1\% and none of the training sets, respectively.

\begin{table}[h]
\centering
\caption{Camera configuration of DL3DV~\cite{dl3dv}, the training data of the feed-forward model, and of Neural 3D~\cite{dynerf} and Technicolor~\cite{technicolor}. For DL3DV, the median and the 10th to 90th percentiles are given. The max.\ axis angle is the maximum angle between the optical axes of the views, and the motion angle is the angle between the baseline of neighboring views and the mean optical axis, where 90$^\circ$ is lateral motion.}
\label{tab:supp_camconfig}
\small
\setlength{\tabcolsep}{4pt}
\begin{adjustbox}{max width=\linewidth}
\begin{tabular}{l ccc}
\toprule
 & Horizontal FoV & Max.\ axis angle & Motion angle \\
\midrule
DL3DV (training) & 96$^\circ$ (89--98) & 46$^\circ$ (13--112) & 65$^\circ$ (41--81) \\
Neural 3D & 86$^\circ$ & 22--28$^\circ$ & 88--89$^\circ$ \\
Technicolor & 45--51$^\circ$ & 1.3--1.7$^\circ$ & 89--90$^\circ$ \\
\bottomrule
\end{tabular}
\end{adjustbox}
\end{table}

\paragraph{Comparison with consistency filtering.}
MVSGaussian~\cite{mvsgaussian} also initializes with feed-forward Gaussian splatting and handles inaccurate depth by removing the Gaussians that are inconsistent across views.
With many views it keeps too many points, and with few views it removes too many (\cref{tab:supp_mvsgs}).
Since DepthSplat cannot take all 15 training views as input, we measure the dense setting with the network and official weights of MVSGaussian.
On all 15 training views of Technicolor, it keeps 2.9--5.9M points, 25 to 51 times as many as our 115K Gaussians.
With the four input views of our initialization, its consistency check applied to the DepthSplat depth removes 63--96\% of the Gaussians.
Multi-View Ray Relocation instead keeps every Gaussian and corrects its depth.
Training DSD-GS with the consistency check of MVSGaussian gives 28.27\,dB, above the 26.85\,dB without correction but below Multi-View Ray Relocation in every scene (\cref{tab:supp_mvsgs_psnr}).

\begin{table}[t]
\centering
\caption{Number of Gaussians on the first frame of Technicolor~\cite{technicolor} before and after the consistency check of MVSGaussian~\cite{mvsgaussian}. Ranges are over the five scenes. DS denotes DepthSplat~\cite{depthsplat}.}
\label{tab:supp_mvsgs}
\scriptsize
\setlength{\tabcolsep}{3pt}
\begin{tabular}{@{}l cc@{}}
\toprule
Method & Before (K) & After (K) \\
\midrule
DS + MVR & 115 & -- \\
DS + MVSGaussian check, 4 views & 115 & 5--42 \\
MVSGaussian, 15 views & 9216 & 2852--5874 \\
\bottomrule
\end{tabular}
\end{table}

\begin{table}[t]
\centering
\caption{PSNR of DSD-GS on Technicolor~\cite{technicolor} with different first-frame initializations. DS denotes DepthSplat~\cite{depthsplat}.}
\label{tab:supp_mvsgs_psnr}
\scriptsize
\setlength{\tabcolsep}{3pt}
\begin{tabular}{@{}l cccccc@{}}
\toprule
Initialization & Birthday & Fabien & Painter & Theater & Train & Avg. \\
\midrule
DS & 28.34 & 33.71 & 25.81 & 23.52 & 22.89 & 26.85 \\
DS + MVSGaussian check & 29.53 & 33.55 & 27.92 & 25.19 & 25.19 & 28.27 \\
DS + MVR & \textbf{30.22} & \textbf{34.39} & \textbf{34.54} & \textbf{29.02} & \textbf{30.23} & \textbf{31.68} \\
\bottomrule
\end{tabular}
\end{table}

\paragraph{Relocation for the baselines.}
To examine whether Multi-View Ray Relocation also benefits other methods, \cref{tab:supp_ffreloc} gives the baselines on Technicolor the same feed-forward initialization with and without Multi-View Ray Relocation, with all other settings unchanged.
Without Multi-View Ray Relocation, the PSNR of every baseline drops below that with COLMAP initialization, by 1.23--8.82\,dB.
With Multi-View Ray Relocation, the PSNR of all five baselines improves by 0.45--5.90\,dB.

\begin{table}[t]
\centering
\caption{Baselines on Technicolor~\cite{technicolor} with COLMAP initialization (SA or D0), feed-forward initialization (FF), and feed-forward initialization with Multi-View Ray Relocation (FF+MVR).}
\label{tab:supp_ffreloc}
\scriptsize
\setlength{\tabcolsep}{3pt}
\begin{tabular}{@{}l cc cc cc@{}}
\toprule
& \multicolumn{2}{c}{COLMAP} & \multicolumn{2}{c}{FF} & \multicolumn{2}{c}{FF+MVR} \\
\cmidrule(lr){2-3}\cmidrule(lr){4-5}\cmidrule(lr){6-7}
Method & PSNR$\uparrow$ & LPIPS$\downarrow$ & PSNR$\uparrow$ & LPIPS$\downarrow$ & PSNR$\uparrow$ & LPIPS$\downarrow$ \\
\midrule
STG~\cite{stg} & 31.83 & 0.104 & 23.01 & 0.297 & 28.91 & 0.146 \\
Taylor~\cite{taylor} & 31.33 & 0.128 & 27.13 & 0.201 & 28.69 & 0.192 \\
4DGS~\cite{4dgs} & 27.96 & 0.208 & 26.73 & 0.233 & 27.18 & 0.220 \\
Swift4D~\cite{swift4d} & 29.82 & 0.132 & 27.72 & 0.154 & 29.50 & 0.134 \\
DeGauss~\cite{degauss} & 23.14 & 0.180 & 20.21 & 0.271 & 22.34 & 0.188 \\
\bottomrule
\end{tabular}
\end{table}

\subsection{Temporally Varying Regions}
\label{sec:supp_region}
\begin{figure*}[t]
\centering
\includegraphics[width=\linewidth]{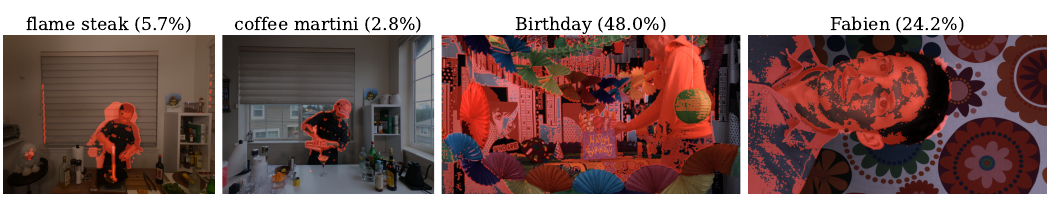}
\caption{Temporally varying regions (red) at $\tau=0.1$ on \emph{flame\_steak} and \emph{coffee\_martini} of Neural 3D~\cite{dynerf} and on \emph{Birthday} and \emph{Fabien} of Technicolor~\cite{technicolor}, with the fraction of pixels in the region. On Technicolor, lighting changes over the sequence enlarge the region beyond the moving objects.}
\label{fig:supp_region_mask}
\end{figure*}

Since image-level metrics are dominated by static regions, we evaluate the regions that change over time separately.
A pixel belongs to the temporally varying region at a frame if its ground-truth color differs from the temporal median of the ground truth at that pixel by more than a threshold $\tau$ in any channel, after dilating the region by 5 pixels,
\begin{equation}
\mathcal{R}_t = \mathrm{dilate}_5\Big(\big\{\mathbf{p} : \max_k \big|G_{t,k}(\mathbf{p})-\tilde{G}_k(\mathbf{p})\big| > \tau\big\}\Big),
\label{eq:supp_region}
\end{equation}
where $G_t$ is the ground-truth image at frame $t$ with values in $[0,1]$, $k$ indexes the color channels, and $\tilde{G}$ is the per-pixel temporal median of the ground truth.
\Cref{fig:supp_region_mask} shows examples.
The region is computed from the ground truth only and is shared by all methods.

\Cref{tab:supp_region} reports the metrics at $\tau=0.1$.
DSD-GS with either initialization ranks first or second in PSNR, SSIM, and LPIPS on both datasets.
With FF+MVR initialization, it improves the PSNR over STG by 2.35\,dB on Neural 3D and 0.50\,dB on Technicolor.
With $\tau=0.05$ and $\tau=0.2$, DSD-GS keeps the first and second places in PSNR, SSIM, and LPIPS on both datasets.

\begin{table*}[h]
\centering
\caption{Quality in temporally varying regions ($\tau=0.1$).}
\label{tab:supp_region}
\small
\setlength{\tabcolsep}{4pt}
\begin{tabular}{l c ccc c ccc}
\toprule
 & & \multicolumn{3}{c}{Neural 3D~\cite{dynerf}} & & \multicolumn{3}{c}{Technicolor~\cite{technicolor}} \\
\cmidrule{3-5}\cmidrule{7-9}
Method & Init. & PSNR$\uparrow$ & SSIM$\uparrow$ & LPIPS$\downarrow$ & & PSNR$\uparrow$ & SSIM$\uparrow$ & LPIPS$\downarrow$ \\
\midrule
STG~\cite{stg} & SA & 24.06 & 0.808 & 0.170 &  & \third{25.97} & \third{0.797} & \third{0.254} \\
Taylor~\cite{taylor} & SA & 23.62 & 0.806 & 0.171 &  & 25.85 & 0.783 & 0.291 \\
4DGS~\cite{4dgs} & D0 & \third{25.03} & \third{0.815} & 0.148 &  & 21.30 & 0.643 & 0.380 \\
Swift4D~\cite{swift4d} & D0 & 24.59 & \second{0.834} & \third{0.127} &  & 23.09 & 0.732 & 0.335 \\
DeGauss~\cite{degauss} & D0 & 23.10 & 0.760 & 0.218 &  & 15.06 & 0.535 & 0.453 \\
\midrule
Ours & D0 & \second{26.39} & \best{0.867} & \second{0.116} &  & \best{26.69} & \best{0.810} & \best{0.234} \\
Ours & FF+MVR & \best{26.41} & \best{0.867} & \best{0.110} &  & \second{26.47} & \second{0.803} & \second{0.243} \\
\bottomrule
\end{tabular}
\end{table*}

\section{Discussion on Static-Caching Rasterization}
\label{sec:supp_scr_discussion}

\subsection{Static-Caching Rasterization Requires Training-Time Consistency}
\label{sec:supp_scr_renderonly}
This section examines whether Static-Caching Rasterization can be applied orthogonally to other dynamic--static decomposition methods.
We find that it applies only to snapshot dynamic--static decomposition.

\paragraph{Rendering-time caching changes the depth order.}
First, the cache cannot be applied only at rendering time, because it changes the depth order of the Gaussians.
3DGS renders a pixel $\mathbf{p}$ by sorting the Gaussians that overlap its tile by depth and alpha-compositing them from front to back (\cref{alg:raster}) as
\begin{align}
C(\mathbf{p}) &= \sum_{i=1}^{N} \mathbf{c}_i\,\alpha_i \prod_{j=1}^{i-1} (1-\alpha_j), \label{eq:supp_composite}\\
\alpha_i &= \sigma_i \exp\!\Big(-\tfrac{1}{2}(\mathbf{p}-\boldsymbol{\mu}^{\prime}_i)^\top \Sigma_i^{\prime\,-1}(\mathbf{p}-\boldsymbol{\mu}^{\prime}_i)\Big), \label{eq:supp_alpha}
\end{align}
where $C(\mathbf{p})$ is the rendered color of pixel $\mathbf{p}$, $i$ indexes the $N$ Gaussians in front-to-back order, and $\alpha_i$ is the effective opacity of the $i$-th Gaussian at $\mathbf{p}$, given by its opacity $\sigma_i$ weighted by its projected 2D Gaussian with mean $\boldsymbol{\mu}^{\prime}_i$ and covariance $\Sigma^{\prime}_i$. $\mathbf{c}_i$ is the color of the $i$-th Gaussian.
The product is the transmittance $T$ remaining in front of the $i$-th Gaussian, so the contribution of each Gaussian depends on which Gaussians lie in front of it.
When static and dynamic Gaussians are rendered together, they are sorted jointly, so a static Gaussian can lie in front of a dynamic one and occlude it, and vice versa.
Static-Caching Rasterization instead composites the dynamic rendering over a cached static image (\cref{eq:scr}), which equals joint rendering only if, along every ray, every dynamic Gaussian lies in front of every static Gaussian.

\begin{algorithm}[h]
\caption{Tile-based rasterization of 3DGS for one pixel $\mathbf{p}$.}
\label{alg:raster}
\begin{algorithmic}[1]
\State $\mathcal{S} \gets$ Gaussians overlapping the tile of the pixel
\State Sort $\mathcal{S}$ by depth \Comment{static and dynamic Gaussians are sorted jointly}
\State $C \gets 0$, $T \gets 1$
\For{$g_i \in \mathcal{S}$ in front-to-back order}
  \State Compute $\alpha_i$ by \cref{eq:supp_alpha}
  \State $C \gets C + T\,\alpha_i\,\mathbf{c}_i$, \quad $T \gets T\,(1-\alpha_i)$
  \If{$T < \epsilon$} \textbf{break} \EndIf
\EndFor
\State \Return $C$
\end{algorithmic}
\end{algorithm}

\begin{figure}[h]
\centering
\includegraphics[width=\linewidth]{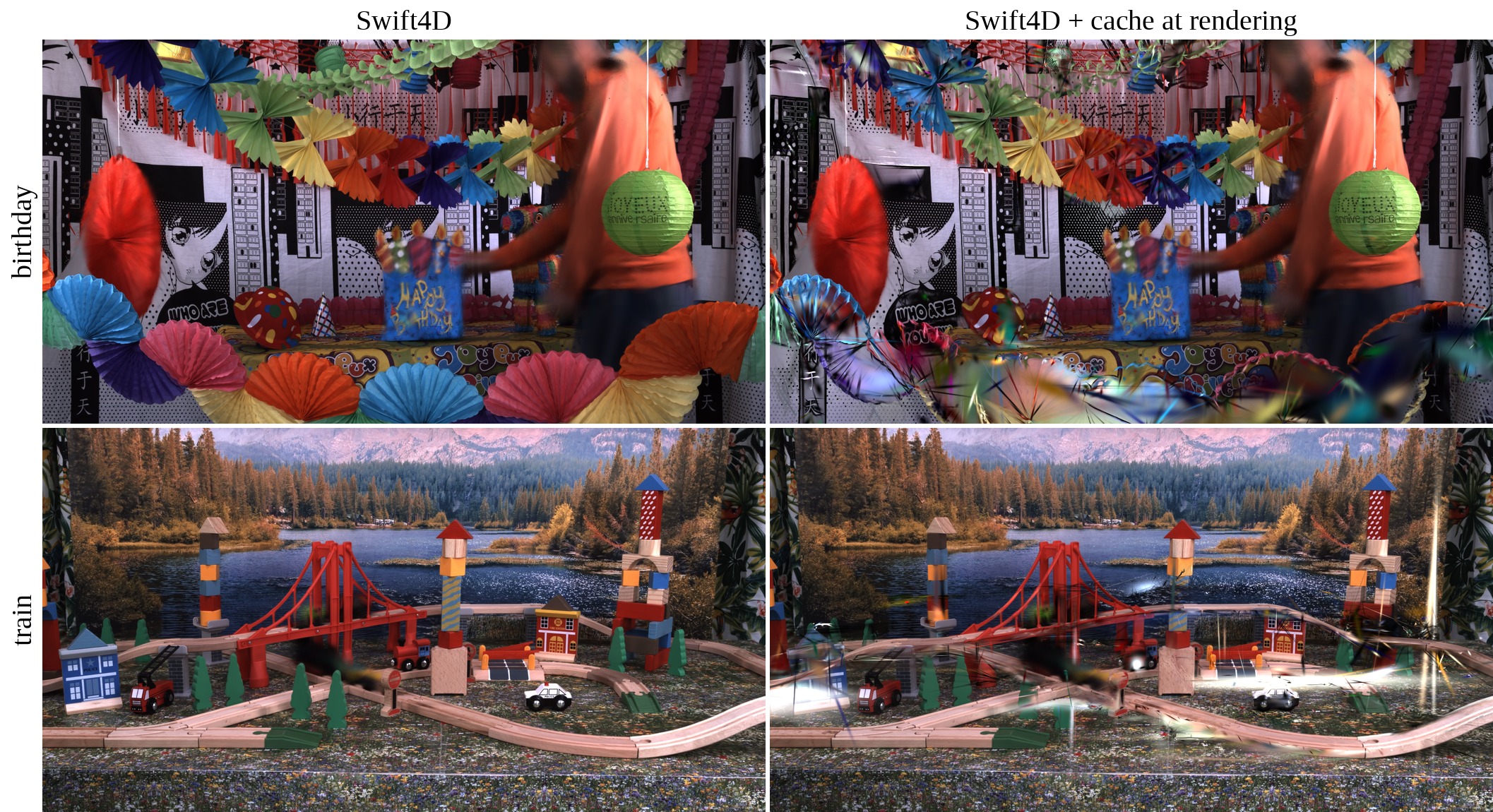}
\caption{Applying Static-Caching Rasterization only at rendering time to Swift4D~\cite{swift4d} on Technicolor~\cite{technicolor}, without retraining.}
\label{fig:supp_scr_renderonly}
\end{figure}

To verify this, we apply the cache only at rendering time to Swift4D~\cite{swift4d} and DeGauss~\cite{degauss} trained on Technicolor, without retraining, and evaluate both renderings with the same script.
As shown in \cref{tab:supp_scr_renderonly}, both methods lose more than 10\,dB.
In Swift4D, dynamic Gaussians cover most of the screen, and regions where static Gaussians are in front of them, such as the decorations in \emph{Birthday} and the tracks in \emph{Train}, are corrupted (\cref{fig:supp_scr_renderonly}).
DeGauss does not composite its foreground and background Gaussians in depth order but blends their renderings with a learned probabilistic mask, so replacing this composition with \cref{eq:scr} exposes the foreground Gaussians that the mask suppresses.
Caching only the background while keeping the mask-based composition preserves the quality but increases the rendering speed by only 13\%.
The background accounts for a small part of the rendering time, since the foreground Gaussians are rendered twice at every frame, once for the color and once for the mask.
Since applying Static-Caching Rasterization only at rendering time thus changes the rendered images, it must be applied from training onward, so that the dynamic Gaussians are optimized under the same composition.

\paragraph{Existing methods cannot cache during training.}
Applying it from training onward, however, requires the static Gaussians to stay fixed during training, which existing methods do not satisfy.
Their dynamic--static labels are determined during training, and their static Gaussians are updated throughout training, so the static rendering cannot be cached.
Swift4D keeps optimizing the static Gaussians even after fixing their labels, DeGauss optimizes the background Gaussians throughout training, and Ex4DGS assigns linear motion to static Gaussians.
In contrast, DSD-GS separates and freezes the static Gaussians in the snapshot stage and uses the same cache in training and rendering, so caching does not change the rendered images.

\begin{table}[h]
\centering
\caption{Quantitative results of applying Static-Caching Rasterization only at rendering time to existing decomposition methods on Technicolor~\cite{technicolor}, without retraining.}
\label{tab:supp_scr_renderonly}
\small
\setlength{\tabcolsep}{4pt}
\begin{adjustbox}{max width=\linewidth}
\begin{tabular}{llcccc}
\toprule
Method & Rendering & PSNR$\uparrow$ & SSIM$\uparrow$ & LPIPS$\downarrow$ & FPS$\uparrow$ \\
\midrule
Swift4D & original & 29.82 & 0.889 & 0.132 & 115 \\
Swift4D & cache & 17.47 & 0.748 & 0.289 & 159 \\
\midrule
DeGauss & original & 23.14 & 0.837 & 0.180 & 93 \\
DeGauss & background cache & 23.14 & 0.837 & 0.180 & 105 \\
DeGauss & cache & 12.96 & 0.447 & 0.715 & 239 \\
\bottomrule
\end{tabular}
\end{adjustbox}
\end{table}

\subsection{Effect on Dynamic Optimization}
\label{sec:supp_scr_optim}
Static-Caching Rasterization also benefits the optimization of the dynamic Gaussians.
When static and dynamic Gaussians are sorted jointly, a frozen static Gaussian can lie in front of a dynamic one and occlude it.
In training, it reduces the transmittance that reaches the dynamic Gaussian, which scales down its contribution and its gradient in \cref{eq:supp_composite}, so the dynamic content behind the frozen static Gaussians is learned poorly.
In rendering, it hides this content.
Static-Caching Rasterization composites the dynamic Gaussians over the cached static image, which prevents both.

To verify this, we measure the gradients of the dynamic Gaussians on \emph{Fabien} over all training views at 10 time steps (\cref{tab:supp_scr_grad}).
At the start of dynamic training, when the frozen static and dynamic Gaussians are sorted jointly, the dynamic Gaussians lying behind static surfaces receive per-Gaussian gradients of only 2\% of those of the other dynamic Gaussians for the position and motion and at most 33\% for the other attributes, whereas with Static-Caching Rasterization they receive at least 55\% of them.

Accordingly, replacing Static-Caching Rasterization with joint depth-sorted rendering of the frozen static and the dynamic Gaussians during dynamic training, with all other settings unchanged, lowers the average PSNR by 0.78\,dB (\cref{tab:ablation}).

\begin{table}[h]
\centering
\caption{Ratio of the mean gradient per dynamic Gaussian lying behind static surfaces to that per other dynamic Gaussian on \emph{Fabien} of Technicolor~\cite{technicolor}. A value of 1 means that the occluded Gaussians receive as much gradient as the others. SCR and Joint denote Static-Caching Rasterization and joint depth sorting. The trained model is the full model trained with Static-Caching Rasterization.}
\label{tab:supp_scr_grad}
\small
\setlength{\tabcolsep}{4pt}
\begin{tabular}{@{}lcccc@{}}
\toprule
 & \multicolumn{2}{c}{Start of training} & \multicolumn{2}{c}{Trained model} \\
\cmidrule(lr){2-3}\cmidrule(lr){4-5}
Attribute & SCR & Joint & SCR & Joint \\
\midrule
Position $\mathbf{m}_{i,0}$ & 0.58 & 0.02 & 0.16 & 0.06 \\
Motion $\mathbf{m}_{i,n\ge1}$ & 0.55 & 0.02 & 0.50 & 0.19 \\
Color & 3.75 & 0.19 & 0.40 & 0.14 \\
Opacity & 8.26 & 0.29 & 1.55 & 0.78 \\
Scale & 9.29 & 0.33 & 0.79 & 0.43 \\
Rotation $\mathbf{q}_{i,0}$ & 3.93 & 0.22 & 0.65 & 0.31 \\
Rotation velocity $\mathbf{q}_{i,1}$ & 3.74 & 0.20 & 1.11 & 0.39 \\
\bottomrule
\end{tabular}
\end{table}

\subsection{Free-Viewpoint Rendering}
\label{sec:supp_freecam}
Static-Caching Rasterization requires the rendering camera to stay fixed over time.
Multi-view dynamic scenes are typically captured with synchronized cameras at fixed positions.
All multi-view benchmarks of dynamic scene reconstruction follow this setting and evaluate on held-out fixed cameras (\cref{tab:supp_fixedcam}), and commercial volumetric capture stages likewise surround the subject with tens to over a hundred fixed cameras.
The training speedup therefore holds in the standard capture setting, and only rendering from a camera that moves over time loses the cache.

\begin{table}[t]
\centering
\caption{Multi-view dynamic scene datasets and commercial volumetric capture stages. All capture with synchronized cameras at fixed positions.}
\label{tab:supp_fixedcam}
\small
\setlength{\tabcolsep}{3pt}
\begin{adjustbox}{max width=\linewidth}
\begin{tabular}{@{}lcc@{}}
\toprule
Dataset & Cameras & Setting \\
\midrule
Neural 3D Video~\cite{dynerf} & 18--21 & Fixed \\
Technicolor~\cite{technicolor} & 16 & Fixed \\
Google Immersive~\cite{broxton2020immersive} & 46 & Fixed \\
Panoptic Studio~\cite{joo2015panoptic} & 31 HD, 480 VGA & Fixed \\
ActorsHQ~\cite{isik2023humanrf} & 160 & Fixed \\
DNA-Rendering~\cite{cheng2023dnarendering} & 60 & Fixed \\
\midrule
\multicolumn{3}{@{}l}{\textit{Commercial capture stages}} \\
Microsoft Mixed Reality Capture Studios & 106 & Fixed \\
Metastage & 106 & Fixed \\
Dimension Studio & 106 & Fixed \\
Fraunhofer HHI (Volucap) & 32 & Fixed \\
\bottomrule
\end{tabular}
\end{adjustbox}
\end{table}

When the rendering camera moves at every frame, the cache cannot be used, so the static Gaussians are rasterized at every frame and the dynamic Gaussians are composited over them in the same order as in Static-Caching Rasterization.
This produces the same images as rendering with the cache at the test views.
\Cref{tab:supp_freecam} reports the rendering speed in this setting.
Without the cache, the rendering speed of DSD-GS drops from 1033 to 189\,FPS on Technicolor and from 1926 to 323\,FPS on Neural 3D, which is comparable to STG on Technicolor and lower than STG on Neural 3D.
This drop arises because the full model applies Render-Cost Regularization only to the dynamic Gaussians.
Since the static Gaussians are rendered once into the cache, their rendering cost does not need to be reduced.
For free-viewpoint rendering, however, the regularization can also be applied to the static Gaussians.
Applying its area term to the static Gaussians during Snapshot Training with $\lambda_c=0.001$ raises the free-viewpoint rendering speed to 402 and 521\,FPS with only a marginal decrease in PSNR.
On a Jetson AGX Orin, it renders at 28.6 and 40.3\,FPS and with $\lambda_c=0.003$ at 31.7 and 46.5\,FPS, above 30\,FPS on average on both benchmarks.
No baseline reaches 30\,FPS on both benchmarks, and all of them stay below 16\,FPS on Technicolor (\cref{tab:main_technicolor,tab:main_dynerf}).

\begin{table}[t]
\centering
\caption{Free-viewpoint rendering, where the static Gaussians are rasterized at every frame. FPS is measured on an RTX 5090 GPU and a Jetson AGX Orin. Static reg.\ applies the area term of Render-Cost Regularization to the static Gaussians. STG is the fastest baseline (\cref{tab:main_technicolor,tab:main_dynerf}).}
\label{tab:supp_freecam}
\scriptsize
\setlength{\tabcolsep}{3pt}
\begin{adjustbox}{max width=\linewidth}
\begin{tabular}{l cccc cccc}
\toprule
& \multicolumn{4}{c}{Technicolor~\cite{technicolor}} & \multicolumn{4}{c}{Neural 3D~\cite{dynerf}} \\
\cmidrule(lr){2-5}\cmidrule(lr){6-9}
Method & PSNR$\uparrow$ & LPIPS$\downarrow$ & \makecell{FPS$\uparrow$\\(5090)} & \makecell{FPS$\uparrow$\\(Orin)} & PSNR$\uparrow$ & LPIPS$\downarrow$ & \makecell{FPS$\uparrow$\\(5090)} & \makecell{FPS$\uparrow$\\(Orin)} \\
\midrule
STG~\cite{stg} & 31.83 & 0.104 & 206 & 12.9 & 30.70 & 0.057 & 484 & 35.6 \\
\midrule
Ours, with cache (fixed view) & 31.68 & 0.104 & 1033 & 105.6 & 30.83 & 0.045 & 1926 & 280.0 \\
Ours, without cache & 31.68 & 0.104 & 189 & 12.7 & 30.83 & 0.045 & 323 & 22.8 \\
\quad + static reg., $\lambda_c=0.001$ & 31.46 & 0.103 & 402 & 28.6 & 30.78 & 0.046 & 521 & 40.3 \\
\quad + static reg., $\lambda_c=0.003$ & 31.48 & 0.103 & 425 & 31.7 & 30.61 & 0.047 & 575 & 46.5 \\
\bottomrule
\end{tabular}
\end{adjustbox}
\end{table}

\subsection{Static Occluders}
\label{sec:supp_static_occluder}
\begin{figure}[t]
\centering
\includegraphics[width=\linewidth]{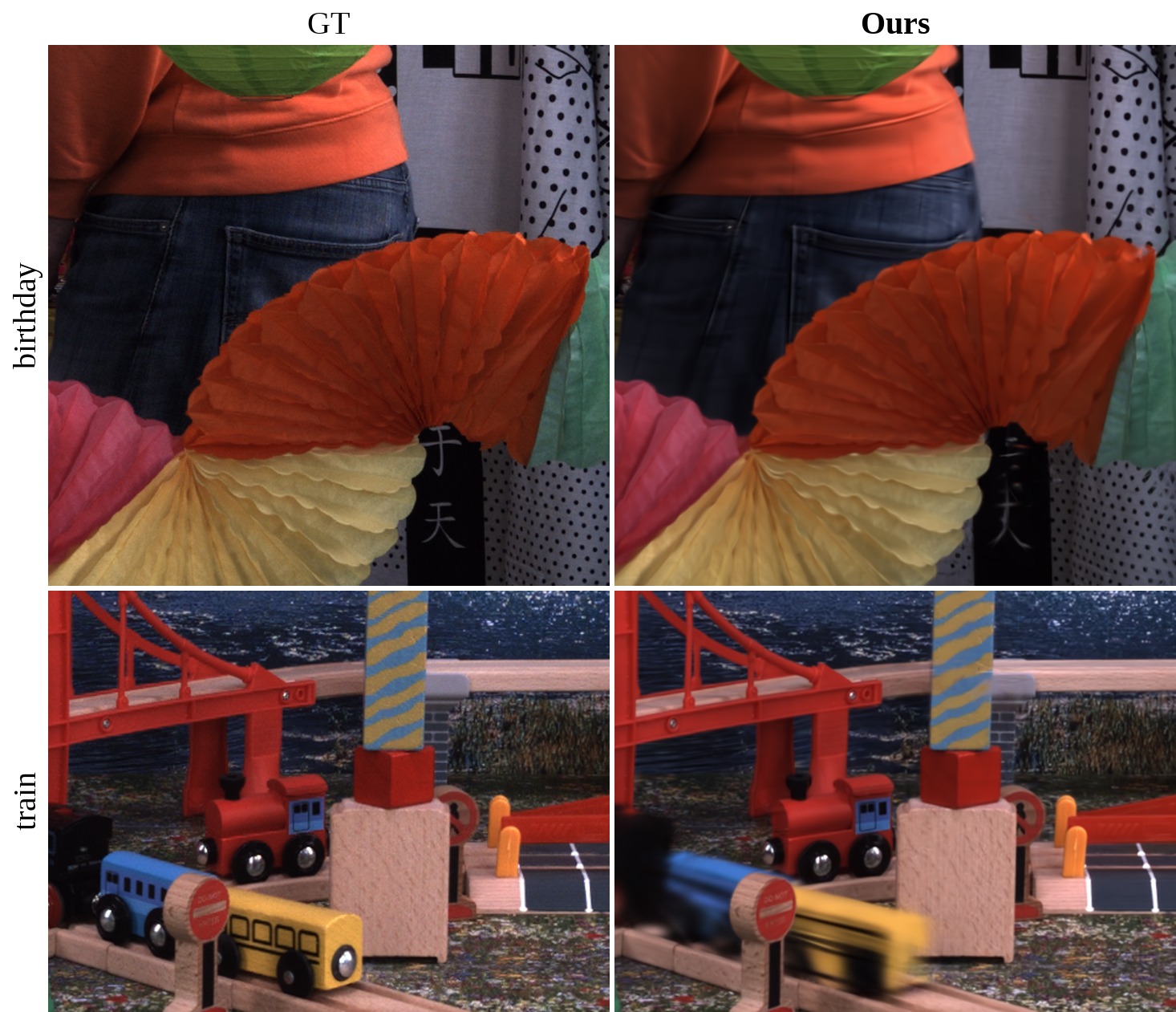}
\caption{Static occluders that occlude moving content in all training views, rendered by Static-Caching Rasterization. Top, paper fans in front of the person in \emph{Birthday}. Bottom, a signal in front of a toy train in \emph{Train}. Both scenes are from Technicolor~\cite{technicolor}.}
\label{fig:supp_static_occluder_ok}
\end{figure}

Static-Caching Rasterization composites the dynamic Gaussians over the cached static image, which assumes that no static surface lies in front of dynamic content.
In most cases, this assumption does not restrict the reconstruction.
Where a static object occludes moving content in the same way in all training views, such as the paper fans in front of the person in \emph{Birthday} or the trees and the signal in front of the toy trains in \emph{Train}, the optimization places no dynamic Gaussians in front of or behind it, so the occluder remains visible in the test view (\cref{fig:supp_static_occluder_ok}).

\begin{figure}[t]
\centering
\includegraphics[width=\linewidth]{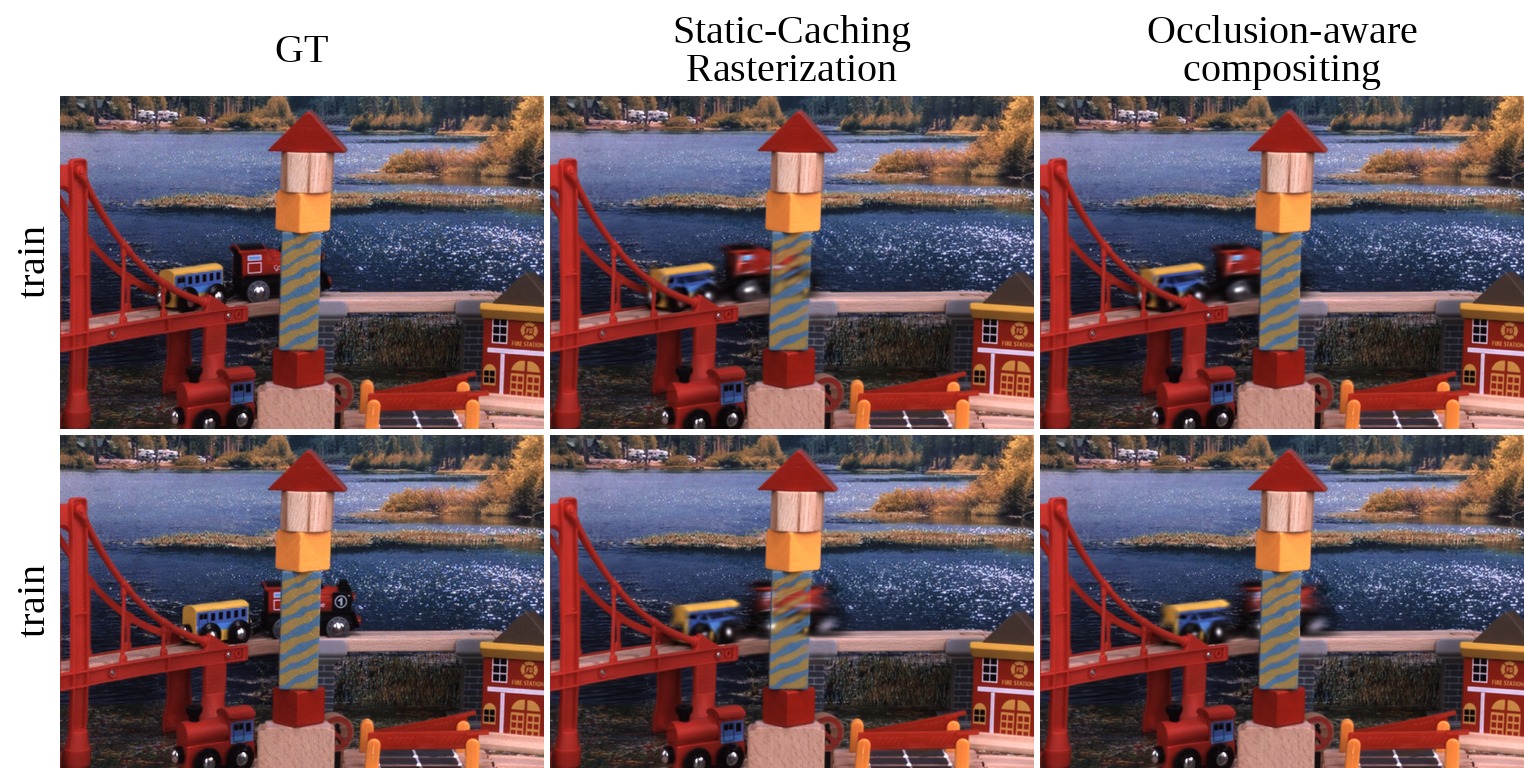}
\caption{A toy train passing behind a thin pillar in \emph{Train} of Technicolor~\cite{technicolor}.}
\label{fig:supp_static_occluder}
\end{figure}

This assumption breaks when a static occluder hides different parts of the moving content in different views.
In \emph{Train}, a toy train passes behind a pillar (\cref{fig:supp_static_occluder}).
Because the pillar is thin, neighboring training views see the part of the train that it hides in the test view.
Dynamic Gaussians are therefore rendered in this part.
In the test view, Static-Caching Rasterization composites these Gaussians over the cached image of the pillar, so the train appears in front of the pillar.
Among the five Technicolor scenes, we observe this case only in \emph{Train}.

\begin{figure}[t]
\centering
\includegraphics[width=\linewidth]{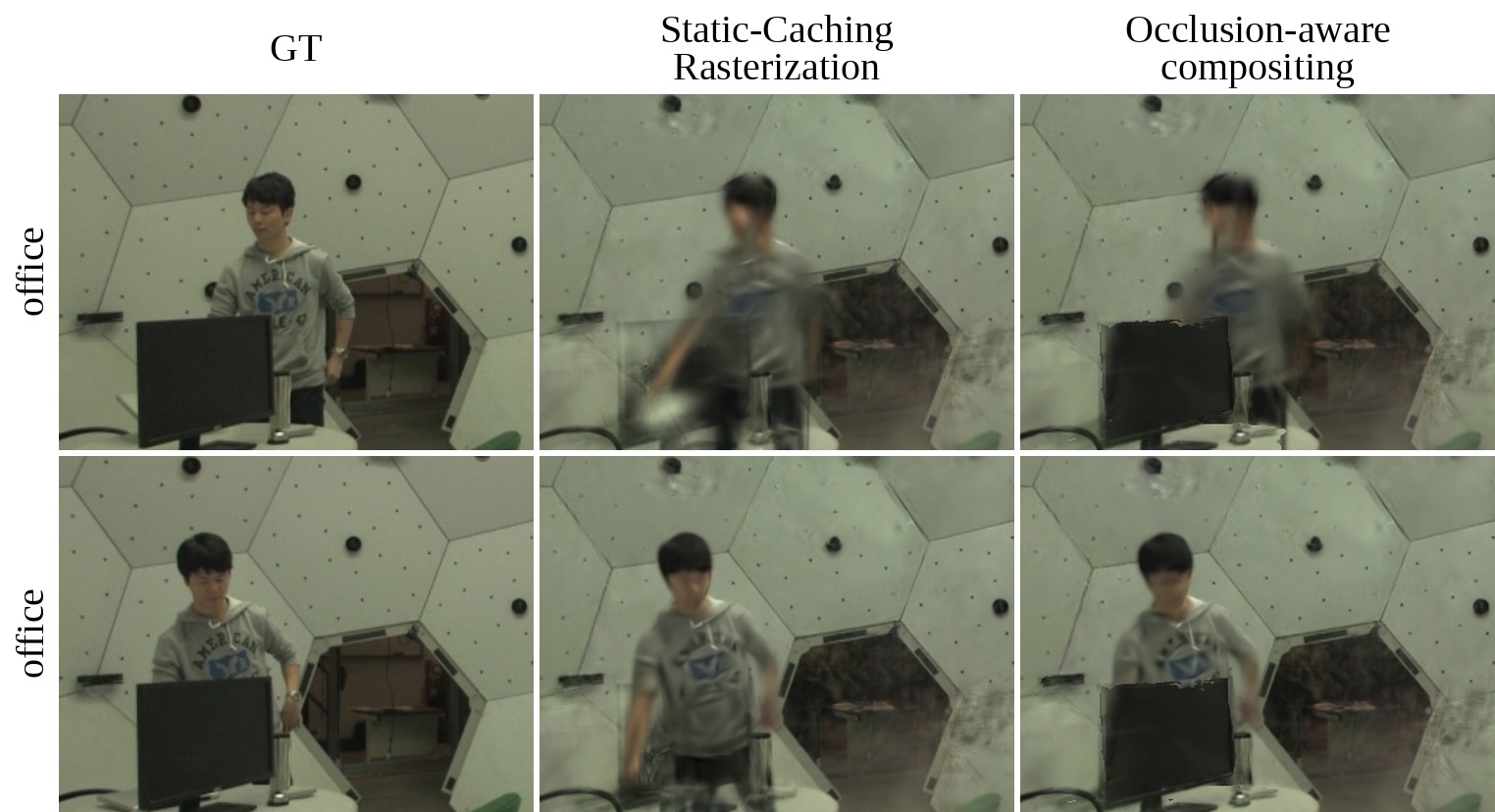}
\caption{A person walking behind a monitor in the Panoptic Studio~\cite{joo2015panoptic} \emph{office} sequence, seen from a held-out camera of the dome. Occlusion-aware compositing is also enabled during training.}
\label{fig:supp_static_occluder_360}
\end{figure}

\begin{figure*}[t]
\centering
\includegraphics[width=\linewidth]{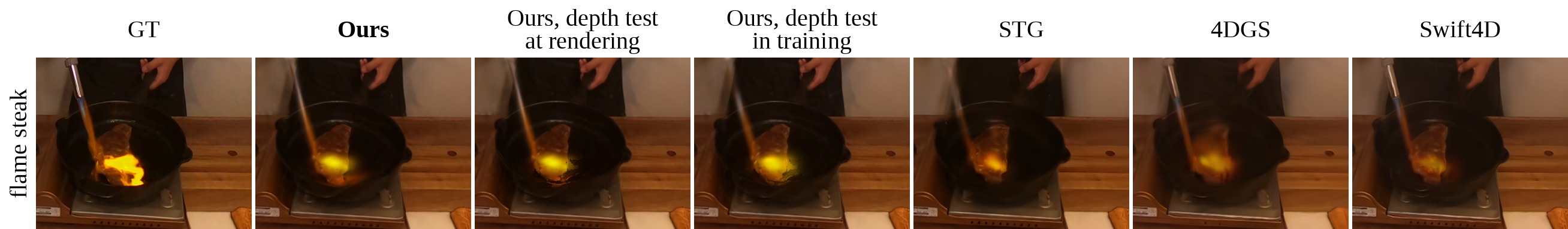}
\caption{Flame occluded by the static pan in \emph{flame steak} of Neural 3D~\cite{dynerf}. Static-Caching Rasterization keeps the static pan from occluding the dynamic flame. The depth test of \cref{sec:supp_static_occluder} lets the pan occlude the flame at rendering, and enabling it also in training reduces but does not remove this effect. The baselines, which sort the static and dynamic Gaussians jointly in training, render a weakened flame.}
\label{fig:supp_scr_flame}
\end{figure*}

\begin{table}[t]
\centering
\caption{Occlusion-aware compositing on Technicolor~\cite{technicolor} and the Panoptic Studio~\cite{joo2015panoptic} \emph{office} sequence. SCR and OAC denote Static-Caching Rasterization and occlusion-aware compositing. Occluded pixels are the pixels that stay static in the ground truth while moving content passes next to them, restricted to the pillar in \emph{Train} on Technicolor.}
\label{tab:supp_static_occluder}
\small
\setlength{\tabcolsep}{4pt}
\begin{adjustbox}{max width=\linewidth}
\begin{tabular}{@{}ll ccc@{}}
\toprule
 & & PSNR$\uparrow$ & \makecell{Occluded\\PSNR$\uparrow$} & FPS$\uparrow$ \\
\midrule
\multirow{3}{*}{Technicolor~\cite{technicolor}} & SCR & 31.68 & 33.06 & 1033 \\
 & OAC (render only) & 31.53 & 35.68 & 1005 \\
 & OAC (train and render) & 31.57 & 33.72 & 1111 \\
\midrule
\multirow{3}{*}{\emph{Office}~\cite{joo2015panoptic}} & SCR & 24.76 & 18.16 & 2641 \\
 & OAC (render only) & 27.03 & 22.24 & 2618 \\
 & OAC (train and render) & 27.14 & 23.55 & 2685 \\
\bottomrule
\end{tabular}
\end{adjustbox}
\end{table}

\paragraph{Occlusion-aware compositing.}
For such scenes, Static-Caching Rasterization can be extended with a per-pixel depth test.
The cache additionally stores the opacity-weighted depth of the static image, and the rasterizer accumulates the opacity-weighted depth of the dynamic Gaussians in the same pass.
Where the depth of the dynamic content lies more than 3\% behind the static depth, the static color is kept.
The test is effective even when applied only at rendering to a model trained with the plain Static-Caching Rasterization.

The test is applied only where the cached static depth is reliable.
It excludes pixels covered by dynamic content in the first frame, pixels whose static depth spreads along the ray, such as object boundaries, and static surfaces that do not stand in front of their surroundings, such as walls.

On \emph{Train}, the depth test restores the pillar (\cref{fig:supp_static_occluder}) and raises the PSNR of the occluded pixels from 33.06 to 35.68\,dB, while the average PSNR on Technicolor stays comparable (\cref{tab:supp_static_occluder}).
Over all scenes, enabling it slightly lowers the PSNR (\cref{tab:supp_static_occluder}).
This quality loss occurs where the depth test lets static Gaussians occlude dynamic ones, which Static-Caching Rasterization otherwise prevents (\cref{fig:supp_scr_flame}).
Enabling the depth test also during training reduces this artifact but does not remove it.

\paragraph{360$^\circ$ capture.}
In Technicolor and Neural 3D, the cameras face the scene from one side, so a static object and the moving content appear in the same depth order in all views.
When the cameras surround the scene, this order differs across views, since a static object lies in front of the moving content in some views and behind it in others.
The training views then also contain such occlusions, which the plain compositing cannot fit, so the depth test should be enabled already during training.
We evaluate this case on the \emph{office} sequence of the Panoptic Studio~\cite{joo2015panoptic}, where a person walks behind a desk with a monitor inside a camera dome.
We train on 27 HD cameras over 150 frames and test on a held-out camera in which the monitor occludes the person.
With the plain compositing, the person is drawn over the monitor, whereas the depth test keeps the monitor in front (\cref{fig:supp_static_occluder_360}).
The PSNR of the test view rises from 24.76 to 27.03\,dB, and that of the occluded pixels from 18.16 to 22.24\,dB (\cref{tab:supp_static_occluder}).
Enabling the depth test also during training accordingly raises them further to 27.14 and 23.55\,dB.

Static-Caching Rasterization prevents static Gaussians from occluding dynamic ones, which improves the reconstruction quality (\cref{sec:supp_scr_optim}).
When the cameras face the scene from one side, as in Technicolor and Neural 3D, this design is sufficient, and the depth test only gives up part of this benefit (31.53\,dB at rendering and 31.57\,dB also in training on Technicolor, compared with 31.68\,dB).
When the cameras surround the scene, this design no longer suffices, and the depth test is needed to keep static occluders in front of dynamic content.
We therefore report the main results with the plain Static-Caching Rasterization and provide the depth test as an option for 360$^\circ$ capture.

\section{Additional Qualitative Comparison}
\label{sec:supp_qual}
\begin{figure*}[t]
\centering
\includegraphics[width=\linewidth]{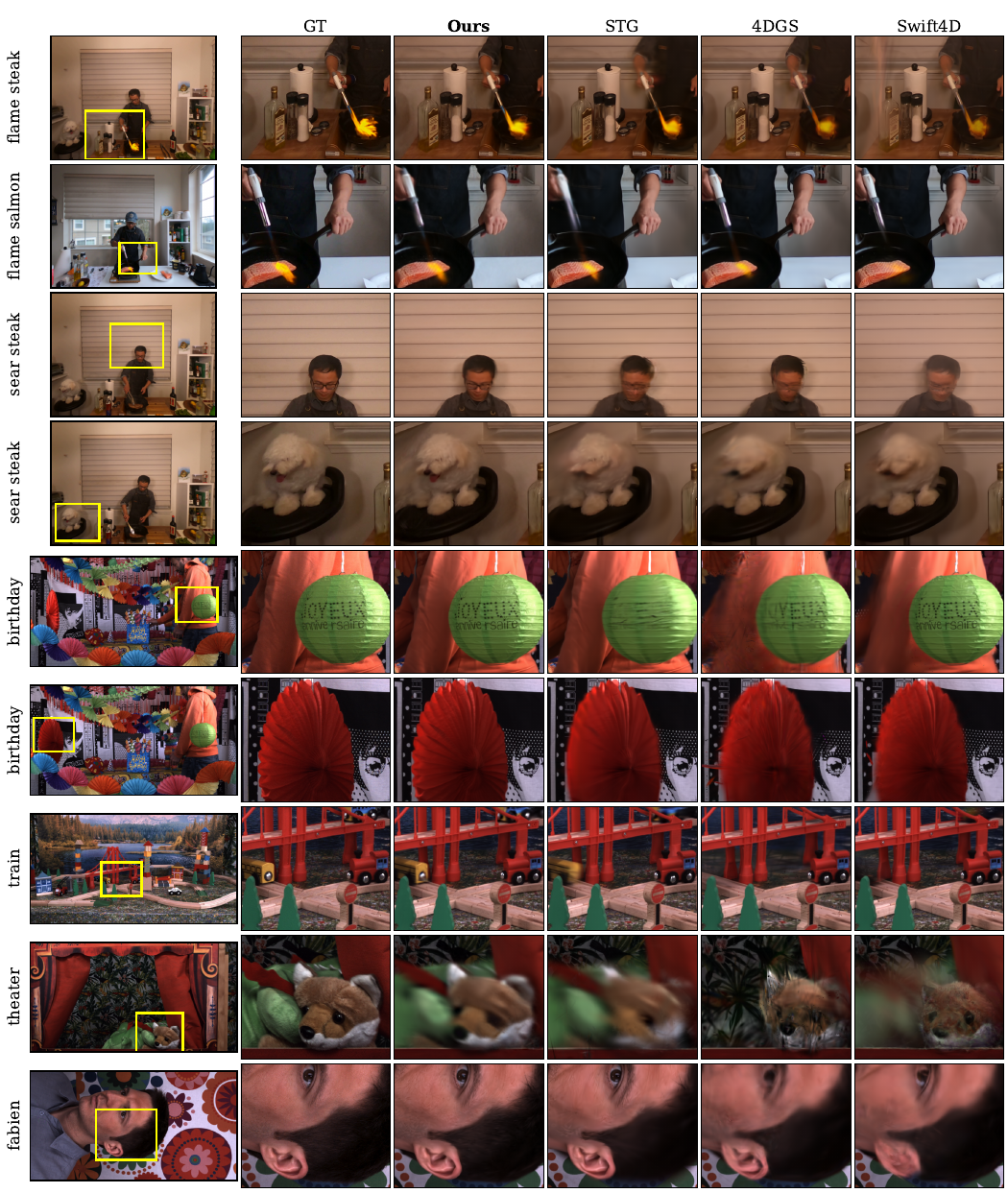}
\caption{Additional qualitative comparison with STG~\cite{stg}, 4DGS~\cite{4dgs}, and Swift4D~\cite{swift4d} on Neural 3D~\cite{dynerf} (top four rows) and Technicolor~\cite{technicolor} (bottom five rows). The yellow box in the first column marks the cropped region.}
\label{fig:qual_supp}
\end{figure*}

\Cref{fig:qual_supp} shows more scenes and frames in the same setting as \cref{fig:qual}.
On Neural 3D, the baselines blur the face of the cook and the dog in the background of \emph{sear\_steak}.
Swift4D produces floating streaks in front of the wall in \emph{flame\_steak}, and STG breaks the shape of the torch in \emph{flame\_salmon}.
DSD-GS reconstructs these regions more sharply.
On Technicolor, STG loses and 4DGS blurs the text on the lantern in \emph{Birthday}, and both distort the red fan in the same scene.
In \emph{Train}, 4DGS and Swift4D lose the yellow train, and STG blurs it.
In \emph{Theater}, STG blurs the moving puppet, and the background shows through it in 4DGS and Swift4D.
In \emph{Fabien}, 4DGS blurs the eye, and Swift4D blurs the ear.
DSD-GS reconstructs all of these objects, keeping the shape of the puppet although it is also blurred.

\section{Per-Scene Results}
\label{sec:supp_perscene}
This section gives the per-scene results of \cref{tab:main_technicolor,tab:main_dynerf} and of the temporally varying regions in \cref{tab:supp_region}, in \crefrange{tab:supp_perscene_Fabien}{tab:supp_perscene_Painter} for Technicolor and in \crefrange{tab:supp_perscene_coffee_martini}{tab:supp_perscene_sear_steak} for Neural 3D.
FF+MVR for the baselines uses the same relocated feed-forward Gaussians as ours on Technicolor (\cref{tab:supp_ffreloc}), and its initialization time adds the relocation of the first frame to that of FF. Ours with FF denotes the feed-forward initialization without Multi-View Ray Relocation, evaluated on Technicolor only (\cref{tab:ablation}).
FPS is measured on an RTX 5090 GPU.
\clearpage

\begin{table*}[p]
\centering
\caption{Per-scene results on \emph{Fabien} (Technicolor~\cite{technicolor}). SA and D0 denote sparse COLMAP on all frames and dense COLMAP on the first frame, FF the feed-forward initialization, and MVR Multi-View Ray Relocation. TV denotes the temporally varying region at $\tau=0.1$ (\cref{sec:supp_region}). Times are in minutes.}
\label{tab:supp_perscene_Fabien}
\small
\setlength{\tabcolsep}{4pt}
\begin{tabular}{@{}ll cc cc ccc c}
\toprule
 & & \multicolumn{2}{c}{Overall} & \multicolumn{2}{c}{TV} & & & & \\
\cmidrule(lr){3-4}\cmidrule(lr){5-6}
Method & Init. & PSNR$\uparrow$ & LPIPS$\downarrow$ & PSNR$\uparrow$ & LPIPS$\downarrow$ & \makecell{Init.\\time$\downarrow$} & \makecell{Train\\time$\downarrow$} & \makecell{Total\\time$\downarrow$} & FPS$\uparrow$ \\
\midrule
STG~\cite{stg} & SA & 33.65 & 0.157 & 30.81 & 0.251 & 8.3 & 56.7 & 65.0 & 220.3 \\
Taylor~\cite{taylor} & SA & 32.27 & 0.204 & 29.43 & 0.350 & 8.3 & 77.6 & 85.9 & 96.3 \\
4DGS~\cite{4dgs} & D0 & 31.84 & 0.216 & 28.62 & 0.373 & 5.4 & 58.0 & 63.4 & 183.0 \\
Swift4D~\cite{swift4d} & D0 & 32.18 & 0.209 & 28.90 & 0.361 & 8.2 & 37.8 & 46.0 & 134.8 \\
DeGauss~\cite{degauss} & D0 & 20.39 & 0.282 & 16.47 & 0.516 & 5.4 & 170.8 & 176.2 & 59.0 \\
\midrule
STG~\cite{stg} & FF & 30.79 & 0.210 & 28.33 & 0.352 & 0.1 & 53.2 & 53.3 & 197.0 \\
Taylor~\cite{taylor} & FF & 27.45 & 0.310 & 26.80 & 0.436 & 0.1 & 69.1 & 69.2 & 114.9 \\
4DGS~\cite{4dgs} & FF & 30.90 & 0.215 & 27.37 & 0.371 & 0.1 & 63.8 & 63.9 & 133.5 \\
Swift4D~\cite{swift4d} & FF & 31.46 & 0.212 & 28.03 & 0.368 & 3.1 & 39.3 & 42.4 & 118.7 \\
DeGauss~\cite{degauss} & FF & 20.17 & 0.291 & 16.33 & 0.528 & 0.1 & 178.7 & 178.8 & 78.7 \\
\midrule
STG~\cite{stg} & FF+MVR & 32.11 & 0.204 & 28.86 & 0.346 & 0.1 & 54.4 & 54.5 & 203.8 \\
Taylor~\cite{taylor} & FF+MVR & 23.60 & 0.413 & 24.07 & 0.493 & 0.1 & 70.1 & 70.2 & 103.2 \\
4DGS~\cite{4dgs} & FF+MVR & 30.63 & 0.222 & 27.09 & 0.388 & 0.1 & 43.4 & 43.5 & 155.0 \\
Swift4D~\cite{swift4d} & FF+MVR & 31.37 & 0.212 & 27.94 & 0.370 & 3.1 & \second{25.9} & \best{29.0} & 161.3 \\
DeGauss~\cite{degauss} & FF+MVR & 20.12 & 0.289 & 16.24 & 0.532 & 0.1 & 181.5 & 181.6 & 79.8 \\
\midrule
Ours & D0 & \best{34.77} & \best{0.141} & \third{32.37} & \best{0.231} & 6.0 & \best{24.3} & \second{30.3} & \best{883.0} \\
Ours & FF & \third{33.71} & \third{0.153} & \second{32.43} & \third{0.236} & 0.3 & \third{33.8} & \third{34.2} & \second{784.4} \\
Ours & FF+MVR & \second{34.39} & \second{0.149} & \best{32.46} & \second{0.233} & 0.4 & 35.3 & 35.7 & \third{721.2} \\
\bottomrule
\end{tabular}
\end{table*}

\begin{table*}[p]
\centering
\caption{Per-scene results on \emph{Birthday} (Technicolor~\cite{technicolor}). SA and D0 denote sparse COLMAP on all frames and dense COLMAP on the first frame, FF the feed-forward initialization, and MVR Multi-View Ray Relocation. TV denotes the temporally varying region at $\tau=0.1$ (\cref{sec:supp_region}). Times are in minutes.}
\label{tab:supp_perscene_Birthday}
\small
\setlength{\tabcolsep}{4pt}
\begin{tabular}{@{}ll cc cc ccc c}
\toprule
 & & \multicolumn{2}{c}{Overall} & \multicolumn{2}{c}{TV} & & & & \\
\cmidrule(lr){3-4}\cmidrule(lr){5-6}
Method & Init. & PSNR$\uparrow$ & LPIPS$\downarrow$ & PSNR$\uparrow$ & LPIPS$\downarrow$ & \makecell{Init.\\time$\downarrow$} & \makecell{Train\\time$\downarrow$} & \makecell{Total\\time$\downarrow$} & FPS$\uparrow$ \\
\midrule
STG~\cite{stg} & SA & 29.41 & 0.067 & 26.76 & 0.090 & 18.9 & 58.0 & 76.9 & 145.0 \\
Taylor~\cite{taylor} & SA & \third{30.10} & \third{0.065} & \third{27.52} & \third{0.082} & 18.9 & 203.9 & 222.8 & 27.7 \\
4DGS~\cite{4dgs} & D0 & 25.10 & 0.161 & 22.08 & 0.210 & 5.4 & 75.1 & 80.5 & 58.6 \\
Swift4D~\cite{swift4d} & D0 & 26.96 & 0.086 & 23.71 & 0.130 & 10.6 & 44.5 & 55.1 & 74.0 \\
DeGauss~\cite{degauss} & D0 & 18.63 & 0.195 & 15.31 & 0.311 & 5.4 & 185.3 & 190.7 & 96.5 \\
\midrule
STG~\cite{stg} & FF & 20.32 & 0.351 & 18.48 & 0.372 & 0.1 & 49.7 & 49.8 & 212.0 \\
Taylor~\cite{taylor} & FF & 27.56 & 0.131 & 25.08 & 0.162 & 0.1 & 183.4 & 183.5 & 36.2 \\
4DGS~\cite{4dgs} & FF & 23.88 & 0.200 & 20.99 & 0.252 & 0.1 & 76.4 & 76.5 & 61.9 \\
Swift4D~\cite{swift4d} & FF & 26.95 & 0.094 & 23.88 & 0.131 & 5.5 & 48.8 & 54.3 & 61.4 \\
DeGauss~\cite{degauss} & FF & 17.44 & 0.292 & 14.68 & 0.373 & 0.1 & 170.0 & 170.1 & 97.6 \\
\midrule
STG~\cite{stg} & FF+MVR & 25.56 & 0.118 & 22.39 & 0.173 & 0.1 & 53.2 & 53.3 & 167.0 \\
Taylor~\cite{taylor} & FF+MVR & 27.40 & 0.142 & 24.90 & 0.165 & 0.1 & 214.8 & 214.9 & 47.2 \\
4DGS~\cite{4dgs} & FF+MVR & 25.17 & 0.164 & 22.13 & 0.211 & 0.1 & 53.1 & 53.2 & 70.6 \\
Swift4D~\cite{swift4d} & FF+MVR & 27.03 & 0.089 & 23.84 & 0.132 & 5.5 & \best{34.6} & \best{40.1} & 71.7 \\
DeGauss~\cite{degauss} & FF+MVR & 18.93 & 0.199 & 15.72 & 0.314 & 0.1 & 188.8 & 188.9 & 101.2 \\
\midrule
Ours & D0 & \best{30.54} & \best{0.055} & \best{27.88} & \second{0.073} & 6.3 & 46.4 & 52.6 & \third{420.8} \\
Ours & FF & 28.34 & 0.078 & 25.65 & 0.101 & 0.6 & \third{43.6} & \third{44.2} & \second{450.8} \\
Ours & FF+MVR & \second{30.22} & \second{0.056} & \second{27.56} & \best{0.070} & 0.6 & \second{42.5} & \second{43.1} & \best{451.4} \\
\bottomrule
\end{tabular}
\end{table*}

\begin{table*}[p]
\centering
\caption{Per-scene results on \emph{Theater} (Technicolor~\cite{technicolor}). SA and D0 denote sparse COLMAP on all frames and dense COLMAP on the first frame, FF the feed-forward initialization, and MVR Multi-View Ray Relocation. TV denotes the temporally varying region at $\tau=0.1$ (\cref{sec:supp_region}). Times are in minutes.}
\label{tab:supp_perscene_Theater}
\small
\setlength{\tabcolsep}{4pt}
\begin{tabular}{@{}ll cc cc ccc c}
\toprule
 & & \multicolumn{2}{c}{Overall} & \multicolumn{2}{c}{TV} & & & & \\
\cmidrule(lr){3-4}\cmidrule(lr){5-6}
Method & Init. & PSNR$\uparrow$ & LPIPS$\downarrow$ & PSNR$\uparrow$ & LPIPS$\downarrow$ & \makecell{Init.\\time$\downarrow$} & \makecell{Train\\time$\downarrow$} & \makecell{Total\\time$\downarrow$} & FPS$\uparrow$ \\
\midrule
STG~\cite{stg} & SA & \second{29.90} & \second{0.146} & \second{23.71} & \third{0.319} & 25.4 & 62.6 & 88.0 & 222.6 \\
Taylor~\cite{taylor} & SA & 28.58 & 0.215 & 23.06 & 0.391 & 25.4 & 133.2 & 158.6 & 43.7 \\
4DGS~\cite{4dgs} & D0 & 26.56 & 0.225 & 19.04 & 0.432 & 5.2 & 65.9 & 71.1 & 96.8 \\
Swift4D~\cite{swift4d} & D0 & 28.51 & 0.169 & 21.68 & 0.373 & 11.7 & 39.5 & 51.2 & 118.9 \\
DeGauss~\cite{degauss} & D0 & 23.58 & 0.193 & 15.27 & 0.486 & 5.2 & 173.7 & 178.9 & 102.4 \\
\midrule
STG~\cite{stg} & FF & 22.31 & 0.310 & 19.91 & 0.473 & 0.1 & 51.6 & 51.7 & 268.2 \\
Taylor~\cite{taylor} & FF & 26.97 & 0.273 & 22.97 & 0.410 & 0.1 & 91.3 & 91.4 & 62.9 \\
4DGS~\cite{4dgs} & FF & 25.14 & 0.263 & 18.30 & 0.466 & 0.1 & 66.8 & 66.9 & 102.7 \\
Swift4D~\cite{swift4d} & FF & 26.19 & 0.201 & 21.26 & 0.394 & 6.5 & 41.1 & 47.6 & 106.8 \\
DeGauss~\cite{degauss} & FF & 21.46 & 0.306 & 15.81 & 0.536 & 0.1 & 183.2 & 183.3 & 100.6 \\
\midrule
STG~\cite{stg} & FF+MVR & 27.08 & 0.183 & 20.56 & 0.398 & 0.1 & 55.0 & 55.1 & 245.6 \\
Taylor~\cite{taylor} & FF+MVR & 28.54 & 0.220 & 22.71 & 0.401 & 0.1 & 86.3 & 86.4 & 56.0 \\
4DGS~\cite{4dgs} & FF+MVR & 25.11 & 0.253 & 17.77 & 0.464 & 0.1 & 48.2 & 48.3 & 93.0 \\
Swift4D~\cite{swift4d} & FF+MVR & 28.03 & 0.171 & 21.11 & 0.378 & 6.5 & 28.2 & 34.7 & 118.0 \\
DeGauss~\cite{degauss} & FF+MVR & 23.50 & 0.203 & 15.52 & 0.487 & 0.1 & 187.1 & 187.2 & 98.2 \\
\midrule
Ours & D0 & \best{30.11} & \best{0.119} & \best{23.89} & \best{0.290} & 6.2 & \third{19.3} & \third{25.5} & \third{958.0} \\
Ours & FF & 23.52 & 0.264 & 22.88 & 0.364 & 0.6 & \best{16.7} & \best{17.3} & \second{1081.6} \\
Ours & FF+MVR & \third{29.02} & \third{0.152} & \third{23.62} & \second{0.311} & 0.7 & \second{16.9} & \second{17.6} & \best{1091.0} \\
\bottomrule
\end{tabular}
\end{table*}

\begin{table*}[p]
\centering
\caption{Per-scene results on \emph{Train} (Technicolor~\cite{technicolor}). SA and D0 denote sparse COLMAP on all frames and dense COLMAP on the first frame, FF the feed-forward initialization, and MVR Multi-View Ray Relocation. TV denotes the temporally varying region at $\tau=0.1$ (\cref{sec:supp_region}). Times are in minutes.}
\label{tab:supp_perscene_Train}
\small
\setlength{\tabcolsep}{4pt}
\begin{tabular}{@{}ll cc cc ccc c}
\toprule
 & & \multicolumn{2}{c}{Overall} & \multicolumn{2}{c}{TV} & & & & \\
\cmidrule(lr){3-4}\cmidrule(lr){5-6}
Method & Init. & PSNR$\uparrow$ & LPIPS$\downarrow$ & PSNR$\uparrow$ & LPIPS$\downarrow$ & \makecell{Init.\\time$\downarrow$} & \makecell{Train\\time$\downarrow$} & \makecell{Total\\time$\downarrow$} & FPS$\uparrow$ \\
\midrule
STG~\cite{stg} & SA & \best{31.12} & \best{0.043} & 19.86 & 0.348 & 22.3 & 62.5 & 84.8 & 264.3 \\
Taylor~\cite{taylor} & SA & \third{30.71} & \best{0.043} & 19.69 & 0.354 & 22.3 & 516.9 & 539.2 & 29.1 \\
4DGS~\cite{4dgs} & D0 & 25.03 & 0.258 & 13.70 & 0.479 & 5.3 & 64.6 & 69.9 & 89.5 \\
Swift4D~\cite{swift4d} & D0 & 28.77 & 0.070 & 15.57 & 0.465 & 10.9 & 38.4 & 49.3 & 103.0 \\
DeGauss~\cite{degauss} & D0 & 26.25 & 0.077 & 10.65 & 0.466 & 5.3 & 190.3 & 195.6 & 102.1 \\
\midrule
STG~\cite{stg} & FF & 18.82 & 0.220 & 13.94 & 0.495 & 0.1 & 48.6 & 48.7 & 243.6 \\
Taylor~\cite{taylor} & FF & 20.83 & 0.132 & 16.06 & 0.455 & 0.1 & 177.3 & 177.4 & 60.6 \\
4DGS~\cite{4dgs} & FF & 23.93 & 0.272 & 13.37 & 0.479 & 0.1 & 68.0 & 68.1 & 77.8 \\
Swift4D~\cite{swift4d} & FF & 23.03 & 0.117 & 15.19 & 0.459 & 5.6 & 41.2 & 46.8 & 84.9 \\
DeGauss~\cite{degauss} & FF & 20.51 & 0.168 & 10.67 & 0.469 & 0.1 & 167.2 & 167.3 & 103.0 \\
\midrule
STG~\cite{stg} & FF+MVR & 27.49 & 0.087 & 16.71 & 0.443 & 0.1 & 54.4 & 54.5 & 227.6 \\
Taylor~\cite{taylor} & FF+MVR & 29.60 & \third{0.052} & 17.50 & 0.439 & 0.1 & 201.2 & 201.3 & 51.4 \\
4DGS~\cite{4dgs} & FF+MVR & 23.98 & 0.278 & 12.27 & 0.486 & 0.1 & 48.8 & 48.9 & 78.7 \\
Swift4D~\cite{swift4d} & FF+MVR & 28.50 & 0.068 & 15.12 & 0.456 & 5.6 & 28.4 & 34.0 & 105.8 \\
DeGauss~\cite{degauss} & FF+MVR & 26.06 & 0.086 & 10.68 & 0.467 & 0.1 & 181.8 & 181.9 & 103.1 \\
\midrule
Ours & D0 & \second{30.90} & \best{0.043} & \best{21.29} & \best{0.293} & 6.2 & \second{13.7} & \third{19.9} & \third{1687.3} \\
Ours & FF & 22.89 & 0.109 & \third{20.12} & \third{0.345} & 0.5 & \best{13.2} & \best{13.7} & \best{2228.4} \\
Ours & FF+MVR & 30.23 & \second{0.051} & \second{20.38} & \second{0.331} & 0.6 & \third{14.8} & \second{15.3} & \second{1788.2} \\
\bottomrule
\end{tabular}
\end{table*}

\begin{table*}[p]
\centering
\caption{Per-scene results on \emph{Painter} (Technicolor~\cite{technicolor}). SA and D0 denote sparse COLMAP on all frames and dense COLMAP on the first frame, FF the feed-forward initialization, and MVR Multi-View Ray Relocation. TV denotes the temporally varying region at $\tau=0.1$ (\cref{sec:supp_region}). Times are in minutes.}
\label{tab:supp_perscene_Painter}
\small
\setlength{\tabcolsep}{4pt}
\begin{tabular}{@{}ll cc cc ccc c}
\toprule
 & & \multicolumn{2}{c}{Overall} & \multicolumn{2}{c}{TV} & & & & \\
\cmidrule(lr){3-4}\cmidrule(lr){5-6}
Method & Init. & PSNR$\uparrow$ & LPIPS$\downarrow$ & PSNR$\uparrow$ & LPIPS$\downarrow$ & \makecell{Init.\\time$\downarrow$} & \makecell{Train\\time$\downarrow$} & \makecell{Total\\time$\downarrow$} & FPS$\uparrow$ \\
\midrule
STG~\cite{stg} & SA & \best{35.06} & \best{0.106} & \third{28.69} & \best{0.264} & 25.1 & 62.1 & 87.2 & 177.0 \\
Taylor~\cite{taylor} & SA & \second{34.98} & 0.113 & \best{29.55} & \third{0.277} & 25.1 & 160.9 & 186.0 & 53.0 \\
4DGS~\cite{4dgs} & D0 & 31.26 & 0.178 & 23.04 & 0.405 & 5.3 & 61.7 & 67.0 & 109.2 \\
Swift4D~\cite{swift4d} & D0 & 32.70 & 0.128 & 25.61 & 0.344 & 11.7 & 36.6 & 48.3 & 143.0 \\
DeGauss~\cite{degauss} & D0 & 26.86 & 0.155 & 17.62 & 0.487 & 5.3 & 171.6 & 176.9 & 103.4 \\
\midrule
STG~\cite{stg} & FF & 22.79 & 0.393 & 20.39 & 0.549 & 0.1 & 45.7 & 45.8 & 267.5 \\
Taylor~\cite{taylor} & FF & 32.86 & 0.157 & 27.22 & 0.348 & 0.1 & 116.3 & 116.4 & 73.0 \\
4DGS~\cite{4dgs} & FF & 29.80 & 0.216 & 22.55 & 0.435 & 0.1 & 68.9 & 69.0 & 79.7 \\
Swift4D~\cite{swift4d} & FF & 30.96 & 0.146 & 24.06 & 0.371 & 6.5 & 40.0 & 46.5 & 101.0 \\
DeGauss~\cite{degauss} & FF & 21.47 & 0.298 & 16.38 & 0.582 & 0.1 & 181.9 & 182.0 & 99.8 \\
\midrule
STG~\cite{stg} & FF+MVR & 32.29 & 0.140 & 25.06 & 0.367 & 0.1 & 51.6 & 51.7 & 211.9 \\
Taylor~\cite{taylor} & FF+MVR & 34.29 & 0.132 & \second{29.18} & 0.306 & 0.1 & 109.6 & 109.7 & 91.7 \\
4DGS~\cite{4dgs} & FF+MVR & 30.99 & 0.182 & 22.80 & 0.409 & 0.1 & 46.0 & 46.1 & 99.3 \\
Swift4D~\cite{swift4d} & FF+MVR & 32.58 & 0.131 & 25.58 & 0.343 & 6.5 & 26.2 & 32.7 & 139.5 \\
DeGauss~\cite{degauss} & FF+MVR & 23.10 & 0.162 & 16.12 & 0.507 & 0.1 & 182.0 & 182.1 & 104.9 \\
\midrule
Ours & D0 & 34.37 & \second{0.108} & 28.03 & 0.283 & 6.3 & \best{18.1} & \third{24.4} & \third{1011.8} \\
Ours & FF & 25.81 & 0.211 & 24.38 & 0.379 & 0.5 & \second{18.9} & \best{19.4} & \second{1085.9} \\
Ours & FF+MVR & \third{34.54} & \third{0.109} & 28.31 & \second{0.269} & 0.6 & \third{20.4} & \second{21.0} & \best{1112.2} \\
\bottomrule
\end{tabular}
\end{table*}

\begin{table*}[p]
\centering
\caption{Per-scene results on \emph{coffee\_martini} (Neural 3D~\cite{dynerf}). SA and D0 denote sparse COLMAP on all frames and dense COLMAP on the first frame, FF the feed-forward initialization, and MVR Multi-View Ray Relocation. TV denotes the temporally varying region at $\tau=0.1$ (\cref{sec:supp_region}). Times are in minutes.}
\label{tab:supp_perscene_coffee_martini}
\small
\setlength{\tabcolsep}{4pt}
\begin{tabular}{@{}ll cc cc ccc c}
\toprule
 & & \multicolumn{2}{c}{Overall} & \multicolumn{2}{c}{TV} & & & & \\
\cmidrule(lr){3-4}\cmidrule(lr){5-6}
Method & Init. & PSNR$\uparrow$ & LPIPS$\downarrow$ & PSNR$\uparrow$ & LPIPS$\downarrow$ & \makecell{Init.\\time$\downarrow$} & \makecell{Train\\time$\downarrow$} & \makecell{Total\\time$\downarrow$} & FPS$\uparrow$ \\
\midrule
STG~\cite{stg} & SA & 27.86 & 0.077 & 23.97 & 0.146 & 15.3 & 29.1 & 44.4 & 376.2 \\
Taylor~\cite{taylor} & SA & 25.28 & 0.093 & 22.55 & 0.133 & 15.3 & 84.0 & 99.3 & 56.6 \\
4DGS~\cite{4dgs} & D0 & \best{28.30} & 0.083 & \best{26.04} & 0.111 & 5.2 & 47.4 & 52.6 & 157.1 \\
Swift4D~\cite{swift4d} & D0 & 27.37 & \third{0.066} & \third{25.58} & \second{0.083} & 10.0 & 27.7 & 37.7 & 147.1 \\
DeGauss~\cite{degauss} & D0 & \third{27.91} & 0.070 & 24.00 & 0.121 & 5.2 & 123.0 & 128.2 & 108.7 \\
\midrule
STG~\cite{stg} & FF & 27.75 & 0.082 & 21.88 & 0.251 & 0.1 & 27.6 & \third{27.7} & \third{378.1} \\
Taylor~\cite{taylor} & FF & 27.12 & 0.095 & 23.43 & 0.198 & 0.1 & 80.2 & 80.3 & 58.5 \\
4DGS~\cite{4dgs} & FF & 27.67 & 0.092 & 24.89 & 0.132 & 0.2 & 50.1 & 50.3 & 130.1 \\
Swift4D~\cite{swift4d} & FF & \second{27.98} & \second{0.062} & \second{25.81} & \best{0.065} & 5.1 & \third{27.4} & 32.5 & 137.1 \\
DeGauss~\cite{degauss} & FF & 27.39 & 0.074 & 25.55 & \third{0.088} & 0.2 & 124.5 & 124.7 & 97.7 \\
\midrule
Ours & D0 & 27.12 & 0.071 & 25.32 & 0.108 & 5.8 & \best{8.7} & \second{14.5} & \second{1953.5} \\
Ours & FF+MVR & 27.41 & \best{0.060} & 24.96 & 0.109 & 0.7 & \second{8.9} & \best{9.6} & \best{2046.3} \\
\bottomrule
\end{tabular}
\end{table*}

\begin{table*}[p]
\centering
\caption{Per-scene results on \emph{cook\_spinach} (Neural 3D~\cite{dynerf}). SA and D0 denote sparse COLMAP on all frames and dense COLMAP on the first frame, FF the feed-forward initialization, and MVR Multi-View Ray Relocation. TV denotes the temporally varying region at $\tau=0.1$ (\cref{sec:supp_region}). Times are in minutes.}
\label{tab:supp_perscene_cook_spinach}
\small
\setlength{\tabcolsep}{4pt}
\begin{tabular}{@{}ll cc cc ccc c}
\toprule
 & & \multicolumn{2}{c}{Overall} & \multicolumn{2}{c}{TV} & & & & \\
\cmidrule(lr){3-4}\cmidrule(lr){5-6}
Method & Init. & PSNR$\uparrow$ & LPIPS$\downarrow$ & PSNR$\uparrow$ & LPIPS$\downarrow$ & \makecell{Init.\\time$\downarrow$} & \makecell{Train\\time$\downarrow$} & \makecell{Total\\time$\downarrow$} & FPS$\uparrow$ \\
\midrule
STG~\cite{stg} & SA & \third{32.36} & 0.050 & 25.50 & 0.179 & 16.1 & 26.2 & 42.3 & \third{521.6} \\
Taylor~\cite{taylor} & SA & 31.40 & 0.059 & 25.43 & 0.180 & 16.1 & 71.8 & 87.9 & 77.8 \\
4DGS~\cite{4dgs} & D0 & 31.50 & 0.052 & \second{26.58} & 0.151 & 6.3 & 24.2 & 30.5 & 182.8 \\
Swift4D~\cite{swift4d} & D0 & \best{32.79} & \best{0.043} & \best{26.60} & \second{0.135} & 11.6 & 25.4 & 37.0 & 162.6 \\
DeGauss~\cite{degauss} & D0 & 30.62 & 0.064 & 21.56 & 0.306 & 6.3 & 124.9 & 131.2 & 92.5 \\
\midrule
STG~\cite{stg} & FF & 31.20 & 0.067 & 22.99 & 0.315 & 0.1 & \third{23.4} & \third{23.5} & 509.4 \\
Taylor~\cite{taylor} & FF & 31.37 & 0.063 & 24.08 & 0.209 & 0.1 & 74.8 & 74.9 & 55.5 \\
4DGS~\cite{4dgs} & FF & 32.35 & 0.060 & 26.41 & 0.161 & 0.1 & 24.9 & 25.0 & 156.9 \\
Swift4D~\cite{swift4d} & FF & 31.15 & \third{0.045} & \third{26.47} & \best{0.133} & 5.3 & 26.0 & 31.3 & 151.3 \\
DeGauss~\cite{degauss} & FF & 31.94 & 0.048 & 26.23 & \third{0.143} & 0.1 & 119.5 & 119.6 & 99.5 \\
\midrule
Ours & D0 & \second{32.71} & 0.046 & 25.94 & 0.156 & 7.2 & \best{8.8} & \second{15.9} & \best{2096.5} \\
Ours & FF+MVR & 32.07 & \second{0.044} & 26.21 & \third{0.143} & 0.7 & \second{9.3} & \best{10.1} & \second{2029.4} \\
\bottomrule
\end{tabular}
\end{table*}

\begin{table*}[p]
\centering
\caption{Per-scene results on \emph{cut\_roasted\_beef} (Neural 3D~\cite{dynerf}). SA and D0 denote sparse COLMAP on all frames and dense COLMAP on the first frame, FF the feed-forward initialization, and MVR Multi-View Ray Relocation. TV denotes the temporally varying region at $\tau=0.1$ (\cref{sec:supp_region}). Times are in minutes.}
\label{tab:supp_perscene_cut_roasted_beef}
\small
\setlength{\tabcolsep}{4pt}
\begin{tabular}{@{}ll cc cc ccc c}
\toprule
 & & \multicolumn{2}{c}{Overall} & \multicolumn{2}{c}{TV} & & & & \\
\cmidrule(lr){3-4}\cmidrule(lr){5-6}
Method & Init. & PSNR$\uparrow$ & LPIPS$\downarrow$ & PSNR$\uparrow$ & LPIPS$\downarrow$ & \makecell{Init.\\time$\downarrow$} & \makecell{Train\\time$\downarrow$} & \makecell{Total\\time$\downarrow$} & FPS$\uparrow$ \\
\midrule
STG~\cite{stg} & SA & 32.31 & 0.050 & 27.78 & 0.159 & 15.4 & 26.0 & 41.4 & \third{528.7} \\
Taylor~\cite{taylor} & SA & \best{33.44} & 0.053 & 28.40 & 0.148 & 15.4 & 83.0 & 98.4 & 57.5 \\
4DGS~\cite{4dgs} & D0 & 30.12 & 0.059 & 28.06 & 0.133 & 5.8 & 24.4 & 30.2 & 174.8 \\
Swift4D~\cite{swift4d} & D0 & 29.80 & 0.048 & 25.85 & 0.130 & 10.9 & 26.6 & 37.5 & 160.1 \\
DeGauss~\cite{degauss} & D0 & 31.75 & \third{0.045} & 28.53 & \second{0.113} & 5.8 & 123.2 & 129.0 & 100.2 \\
\midrule
STG~\cite{stg} & FF & 31.37 & 0.068 & 24.82 & 0.279 & 0.1 & \third{23.4} & \third{23.5} & 520.8 \\
Taylor~\cite{taylor} & FF & \third{32.91} & 0.051 & 28.16 & 0.131 & 0.1 & 68.8 & 68.9 & 68.8 \\
4DGS~\cite{4dgs} & FF & 32.80 & 0.060 & 28.72 & 0.140 & 0.1 & 25.2 & 25.3 & 152.5 \\
Swift4D~\cite{swift4d} & FF & \second{33.14} & \best{0.041} & \best{30.07} & \best{0.097} & 5.0 & 26.3 & 31.3 & 156.9 \\
DeGauss~\cite{degauss} & FF & 32.56 & 0.046 & 28.57 & 0.118 & 0.1 & 119.9 & 120.0 & 98.1 \\
\midrule
Ours & D0 & 32.81 & \third{0.045} & \third{28.82} & 0.128 & 6.6 & \best{10.0} & \second{16.6} & \best{1683.1} \\
Ours & FF+MVR & 32.49 & \second{0.042} & \second{28.90} & \third{0.116} & 0.7 & \second{10.1} & \best{10.8} & \second{1671.8} \\
\bottomrule
\end{tabular}
\end{table*}

\begin{table*}[p]
\centering
\caption{Per-scene results on \emph{flame\_salmon} (Neural 3D~\cite{dynerf}). SA and D0 denote sparse COLMAP on all frames and dense COLMAP on the first frame, FF the feed-forward initialization, and MVR Multi-View Ray Relocation. TV denotes the temporally varying region at $\tau=0.1$ (\cref{sec:supp_region}). Times are in minutes.}
\label{tab:supp_perscene_flame_salmon}
\small
\setlength{\tabcolsep}{4pt}
\begin{tabular}{@{}ll cc cc ccc c}
\toprule
 & & \multicolumn{2}{c}{Overall} & \multicolumn{2}{c}{TV} & & & & \\
\cmidrule(lr){3-4}\cmidrule(lr){5-6}
Method & Init. & PSNR$\uparrow$ & LPIPS$\downarrow$ & PSNR$\uparrow$ & LPIPS$\downarrow$ & \makecell{Init.\\time$\downarrow$} & \makecell{Train\\time$\downarrow$} & \makecell{Total\\time$\downarrow$} & FPS$\uparrow$ \\
\midrule
STG~\cite{stg} & SA & 28.34 & 0.070 & 21.43 & 0.144 & 16.6 & 29.3 & 45.9 & \third{376.5} \\
Taylor~\cite{taylor} & SA & 24.83 & 0.144 & 19.69 & 0.165 & 16.6 & 84.7 & 101.3 & 66.0 \\
4DGS~\cite{4dgs} & D0 & \second{28.68} & 0.085 & \second{24.07} & 0.089 & 5.5 & 48.2 & 53.7 & 151.0 \\
Swift4D~\cite{swift4d} & D0 & \best{28.99} & \second{0.061} & 23.36 & \third{0.075} & 10.7 & \third{27.0} & 37.7 & 146.6 \\
DeGauss~\cite{degauss} & D0 & \third{28.63} & \third{0.063} & \third{23.91} & \second{0.066} & 5.5 & 128.7 & 134.2 & 80.6 \\
\midrule
STG~\cite{stg} & FF & 27.58 & 0.074 & 18.21 & 0.221 & 0.1 & 27.3 & \third{27.4} & 362.5 \\
Taylor~\cite{taylor} & FF & 27.52 & 0.088 & 22.26 & 0.137 & 0.1 & 73.3 & 73.4 & 67.0 \\
4DGS~\cite{4dgs} & FF & 28.18 & 0.098 & 23.51 & 0.116 & 0.1 & 49.8 & 49.9 & 133.6 \\
Swift4D~\cite{swift4d} & FF & 28.31 & \third{0.063} & \best{24.25} & \best{0.064} & 5.1 & 27.5 & 32.6 & 137.0 \\
DeGauss~\cite{degauss} & FF & 27.93 & 0.066 & 23.43 & 0.078 & 0.1 & 124.5 & 124.6 & 85.5 \\
\midrule
Ours & D0 & 27.01 & 0.069 & 22.22 & 0.126 & 6.3 & \best{8.8} & \second{15.1} & \second{2096.0} \\
Ours & FF+MVR & 27.24 & \best{0.057} & 22.48 & 0.121 & 0.7 & \second{8.9} & \best{9.6} & \best{2120.9} \\
\bottomrule
\end{tabular}
\end{table*}

\begin{table*}[p]
\centering
\caption{Per-scene results on \emph{flame\_steak} (Neural 3D~\cite{dynerf}). SA and D0 denote sparse COLMAP on all frames and dense COLMAP on the first frame, FF the feed-forward initialization, and MVR Multi-View Ray Relocation. TV denotes the temporally varying region at $\tau=0.1$ (\cref{sec:supp_region}). Times are in minutes.}
\label{tab:supp_perscene_flame_steak}
\small
\setlength{\tabcolsep}{4pt}
\begin{tabular}{@{}ll cc cc ccc c}
\toprule
 & & \multicolumn{2}{c}{Overall} & \multicolumn{2}{c}{TV} & & & & \\
\cmidrule(lr){3-4}\cmidrule(lr){5-6}
Method & Init. & PSNR$\uparrow$ & LPIPS$\downarrow$ & PSNR$\uparrow$ & LPIPS$\downarrow$ & \makecell{Init.\\time$\downarrow$} & \makecell{Train\\time$\downarrow$} & \makecell{Total\\time$\downarrow$} & FPS$\uparrow$ \\
\midrule
STG~\cite{stg} & SA & 30.87 & 0.049 & 21.91 & 0.189 & 15.9 & 26.6 & 42.5 & 538.3 \\
Taylor~\cite{taylor} & SA & 29.76 & 0.062 & 19.92 & 0.212 & 15.9 & 81.5 & 97.4 & 61.7 \\
4DGS~\cite{4dgs} & D0 & 29.55 & 0.052 & 21.02 & 0.157 & 6.3 & 24.1 & 30.4 & 183.6 \\
Swift4D~\cite{swift4d} & D0 & 29.22 & 0.060 & \third{23.15} & \third{0.112} & 11.6 & 26.6 & 38.2 & 153.0 \\
DeGauss~\cite{degauss} & D0 & 28.80 & 0.064 & 16.35 & 0.429 & 6.3 & 124.9 & 131.2 & 96.3 \\
\midrule
STG~\cite{stg} & FF & 29.78 & 0.064 & 18.98 & 0.335 & 0.1 & \third{22.7} & \third{22.8} & \third{555.0} \\
Taylor~\cite{taylor} & FF & 29.58 & 0.063 & 18.85 & 0.243 & 0.1 & 66.0 & 66.1 & 72.7 \\
4DGS~\cite{4dgs} & FF & 29.73 & 0.058 & 22.34 & 0.153 & 0.1 & 24.3 & 24.4 & 171.7 \\
Swift4D~\cite{swift4d} & FF & 28.96 & 0.049 & 22.34 & 0.131 & 5.3 & 26.1 & 31.4 & 149.3 \\
DeGauss~\cite{degauss} & FF & \third{31.50} & \third{0.045} & 21.82 & 0.145 & 0.1 & 120.3 & 120.4 & 96.3 \\
\midrule
Ours & D0 & \second{32.50} & \second{0.036} & \second{26.21} & \second{0.097} & 7.1 & \best{9.8} & \second{16.9} & \best{2100.2} \\
Ours & FF+MVR & \best{32.59} & \best{0.034} & \best{26.29} & \best{0.094} & 0.7 & \second{10.4} & \best{11.0} & \second{1799.5} \\
\bottomrule
\end{tabular}
\end{table*}

\begin{table*}[p]
\centering
\caption{Per-scene results on \emph{sear\_steak} (Neural 3D~\cite{dynerf}). SA and D0 denote sparse COLMAP on all frames and dense COLMAP on the first frame, FF the feed-forward initialization, and MVR Multi-View Ray Relocation. TV denotes the temporally varying region at $\tau=0.1$ (\cref{sec:supp_region}). Times are in minutes.}
\label{tab:supp_perscene_sear_steak}
\small
\setlength{\tabcolsep}{4pt}
\begin{tabular}{@{}ll cc cc ccc c}
\toprule
 & & \multicolumn{2}{c}{Overall} & \multicolumn{2}{c}{TV} & & & & \\
\cmidrule(lr){3-4}\cmidrule(lr){5-6}
Method & Init. & PSNR$\uparrow$ & LPIPS$\downarrow$ & PSNR$\uparrow$ & LPIPS$\downarrow$ & \makecell{Init.\\time$\downarrow$} & \makecell{Train\\time$\downarrow$} & \makecell{Total\\time$\downarrow$} & FPS$\uparrow$ \\
\midrule
STG~\cite{stg} & SA & \third{32.45} & \third{0.047} & 23.77 & 0.202 & 16.0 & 26.0 & 42.0 & 560.8 \\
Taylor~\cite{taylor} & SA & 32.44 & 0.055 & \third{25.77} & \third{0.186} & 16.0 & 85.8 & 101.8 & 55.5 \\
4DGS~\cite{4dgs} & D0 & 32.37 & 0.056 & 24.39 & 0.250 & 6.3 & 24.2 & 30.5 & 183.4 \\
Swift4D~\cite{swift4d} & D0 & 30.25 & 0.053 & 22.99 & 0.228 & 11.7 & 26.0 & 37.7 & 161.3 \\
DeGauss~\cite{degauss} & D0 & 32.05 & 0.053 & 24.23 & 0.273 & 6.3 & 123.3 & 129.6 & 99.3 \\
\midrule
STG~\cite{stg} & FF & 31.71 & 0.056 & 22.81 & 0.272 & 0.1 & \third{22.2} & \third{22.3} & \third{573.3} \\
Taylor~\cite{taylor} & FF & 30.57 & 0.098 & 24.07 & 0.239 & 0.1 & 64.4 & 64.5 & 77.7 \\
4DGS~\cite{4dgs} & FF & 30.96 & 0.065 & 24.63 & 0.255 & 0.1 & 24.4 & 24.5 & 172.9 \\
Swift4D~\cite{swift4d} & FF & 32.23 & \third{0.047} & 25.15 & 0.202 & 5.2 & 25.8 & 31.0 & 154.7 \\
DeGauss~\cite{degauss} & FF & 31.53 & 0.056 & 23.91 & 0.284 & 0.1 & 119.9 & 120.0 & 95.3 \\
\midrule
Ours & D0 & \best{33.29} & \second{0.033} & \best{29.81} & \second{0.078} & 7.2 & \best{9.6} & \second{16.8} & \best{2220.8} \\
Ours & FF+MVR & \second{33.17} & \best{0.031} & \second{29.61} & \best{0.077} & 0.7 & \second{9.9} & \best{10.6} & \second{1889.3} \\
\bottomrule
\end{tabular}
\end{table*}

\end{document}